\documentclass[11pt,fullpage,letterpaper]{article} 
  \usepackage[margin=0.9in]{geometry}
  \usepackage{listings} 
    
\usepackage{mathrsfs} 
\usepackage{kpfonts,comment}
\usepackage[T1]{fontenc}
 
\usepackage{kz_style}

\newcommand{\kzedit}[1]{{\color{black}#1}} 

\newcommand{\xy}[1]{}{} 
 
\newcommand{\xynew}[1]{{\color{black}#1}}

\newenvironment{prevproof}[1]{\begin{proof}[Proof of \Cref{#1}]}{\end{proof}}

 \newcommand{\cth}{\text{ctt}(h)}

\title{\LARGE 
Partially Observable  Multi-Agent Reinforcement Learning\\ with Information Sharing} 
\author{Xiangyu Liu\thanks{University of Maryland, College Park. Email: \url{{xyliu999,kaiqing}@umd.edu}. A preliminary version of the paper has been accepted to the International Conference on Machine Learning (ICML) 2023 \citep{liu2023partially}.} 
\and \qquad\qquad\qquad  Kaiqing Zhang$^\dagger$   
}  
\date{}
     
\usepackage{cleveref}

\begin{document} 

\maketitle     

\begin{abstract}   
We study provable multi-agent reinforcement learning (RL) in the general framework of partially observable stochastic games (POSGs). To circumvent the known hardness results and the use of computationally intractable oracles, we advocate leveraging the potential \emph{information-sharing} among agents, a common practice in empirical multi-agent RL, and a standard model for multi-agent control systems with communication. We first establish several computational complexity results to justify the necessity of information-sharing, as well as the observability assumption that has enabled quasi-polynomial time and sample single-agent RL with partial observations, for  tractably solving POSGs. Inspired by the inefficiency of planning in the ground-truth model, we then propose to further \emph{approximate} the shared common information to construct an approximate model of the POSG, in which an approximate \emph{equilibrium} (of the original POSG) can be found in quasi-polynomial-time, under the aforementioned assumptions. Furthermore, we develop a partially observable multi-agent RL algorithm whose time and sample complexities are \emph{both} quasi-polynomial. Finally, beyond equilibrium learning, we extend our algorithmic framework to finding the \emph{team-optimal solution} in cooperative POSGs, i.e., decentralized partially observable Markov decision processes, a more challenging goal. We establish concrete computational and sample complexities under several structural assumptions of the model.  We hope our study could open up the possibilities of leveraging and even designing different  \emph{information structures}, a well-studied notion in control theory, for developing both sample- and computation-efficient partially observable multi-agent RL.  
\end{abstract}

\section{Introduction}\label{sec:Introduction}

Recent years have witnessed  the fast development  of reinforcement learning (RL) in a wide range of applications, including playing  Go games \citep{silver2017mastering}, robotics  \citep{lillicrap2016continuous,long2018towards}, video games  \citep{vinyals2019grandmaster,berner2019dota}, and autonomous driving \citep{shalev2016safe,sallab2017deep}. Many of these application domains by nature involve \emph{multiple decision-makers} operating in a common environment, with either aligned or misaligned objectives that are affected by their joint behaviors. This has thus inspired  surging research interests in multi-agent RL (MARL), with both deeper theoretical and  empirical understandings \citep{bu2008comprehensive,zhang2021multi,hernandez2019survey}. 

One central challenge of multi-agent learning in these  applications is the \emph{imperfection} of information, or more generally, the \emph{partial observability} of the  environments and other decision-makers. Specifically, each agent may possess \emph{different} information about the state
and action processes while making decisions. For example, in  vision-based  multi-robot learning and autonomous driving, each agent only  accesses a first-person camera to stream noisy  measurements of the object/scene,  
without accessing the observations or past actions of other agents. This is also sometimes referred to as \emph{information asymmetry} in game theory and decentralized decision-making \citep{behn1968class,milgrom1987informational,nayyar2013common,shi2016leader}. Despite its ubiquity in practice, theoretical understandings  
of MARL in partially observable settings remain scant. This is somewhat expected since even in  single-agent settings, planning and learning   under partial observability suffer from well-known computational and statistical hardness results  \citep{papadimitriou1987complexity,mundhenk2000complexity,jin2020sample}. The challenge  is known to be amplified for multi-agent decentralized decision-making \citep{witsenhausen1968counterexample,tsitsiklis1985complexity}. Existing partially observable MARL algorithms with finite-time/sample guarantees  either only apply to a small subset of highly structured (tree-like)  problems   \citep{zinkevich2007regret,kozuno2021learning}, or require computationally intractable oracles \citep{liu2022sample}.

With these hardness results that can be doubly exponential in the worst case,  even a \emph{quasi-polynomial} (time and sample complexity) algorithm could represent a non-trivial  improvement in partially observable MARL. 
In particular, we ask and attempt to answer  the following  question:
\begin{center}
    \emph{Can 
    partially observable MARL be made 
    \emph{both}  
    statistically \emph{and}  computationally  efficient?}  
\end{center} 
We provide some results towards answering the question positively, by leveraging the potential \emph{information sharing} among agents, together with a careful compression of the shared information. Indeed, the idea of information sharing has been widely used in empirical MARL, e.g., \emph{centralized}  training that aggregates all agents' information for more efficient training \citep{lowe2017multi,rashid2020monotonic}; it has also been widely used to model practical multi-agent  systems in decentralized control, e.g., those with delayed communication among agents  \citep{witsenhausen1971separation,nayyar2010optimal}. We detail our contributions below. 

\paragraph{Contributions.} We study provable multi-agent RL under the   framework of partially observable stochastic games (POSGs), with potential \emph{information  sharing} among  agents. First, we establish several computational   complexity results of solving POSGs \emph{in the presence of {information sharing}}, justifying its necessity, together with the necessity of the  \emph{observability}  assumption made in the recent  literature, which enabled single-agent partially observable RL without computationally intractable  oracles.   Second, we propose to further approximate the shared common information to construct an \emph{approximate model}, and characterize the computational  complexity of planning in this  model. We show that for several standard information-sharing structures, a simple \emph{finite-memory} compression can lead to \emph{expected}  approximate common information models in which planning an approximate equilibrium (in terms of solving the original POSG) has quasi-polynomial time complexity. Third, based on the planning results, we develop a  partially observable multi-agent RL algorithm \kzedit{whose time and sample complexities are \emph{both} quasi-polynomial, which we refer to as being \emph{quasi-efficient} for short (given that \emph{polynomial}-complexity algorithms are generally deemed as being \emph{efficient}).} {Fourth, beyond \emph{equilibrium} learning, we extend our framework and algorithm to \kzedit{finding} the \emph{team-optimal} solution in  cooperative POSGs, i.e., decentralized partially observable Markov decision processes (Dec-POMDPs), a more challenging goal. To this end, we identify several structural assumptions on the model under which quasi-efficient planning and learning become attainable.} 
To the best of our knowledge, this is the first study of  provable partially observable MARL with (quasi-)efficiency, with  \emph{both}  sample and  computational complexities. 
Key to our results is to carefully incorporate insights from both \emph{information structures/sharing}, a well-studied framework in decentralized stochastic control theory,  and the tractability conditions   investigated in recent reinforcement learning  theory.

\subsection{Related Work}\label{sec:detailed_related_work}

\paragraph{Information sharing in theory and practice.}  The idea of information-sharing and the study of more general \emph{information structures} have  been extensively studied in decentralized stochastic control \citep{witsenhausen1971separation,nayyar2010optimal,nayyar2013decentralized}, as well as dynamic games 
\citep{nayyar2013common,gupta2014common,ouyang2016dynamic}. The common-information-based approach in  the seminal works \cite{nayyar2013common,nayyar2013decentralized} provided significant   inspiration for our work. 
The information-sharing structures in these works have enabled  \emph{backward-induction-based}  planning algorithms even in this decentralized setting. Performance bounds of information compression in such a framework were later derived in \cite{mao2020information,kao2022common}. 
However, neither computation nor sample complexities of the algorithms were analyzed in these  works. On the other hand, information-sharing has become a normal practice in empirical MARL 
\citep{lowe2017multi,sunehag2018value,rashid2020monotonic}, usually instantiated via the 
 so-called \emph{centralized training}, where all agents' information was shared in the training  to improve learning efficiency. 
 However, information-sharing/structure has not been fully investigated  in the theoretical studies of MARL.

\paragraph{Decentralized stochastic control and decision-making.} Decentralized stochastic control and decision-making are known to have unique challenges, compared to the  single-agent and centralized counterpart, since the seminal works \cite{witsenhausen1968counterexample,tsitsiklis1985complexity}. In particular, \cite{tsitsiklis1985complexity} showed that variations of the classical ``team decision problem'' can be \texttt{NP-hard}. Later, \cite{bernstein2002complexity} showed that planning in Dec-POMDPs, a special class of POSGs with an identical reward function shared across agents, can be \texttt{NEXP-hard} in finding the team-optimal solution. \cite{hansen2004dynamic} provided a  popular POSG planning algorithm, though without any complexity guarantees. There also exist other approximate/heuristic algorithms for solving POSGs \citep{emery2004approximate,kumar2009dynamic,horak2017heuristic}. 

\paragraph{RL in partially observable environments.} It is known that in general,  planning in even \emph{single-agent}  POMDPs can be \texttt{PSPACE-complete} \citep{papadimitriou1987complexity} and thus computationally hard. Statistically,  learning POMDPs can also be hard in general  \citep{krishnamurthy2016pac,jin2020sample}. There has thus been a growing body of literature on RL in POMDPs with additional assumptions, e.g., \cite{azizzadenesheli2016reinforcement,jin2020sample,liu2022partially}. However, these works only focused on \emph{statistical} efficiency, and the algorithms usually required \emph{computationally intractable}  oracles. More recently, \cite{golowich2022planning} has identified the condition of $\gamma$-observability in POMDPs (firstly introduced in \cite{even2007value}), which enabled a quasi-polynomial-time planning algorithm for such POMDPs. Subsequently, \cite{golowich2022learning} has developed an RL algorithm based on the planning one in \cite{golowich2022planning}, which was both sample and computation (quasi-)efficient. {The key enabler of these (quasi-)efficient algorithms is the use  of the \emph{finite-memory} policy class, whose (near-)optimality has also been studied lately in \cite{kara2022near,kara2022convergence}, under different assumptions on both the transition dynamics and the observation channels.}   Rather than  statistical and computational complexity guarantees, \cite{subramanian2022approximate} has analyzed the performance bounds of general approximate information states (AIS) in partially observable environments. Our finite-memory compression may be viewed as a kind of AIS, although it does not satisfy the \emph{uniform}  approximation conditions in \cite{subramanian2022approximate} (and also in \cite{mao2020information,kao2022common}). In fact, relaxing such conditions to \emph{expected} versions is the key to obtaining our (quasi-)efficient sample and computational complexities (cf. Remark \ref{remark:compare_kao_mao}). Other information compression results include \cite{tang2024information} for dynamic games, and \cite{sinha2023asymmetric,caiprovable} for RL with asymmetric information.

\paragraph{Provable multi-agent reinforcement learning.} There has been a fast-growing literature on provable MARL algorithms with sample efficiency guarantees, e.g.,  \cite{bai2020near,liu2021sharp,zhang2020model,xie2020learning,zhang2021derivative,wei2021last,daskalakis2020independent,jin2021v,song2021can,daskalakis2022complexity,mao2022improving,leonardos2022global,zhang2021gradient,ding2022independent,chen2023finite}. However, these works have been exclusively focused on the fully observable setting of Markov/stochastic  games. The only  MARL algorithms under partial observability that enjoy finite-sample guarantees, to the best of our knowledge, are those in \cite{liu2022sample,kozuno2021learning}. However, the algorithm in \cite{kozuno2021learning} only applied to POSGs with certain tree-structured transitions, while that in \cite{liu2022sample} required computationally intractable oracles. In general, information-sharing/structure has not been fully investigated  in the theoretical studies of MARL with \emph{finite-sample} and \emph{computation complexities}. One exception is \cite{kao2022decentralized}, which exploited a special \emph{hierarchical}   information structure in the bandits and MDP settings. Another exception is  \cite{altabaa2024role},  which appeared online after the acceptance of the conference version of this paper \citep{liu2023partially}, and also incorporated (general)   \emph{information structure} considerations into the algorithm design and analyses. However, 
the algorithms in \cite{altabaa2024role} also required   {computationally intractable} oracles, with a focus on statistical-tractability only.  

\paragraph{Independent result in \cite{golowich2023planning}.} We note that after the acceptance to ICML 2023 of the preliminary version of the paper, an updated version of  \cite{golowich2022planning}  in its proceedings form appeared online, i.e., \cite{golowich2023planning}. In \cite{golowich2023planning}, a quasi-polynomial-time  planning algorithm for solving a class of partially observable stochastic games was also discussed. There are several differences compared to our results. First,  in the class of POSGs considered in \cite{golowich2023planning}, the observation is identical for all the agents, and each agent  has access to the joint action history of all the agents. Notably, this setting exactly corresponds to the \emph{fully-sharing/symmetric-information} case covered by our information-sharing framework (see \Cref{exam:sym} in \Cref{sec:example}). Moreover, we study both Nash equilibrium (NE) in cooperative/zero-sum games and correlated equilibrium (CE), coarse correlated equilibrium (CCE) in general-sum games, while \cite{golowich2023planning} only focused on finding   CCE in general-sum games; 
we also establish a result for \emph{learning} equilibria in POSGs with both quasi-polynomial sample and computational complexities,  while \cite{golowich2023planning} only focused    planning with model knowledge. Additionally, we also establish results for team-optimum learning for Dec-POMDPs under certain structural conditions.

\paragraph{Notation.} For two sets $B$ and $D$, we define $B\setminus{D}$ as the set of elements that are in $B$ but not in  $D$. We use $\emptyset$ to denote the empty set and $[n]:=\{1, \cdots, n\}$.
For integers $a\le b$, we denote a sequence $(x_a, x_{a+1}, \cdots, x_{b})$ by $x_{a:b}$. If $a>b$, then it denotes an empty sequence. When the sequence index starts from $m$ and ends at $n$, we will treat $x_{a:b}$ as $x_{\max\{a,m \}: \min\{b, n\}}$. For an event $\cE$, we use $\bm{1}$ to denote the indicator function such that $\bm{1}(\cE) = 1$ if the event $\cE$ is true and $0$ otherwise. 
For a finite set $B$, we let $\Delta(B)$ denote the set of distributions over  $B$. {For two probability distributions $p$, $q$, we define the 2-R\'enyi  divergence as  $D_2(p||q):=\log\EE_{x\sim p}\left[\frac{p(x)}{q(x)}\right]$. We also define $p\ll q$ if $q(x) = 0$ implies $p(x)=0$.}

\section{Preliminaries}\label{sec:prelim}

\subsection{POSGs and information sharing}\label{sec:prelim_model}

\paragraph{Model.} 
Formally, we define a finite-horizon POSG with $n$ agents by a tuple $\cG =(H, \mathcal{S}, \{\mathcal{A}_i\}_{i=1}^{n},\{\mathcal{O}_i\}_{i=1}^{n},\mathbb{T},\mathbb{O}, \mu_1, \{r_i\}_{i=1}^{n})$, where  $\mathcal{S}$ denotes the state space with $|\mathcal{S}|=S$, $\mathcal{A}_i$ denotes the action space for the $i^{th}$ agent with $|\mathcal{A}_i|=A_i$, and $H$ denotes the length of each episode. We denote by $a_h:= (a_{1,h}, \cdots, a_{n,h})$ the joint action of all the $n$ agents at time step  $h$, and by $\mathcal{A}=\mathcal{A}_1\times \cdots\times \mathcal{A}_n$ the joint action space with $|\mathcal{A}| = A=\prod_{i=1}^{n}A_i$. We use $\mathbb{T}=\{\mathbb{T}_h\}_{h\in [H]}$ to denote the collection of the transition matrices, so that $\mathbb{T}_h(\cdot\given s, {a})\in\Delta({\mathcal{S}})$ gives the probability of the next state if joint action ${a}$ is taken at state $s$ and step $h$. In the following discussions, for any given $a$, we treat $\TT_h(a)\in\RR^{|\cS|\times|\cS|}$ as a matrix, where each row gives the probability for the next state.  {We use} $\mu_1$ {to} denote the distribution of the initial state $s_1$, and $\cO_i$ {to} denote the observation space for the $i^{th}$ agent with $|\mathcal{O}_i|=O_i$.  We denote by ${o}:=(o_1, \dots, o_n)$ the joint observation of all the $n$ agents, and by $\mathcal{O}:=\mathcal{O}_1\times\dots\times\mathcal{O}_n$ with $|\mathcal{O}|=O=\prod_{i=1}^{n}O_i$. We use $\mathbb{O}=\{\mathbb{O}_h\}_{h\in [H + 1]}$ to denote the collection of the joint emission matrices, so that $\mathbb{O}_h(\cdot\given s)\in\Delta({\mathcal{O}})$ gives the emission distribution over the joint observation space $\mathcal{O}$ at state $s$ and step $h$.  For notational convenience, we will at times adopt the matrix convention, where $\OO_h$ is a matrix with rows $\OO_h(\cdot\given s)$. We also denote by $\OO_{i, h}(\cdot\given s)\in \Delta(\cO_i)$ the marginalized emission for the $i^{th}$ agent at state $s$. Finally, $r_i=\{r_{i, h}\}_{h\in [H]}$ is a collection of reward functions, so that $r_{i, h}(s_h, a_h)\in[0, 1]$ is the reward of the $i^{th}$ agent given the state $s_h$ and (joint) action $a_h$ taken at step $h$. This general formulation of POSGs includes several  important subclasses. For example, decentralized partially observable Markov decision processes (i.e., Dec-POMDPs)  are POSGs {where the agents share} a common reward function, {i.e.,} $r_i = r, \forall i\in [n]$; zero-sum POSGs are POSGs with $n=2$ and $r_1 + r_2= 1$. Note that we require $r_1 + r_2$ to be $1$ instead of $0$ to be consistent with our assumption that $r_{i, h}\in[0, 1]$ for each $i\in\{1,2\}$  and $h\in[H]$, and this requirement does not lose any optimality as one can always subtract the constant-sum offset to attain a zero-sum structure. Hereafter, we may use the terminology \emph{cooperative POSG} and \emph{Dec-POMDP}  interchangeably.

\paragraph{Information sharing, common and private information.} 
The $i^{th}$ agent at step $h$ in the POSG maintains its own information,  $\tau_{i, h}$, a collection of (potentially partial) historical observations and actions  at step $h$, namely,  $\tau_{i, h} \subseteq \{o_1, a_{1}, o_2, \cdots, a_{h-1}, o_{h}\}$, and the collection of such histories at step $h$ is denoted by $\mathcal{T}_{i, h}$.  
{In many practical examples (see some concrete ones in \Cref{sec:example}), agents may share part of the history with each other, which may introduce more structures in the game that enable both sample and computation efficient learning. The information sharing splits  the full history into the \emph{common/shared} and the \emph{private}  information for each agent.} 
The \emph{common information} at step $h$ is a subset of the joint history $\tau_h$: ${c}_h \subseteq \{o_1, a_1, o_2, \cdots, a_{h-1}, o_{h}\}$, which is available to \emph{all the agents} in the system, and the collection of the common information is denoted as $\cC_h$ and  we define $C_h=|\cC_h|$. Given the common information ${c}_h$, each agent also has the private information $p_{i, h} = \tau_{i, h}\setminus{c_h}$, where the collection of the private information for the $i^{th}$ agent is denoted as $\cP_{i, h}$ and its  cardinality as  $P_{i, h}$. The joint private information at step $h$ is denoted as $p_{h}$, where the collection of the joint private history is given by $\cP_h = \cP_{1, h}\times\cdots\times\cP_{n, h}$ and the corresponding cardinality is $P_{h} = \prod_{i=1}^nP_{i, h}$. We allow $c_h$ or $p_{i, h}$ to take the special value of  $\emptyset$ when there is no common or private information. {In particular, when $\cC_h = \{\emptyset\}$, the problem reduces to a general POSG without any favorable information structure; when $\cP_{i, h} = \{\emptyset\}$, every agent holds the same history, and it reduces to a POMDP when the agents share a common reward function, for which the goal is usually to find the team-optimal policy.}

{Throughout,} we also assume that the common information and private information evolve over time properly, as formalized below.  

\begin{assumption}[Evolution of common and private information]\label{evo}
We assume that common information and private information evolve over time as follows:
\begin{itemize}
    \item Common information $c_h$ is non-decreasing with time, that is, $c_h\subseteq c_{h+1}$ for all $h$. Let $z_{h+1} = c_{h+1}\setminus c_h$. Thus, $c_{h+1}=c_h \cup z_{h+1}$. Further, we have 
   \begin{equation}
   	 \label{common_trans}
        z_{h+1} = \chi_{h+1}(p_h, a_h, o_{h+1}),
    \end{equation}
    where $\chi_{h+1}$ is a fixed transformation. We use $\cZ_{h+1}$ to denote the collection of all $z_{h+1}$ at step $h$. 
    \item Private information evolves according to: 
    \begin{equation}\label{private_trans}
        p_{i, h+1} = \xi_{i, h+1}(p_{i, h}, a_{i, h}, o_{i, h+1}),
    \end{equation}
    where $\xi_{i, h+1}$ is a fixed transformation.
\end{itemize}
\end{assumption} 

Equation \eqref{common_trans} states that the increment in the common information, and thus the common information at the next step $c_{h+1}$, depends on the ``new'' information $\{a_h, o_{h+1}\}$ generated between steps $h$ and $h+1$ and part of the ``old'' information $p_h$. {The incremental common information can be generated by certain sharing and communication protocols among agents.} Equation \eqref{private_trans} implies that the evolution of private information only depends on the newly generated private information $a_{i, h}$ and $o_{i, h+1}$. These evolution rules are standard in the literature \citep{nayyar2013common, nayyar2013decentralized}, specifying the source of common information and private information. {Based on such evolution rules, we define $\{f_h\}_{h\in [H+1]}$ and $\{g_h\}_{h\in [H+1]}$, where $f_h:\cA^{h-1}\times\cO^{h}\rightarrow \cC_h$ and $g_h:\cA^{h-1}\times\cO^{h}\rightarrow \cP_h$ for $h\in [H+1]$, as the mappings that map the joint history to common information and joint private information, respectively.}

\subsection{Policies and value functions}
We define a stochastic policy for the $i^{th}$ agent at step $h$ as: 
\begin{equation}\label{policy}
    {\pi}_{i, h}: \Omega_h\times\cP_{i, h}\times \cC_{h}\rightarrow \Delta({\mathcal{A}_i}),
\end{equation}
where $\Omega_h$ is a space of random seeds   shared among agents. 
The corresponding policy class is denoted as ${\Pi}_{i, h}$. {Hereafter, unless otherwise noted, when referring to \emph{policies}, we mean the policies given in the form of \eqref{policy}, which map the \emph{available information} of the $i^{th}$ agent, i.e., the private information and the common information, together with the potentially local random seed $\omega_{i, h}\in \Omega_h$, to the distribution over her actions.} 
We further denote by $\Pi_i = \times_{h\in [H]} \Pi_{i, h}$ the policy space for the $i^{th}$ agent and $\Pi$ as the joint policy space. 
As a special case, we define the {space of}  
\emph{deterministic}  policy {as $\Tilde{\Pi}_i$, where $\Tilde{\pi}_i\in\Tilde{\Pi}_i$} maps the private information and common information to a \emph{deterministic} action for the $i^{th}$ agent, and denote the joint space of such policies as $\Tilde{\Pi}$. 

One important  concept   in 
the {common-information-based   framework} 
is called the \emph{prescription} \citep{nayyar2013decentralized,nayyar2013common}, defined for the $i^{th}$ agent at step $h$ as
$
\gamma_{i, h}: \cP_{i, h}\rightarrow \Delta(\cA_i).
$ 
With such a prescription function, agents can take actions purely based on their local private information. We define $\Gamma_{i, h}$ as the function class for prescriptions, and $\Gamma_h$ as the function class of joint prescriptions. Intuitively, the \emph{partial} function $\pi_{i, h}(\cdot\given\omega_{i, h}, \cdot, c_h)$  is a prescription given some $\omega_{i, h}$ and $c_h$. We will define $\pi_i$ as a sequence of policies for the $i^{th}$ agent at all steps $h\in[H]$, i.e., $\pi_i = \{\pi_{i, 1}, \cdots, \pi_{i, H}\}$. 
A (potentially correlated) joint policy 
is denoted as $\pi=\pi_1\odot\pi_2\cdots\odot\pi_n\in \Pi$.
 A \emph{product} policy is denoted as $\pi=\pi_1\times\pi_2\cdots\times\pi_n\in \Pi$ if {the distributions of drawing each seed $\omega_{i, h}$ for different agents are \emph{independent}}. 
 Furthermore, sometimes, we might resort to 
{\emph{deterministic}  joint policies  with \emph{joint history} as input} (which could potentially go beyond $\Pi$): $\pi=\{\pi_{1}, \pi_2, \cdots, \pi_H\}$,  where $\pi_h$ is defined as: $\pi_h: \cA^{h-1}\times\cO^{h}\rightarrow \cA.$ 
We denote the collection of such policies as $\Pi^{\mathrm{det}}$, and note that $\Pi\subseteq \Delta(\Pi^{\mathrm{det}})$. 
{For any policy $\pi\in\Delta(\Pi^{\mathrm{det}})$  and event $\cE$, we write $\PP^{\cG}_{s_{1:h}, a_{1:h-1}, o_{1:h}\sim \pi_{1:h-1}}(\cE)$ to denote the probability of $\cE$ when $\{s_{1:h},a_{1:h-1},o_{1:h}\}$ is drawn from a trajectory following the policy $\pi_{1:h-1}$ from step $1$ to $h-1$ in the model $\cG$. We will use the shorthand notation $\PP_{h}^{\pi_{1:h-1}, \cG}(\cdot)$ if the definition of $\{s_{1:h},a_{1:h-1},o_{1:h}\}$ is evident. At times, if the time index $h$ is evident, we will write it as $\PP^{\pi, \cG}_{h}(\cdot)$. If the event $\cE$ does not depend on the choice of $\pi$, we will use $\PP_{h}^{\cG}(\cdot)$ and omit $\pi$. Moreover, we will write $\EE_{s_{1:h}, a_{1:h-1}, o_{1:h}\sim \pi}^\cG[\cdot]$ or $\EE_{\pi}^\cG[\cdot]$ 
to denote the  expectations over the trajectories under policy $\pi$, and use the shorthand notation $\EE^\cG[\cdot]$ if the expectation does not depend on the choice of $\pi$. Furthermore, if we are given some model $\cM$ (other than $\cG$), 
the notation of $\PP_{h}^\cM(\cdot)$, $\EE^{\cM}_\pi[\cdot]$, and $\EE^{\cM}[\cdot]$ are defined in the same way.} We will hereafter use \emph{strategy} and   \emph{policy} interchangeably. 
We are now ready to define the \emph{value function} for each agent under our framework:
\begin{definition}[Value function {with information sharing}]\label{def:g_value_functioin}
For each agent $i\in [n]$ and step $h\in [H]$, given common information $c_h$ and joint policy  ${\pi}=\{\pi_{i}\}_{i=1}^{n}\in\Pi$, 
the 
\emph{value function conditioned on the common information} of the $i^{th}$ agent is defined as: 
    \[V^{{\pi}, \cG}_{i, h}(c_h) := \EE_{\pi}^\cG\left[\sum_{h^{\prime}=h}^{H}r_{i, h^{\prime}}(s_{h^\prime}, a_{h^\prime})\biggiven c_{h}\right],\] where the expectation   is taken over the randomness from the model $\cG$, policy $\pi$, and the random seeds. For any $ c_{H+1}\in \cC_{H+1}$: $V^{{\pi}, \cG}_{i, H+1}(c_{H+1}) := 0$. For the value function at the first step, we denote $V_{i, 1}^{\pi, \cG}(\emptyset):=\EE^\cG[V_{i, h}^{\pi, \cG}(c_1)]=\EE_{\pi}^\cG[\sum_{h=1}^H r_{i, h}(s_h, a_h)]$, where the expectation is taken over the randomness of $c_1$, which is a function of $o_1$ and does not depend on $\pi$.  
\end{definition} 

Correspondingly, we can define the prescription-value function $Q_{i, h}^{\pi, \cG}(c_h, \gamma_h)$, a generalization of the \emph{action-value} function in  MDPs, indicating the expected return for {the $i^{th}$ agent} when all the agents firstly adopt the prescriptions $\{\gamma_{j, h}\}_{j\in [n]}$ at step $h$ and then follow $\pi$ (cf. \Cref{def:pres_value}).

\subsection{Solution concepts}

With the definition of the value functions, we can accordingly define the solution concepts,  \emph{$\epsilon$-NE} (and similarly \emph{$\epsilon$-CCE}, \emph{$\epsilon$-CE}), and \emph{$\epsilon$-team optimum} under the information-sharing framework as follows.  

\begin{definition}[$\epsilon$-approximate Nash equilibrium {with information sharing}]
For any $\epsilon\ge 0$, a product policy $\pi^{\star}\in\Pi$ is an $\epsilon$-Nash equilibrium  of the POSG $\cG$ if
\[
    \operatorname{NE-gap}(\pi^\star):=\max_i \left(\max_{\pi_{i}^{\prime}\in \Pi_i}V^{\pi_i^{\prime}\times {\pi^{\star}_{-i}}, \cG}_{i, 1}(\emptyset)-V^{{\pi}^{\star}, \cG}_{i, 1}(\emptyset)\right)\le  \epsilon.
\]
\end{definition}
\begin{definition}[$\epsilon$-approximate coarse correlated equilibrium {with information sharing}]
For any $\epsilon\ge 0$, a joint policy $\pi^{\star}\in\Pi$
is an $\epsilon$-approximate coarse correlated equilibrium of the POSG $\cG$ with information sharing if:
\[
    \operatorname{CCE-gap}(\pi^\star):=\max_i \left(\max_{\pi_{i}^{\prime}\in \Pi_i}V^{\pi_i^{\prime}\times {\pi^{\star}_{-i}}, \cG}_{i, 1}(\emptyset)-V^{{\pi}^{\star}, \cG}_{i, 1}(\emptyset)\right)\le  \epsilon.
\]
\end{definition}

\begin{definition}[$\epsilon$-approximate correlated equilibrium {with information sharing}]\label{def:CCE}
For any $\epsilon\ge 0$, a joint policy $\pi^{\star}\in\Pi$ is an $\epsilon$-approximate correlated equilibrium of the POSG $\cG$ with information sharing if: 
\[
    \operatorname{CE-gap}(\pi^\star):=\max_i \left(\max_{\phi_i}V^{(\phi_i\diamond \pi_i^\star)\odot {\pi^{\star}_{-i}}, \cG}_{i, 1}(\emptyset)-V^{{\pi}^{\star}, \cG}_{i, 1}(\emptyset)\right)\le  \epsilon,
\] 
where $\phi_i$ is called a \emph{strategy modification} and $\phi_i = \{\phi_{i, h, c_h, p_{i, h}}\}_{h, c_h, p_{i, h}}$, with  each $\phi_{i, h, c_h, p_{i, h}}:\cA_i\rightarrow \cA_i$ being a mapping from the action set to itself. The space of $\phi_i$ is denoted as $\Phi_i$. The composition $\phi_i\diamond\pi_i$ will work as follows: at the step $h$, when the $i^{th}$ agent is given $c_h$ and $p_{i, h}$, the action chosen to be $(a_{1, h}, \cdots, a_{i, h}, \cdots,a_{n, h})$ will be modified to $(a_{1, h}, \cdots, \phi_{i, h, c_h, p_{i, h}}(a_{i, h}), \cdots, a_{n, h})$. Note that this definition extends those in  \cite{song2021can,liu2021sharp,jin2021v} to our settings when there exists common information, and is a natural generalization of the definition in the normal-form game case  \citep{roughgarden2010algorithmic}.
\end{definition}

\begin{definition}[$\epsilon$-approximate team-optimum in Dec-POMDPs {with information sharing}]\label{def:team} When the reward functions $r_{i,h}$ are identical for all $i\in[n]$, i.e., $r_{i, h} = r_h$,  the POSG reduces to a Dec-POMDP, and a policy ${\pi}^{\star}\in \Tilde{\Pi}$ is {an $\epsilon$-approximate} team-optimal policy if: 
$V^{{\pi}^{\star}, \cG}_{1}(\emptyset)\ge \max_{\pi^{\prime}\in \Tilde{\Pi}}V^{\pi^{\prime}, \cG}_{1}(\emptyset)-\epsilon$, where we have omitted the agent index for the value function. 
\end{definition}

It is also worth noting that, under given information-sharing structures, the team-optimal solution is always a NE {in the Dec-POMDP setting}, and in general, 
a NE is always a CE, and a CE is always a CCE.

\section{Information Sharing in Applications}\label{sec:example}

The information-sharing structure can indeed be common in real-world applications. 
For example, for a self-driving car to avoid collision and successfully navigate, the other cars from the same fleet/company would usually  communicate with each other (possibly with delays) about the road situation. The separation between common information and private information then arises naturally \citep{gong2016constrained}. Similar examples can also be found in cloud computing and power systems \citep{altman2009stochastic}. Here, we outline several  representative  information-sharing structures that were  firstly introduced by \cite{nayyar2013common} and can 
fit into our algorithmic framework.

\begin{example}[One-step delayed 
sharing]\label{exam-1} At any step $h\in[H+1]$, the common and private information are  given as $c_h = \{o_{1:h-1}, a_{1:h-1}\}$ and $p_{i, h} = \{o_{i, h}\}$, respectively. In other words, the agents  share all the action-observation history until the previous step $h-1$, with only the new observation being the private information. This model has been shown useful for power control \citep{altman2009stochastic}.
\end{example}  

\begin{example}[State controlled by one controller with asymmetric delay sharing]\label{exam-2} We assume there are $2$ agents for convenience. It extends naturally to $n$-agent settings. Consider the case where the state dynamics are controlled by agent  $1$, i.e., $\TT_h(\cdot\given s_h, a_{1, h}, a_{2, h}) = \TT_{h}(\cdot\given s_h, a_{1, h}, a_{2, h}^\prime)$  for any $h, s_h, a_{1, h}, a_{2, h},a_{2, h}^\prime$. For the cooperative setting that aims to find approximate \emph{team-optimum}  later (cf. \Cref{dec-pomdp}), we additionally assume, for this example, that the reward function has an additive structure, i.e., $r_h(s_h, a_h)=\sum_{j\in[n]}r_{j, h}(s_h, a_{j, h})$ for some functions $\{r_{j, h}\}_{j\in[n]}$. The information structure is given as $c_{h} = \{o_{1, 1:h}, o_{2, 1:h-d}, a_{1, 1:h-1}\}$, $p_{1, h} = \emptyset$, $p_{2, h} = \{ o_{2, h-d+1:h}\}$, i.e., agent $1$'s observations are available to agent $2$ instantly, while agent $2$'s observations are available to agent $1$ with a delay of $d\ge 1$ time steps. {We will regard $d$ as a constant throughout.} 
This kind of asymmetric sharing is common in  network routing \citep{pathak2008measurement}, {where packages arrive at different hosts with different delays, leading to asymmetric delay sharing among the hosts.} 
\end{example} 

\begin{example}[Symmetric information game]\label{exam:sym} Consider the case when all observations and actions are available for all the agents, and there is no private information. Essentially, we have $c_h = \{o_{1:h}, a_{1:h-1}\}$ and $p_{i, h} = \emptyset$. We will also denote this structure  as \emph{fully sharing}  hereafter.	
\end{example}

\begin{example}[Information sharing with one-directional-one-step delay]\label{exam-4} Similar to the previous cases, we also assume there are $2$ agents for ease of exposition,  and the case can be generalized to multi-agent cases straightforwardly. Similar to the one-step delay case, we consider the situation where all observations of agent $1$ are available to agent $2$, while the observations of agent $2$ are available to agent $1$ with one-step delay. All the past actions are available to both agents. That is, in this case, $c_{h} = \{o_{1, 1:h}, o_{2,2:h-1}, a_{1:h-1}\}$, and agent $1$ has no private information, i.e., $p_{1, h} =\emptyset$, and agent $2$ has private information  $p_{2, h} = \{o_{2, h}\}$.
\end{example} 

\begin{example}[Uncontrolled state process]\label{exam-5} Consider the case where the state transition does not depend on the actions, that is, $\TT_h(\cdot\mid s_h, a_h) = \TT_h(\cdot\mid s_h, a_h^{\prime})$ for any $h, s_h, a_h, a_h^{\prime}$. For the cooperative setting that aims to find approximate \emph{team-optimum}  later (cf. \Cref{dec-pomdp}), as in \Cref{exam-2}, we additionally assume, for this example, that the reward function has an additive structure, i.e., $r_h(s_h, a_h)=\sum_{j\in[n]}r_{j, h}(s_h, a_{j, h})$ for some functions $\{r_{j, h}\}_{j\in[n]}$.
    An example of this case is the information structure where controllers share their observations with a general delay of $d\ge 1$ time steps. In this case, the common information is $c_h = \{o_{1:h-d}\}$ and the private information is $p_{i, h} = \{o_{i, h-d+1:h}\}$. Such information structures {can be used to model} repeated games with incomplete information \citep{aumann1995repeated}.
\end{example}

\section{Hardness and  Planning with Exact Model}\label{sec:main_1}

\subsection{Hardness on finding equilibria}\label{subsec:hardness_main_body}

Recently,  reference \cite{golowich2022planning} considered \emph{observable} POMDPs (firstly introduced in \cite{even2007value}) that rule out the ones  with uninformative observations, {for which} computationally (quasi)-efficient  algorithms can be developed. In the hope of obtaining computational (quasi)-efficiency for  POSGs (including Dec-POMDPs), we thus make a similar observability assumption on the \emph{joint} emission matrix as below. Note that this is weaker than making the assumption on the \emph{individual} emission matrix of each agent.

\begin{assumption}[$\gamma$-observability]
Let $\gamma> 0$. For $h\in [H]$, we say that the matrix $\OO_h$ satisfies the $\gamma$-observability assumption if for each $h\in[H]$, any $b, b^\prime\in \Delta(\cS)$,
\[
\left\|\mathbb{O}_{h}^{\top} b-\mathbb{O}_{h}^{\top} b^{\prime}\right\|_{1} \geq \gamma\left\|b-b^{\prime}\right\|_{1}.
\]
A POSG (Dec-POMDP) satisfies $\gamma$-observability if all its $\OO_h$ for  $h\in[H]$ do so.\label{observa}
\end{assumption}

Examples of an observation matrix which satisfies $\gamma$-observability include
the random channel which outputs the hidden state with probability $\gamma$, and otherwise outputs a random state uniformly (i.e., from a ``noisy sensor'') or an extra dummy observation $\emptyset$ deterministically (i.e., from a ``failure mode''). 
Meanwhile, although the tractability of NE/CE/CCE in 
normal-form games has been extensively studied, its formal tractability in POSGs has been less studied.
Here by the following proposition, we show that both Assumption \ref{observa} and {some favorable} information-sharing {structure} are \emph{necessary} for   NE/CE/CCE to be  {computationally} tractable, even for the special classes of zero-sum POSGs and  cooperative POSGs. Specifically, they are necessary in the sense that missing either one of them would make seeking approximate NE/CE/CCE computationally hard, whose proof  is deferred to \Cref{subsec:hardness_proof}.

\begin{proposition}\label{posg_hardness_1} 
For zero-sum or cooperative POSGs with only information-sharing structures, {or {only} Assumption  \ref{observa}, but not both,} computing $\epsilon$-NE{/CE/CCE} is \texttt{PSPACE-hard}. 
\end{proposition}

{Hence, we will now focus on planning and learning under these assumptions.

\subsection{Planning with strategy-independent common belief}\label{subsec:plan-exact}

For both optimal and equilibrium policy computation, it is known that  
\emph{backward induction} is  one of the most useful  approaches for solving (fully-observable) stochastic games. However, the essential impediment to applying backward induction in \emph{asymmetric-information/partially observable}  dynamic   games is the fact that an agent's posterior beliefs about the system state and about other agents' information may depend on the \emph{strategies}
 used by the agents in the past. If the {nature of system dynamics} and the {information structure} of the game ensure that the agents' posterior beliefs are \emph{strategy independent}, then a backward induction can be derived for equilibrium computation  \citep{nayyar2013common,gupta2014common}. We formalize this conceptual argument as the following assumption}.

\begin{assumption}[Strategy independence of beliefs]\label{str_indi}
Consider any step $h\in[H]$, any choice of joint policies ${\pi}\in\Pi$, and any realization of common information $c_h$ that has a non-zero probability under the trajectories generated by ${\pi}_{1:h-1}$. Consider any other policies ${\pi}^{\prime}_{1:h-1}$, which also give a non-zero probability to $c_h$. Then, we assume that: for any {such} $c_h\in\cC_h$, {and any} $p_{h}\in\cP_h, s_h\in \cS$, 
    $\PP^{{\pi}_{1:h-1}, \cG}_{h}\left(s_h, p_{h}\given c_h\right) = \PP^{{\pi}^{\prime}_{1:h-1}, \cG}_{h}\left(s_h, p_{h}\given c_h\right).$
\end{assumption}  

This assumption has been made  in the literature \citep{nayyar2013common, gupta2014common}, which is  related to the notion of \emph{one-way separation} in stochastic control, that is, the estimation (of the state in standard stochastic control and of the state and private information in our case)  in Assumption  \ref{str_indi} is \emph{independent} of the control strategy. 
 For more detailed discussions, we refer to \cite{nayyar2013common}.
 Before proceeding with further analysis, {we introduced some common examples in \Cref{sec:example} that  satisfy this  assumption (see \cite{nayyar2013common} and also \Cref{subsec:finite_memory}).

With Assumption \ref{str_indi}, we are able to develop the planning algorithm (summarized in Algorithm \ref{alg:vi}) with the following time complexity.
The algorithm is based on \emph{value iteration} on the common information space, which runs in a backward way, enumerating all possible $c_h$ at each step $h$ and computing the corresponding equilibrium in the prescription space. 
Note that 
a value-iteration algorithm for \emph{NE  computation} 
was firstly also studied in \cite{nayyar2013common}, over the space of \emph{common-information-based beliefs} (instead of that of common information). By planning over the common-information space, we can establish its computational complexity, which was not established in \cite{nayyar2013common}, and enables a more efficient planning algorithm later by truncating the common information properly (cf. \Cref{sec:planning_ais}). 
We now establish the computational  complexity of Algorithm \ref{alg:vi} more concretely.   

\begin{theorem}\label{thm:planning_ind_common_belief}
Fix $\epsilon>0$. For the POSG $\cG$ {that satisfies  Assumptions \ref{evo} and \ref{str_indi}}, given access to the belief $\PP_{h}^\cG(s_h, p_h\given c_h)$,
Algorithm \ref{alg:vi} computes an {$\epsilon$-NE if $\cG$ is zero-sum or cooperative, and an $\epsilon$-CE/CCE if $\cG$ is general-sum,} with time complexity $\max_{h\in[H]}~C_h\cdot \texttt{poly}(S, A, P_h, H, \frac{1}{\epsilon})$.   
\end{theorem}

{To prove this, we will prove a more general theorem (see Theorem \ref{thm:struc} later), of which Theorem \ref{thm:planning_ind_common_belief} is a special case}. This theorem characterizes the dependence of computational  complexity on the cardinality of the common information set and private information set. 
To get a sense of how large $C_hP_h$ could be, we consider one   common scenario  where each agent has \emph{perfect recall}, i.e., she remembers   what she did in prior moves, and also  remembers everything that she knew before. 

\begin{definition}[Perfect recall]\label{def:perfect_recall}
    We say that  agent $i$ has perfect recall if for any $h\in [H]$, it holds that $\{a_{i, 1:h-1}, o_{i, 1:h}\}\subseteq\tau_{i, h}$, and $\tau_{i, h}\subseteq\tau_{i, h+1}$.
\end{definition}

If each agent has perfect recall as defined above,  we can show that $C_hP_h$ must be exponential in the horizon index  $h$. Proof of the result below can be found in \Cref{subsec:hardness_proof}. 

\begin{lemma}\label{lemma:cardi}
Fix any $h\in [H]$, and suppose Assumption \ref{evo} holds.  Then, if each agent has perfect recall as  given in Definition \ref{def:perfect_recall}, then for any information-sharing  structure, we have 
$C_hP_h\ge (OA)^{h-1}$. 
\end{lemma}

From this result, {together with \Cref{thm:planning_ind_common_belief}}, we know that the computational complexity of such a naive planning algorithm must suffer from the exponential dependence of $\Omega((OA)^h)$. {This negative result implies that it is barely possible to get computational efficiency for planning in the true model $\cG$, since the cardinality $C_hP_h$ has to be very large oftentimes. Meanwhile, it is worth noting that for obtaining Theorem \ref{thm:planning_ind_common_belief},  we have not yet leveraged   our Assumption \ref{observa}. Thus, this negative result is in line with our fundamental hardness results in Proposition \ref{posg_hardness_1}. 
 
\section{Planning and Learning with Approximate Common Information}\label{sec:alg}

\subsection{Computationally (quasi-)efficient planning}
\label{sec:planning_ais}

Previous exponential complexity comes from the fact that $C_h$ and $P_h$ could not be made \emph{simultaneously} small {in the standard scenario with perfect recall}. {To address this issue, we propose to further \emph{compress} the information available to the agent under  certain regularity conditions, while approximately maintaining the optimality of the policies computed/learned from the compressed information.  Notably, there is a trade-off between \emph{compression error} and \emph{computational tractability}. We show next that by  properly compressing only the \emph{common information}, we can obtain efficient planning (and learning) algorithms with favorable suboptimality guarantees.}
To introduce  the idea, we first define the \emph{approximate} common information model in our setting.

\begin{definition}[{Approximate common information model}]\label{def:ais}
We define an \emph{expected approximate common information model} of $\cG$ as
\vspace{-2mm}
\[
\cM:=\Big(\{\hat{\cC}_h\}_{h\in[H+1]},\{\hat{\phi}_{h+1}\}_{h\in[H]},\{\PP^{\cM, z}_h\}_{h\in[H]}, \Gamma,\hat{r}^\cM\Big),
\vspace{-1mm}
\]
where $\Gamma=\times_{h\in[H]}\Gamma_h$ is the function class for joint prescriptions, {$\hat{\cC}_h$ is the space of approximate common information at  step $h$,} $\PP^{\cM, z}_h:\hat{\cC}_h\times \Gamma_h\to \Delta(\cZ_{h+1})$  gives the probability of $z_{h+1}$  given $\hat{c}_h\in\hat{\cC}_h$ and $\{\gamma_{i, h}\}_{i\in [n]}\in \Gamma_{h}$, with $\cZ_{h+1}$ being the space of incremental  common information. Similarly, for $\hat{r}^{\cM}=\{\hat{r}_{i, h}^{\cM}\}_{i\in[n],h\in[H]}$, $\hat{r}_{i, h}^{\cM}:\hat{\cC}_h\times \Gamma_h\to [0, 1]$ gives the reward of the $i^{th}$ agent at step $h$ given $\hat{c}_h\in\hat{\cC}_h$ and $\{\gamma_{i, h}\}_{i\in [n]}\in \Gamma_{h}$. 
We denote $\hat{C}_h:=|\hat{\cC}_h|$ for any $h\in [H+1]$. We say $\cM$ is an $(\epsilon_r(\cM), \epsilon_z(\cM))$-\emph{expected-approximate common information model} of $\cG$ with the \emph{approximate common information} defined by $\{\hat{c}_{h}\}_{h\in [H+1]}$ for some compression functions  $\{\operatorname{Compress}_{h}\}_{h\in[H+1]}$  that yield $\hat{c}_h = \operatorname{Compress}_{h}(c_h)$, if it satisfies  the {following}: 

\begin{itemize}
    \item It evolves in a recursive manner, i.e., for each $h\in[H]$, there exists a transformation function $\hat{\phi}_{h+1}$ such that  
    \begin{equation}
        \label{def:ais_evo}\hat{c}_{h+1} = \hat{\phi}_{h+1}(\hat{c}_{h}, z_{h+1}),
    \end{equation}
    {where we recall that $z_{h+1} = c_{h+1}\setminus c_h$ is the common information increment.}
    \item It suffices for approximately evaluating the performance, i.e., for {any $i\in[n]$ and $h\in[H]$,} any prescription ${\gamma}_h\in\Gamma_h$ and joint policy $\pi^\prime\in \Pi^{\mathrm{det}}$, it holds that
    \begin{equation}\label{def:ais_2}
       \EE_{a_{1:h-1}, o_{1:h}\sim \pi^\prime}^{\cG}\Big|\EE^\cG[r_{i, h}(s_h, a_h)\mid c_h, {\gamma}_h]- 
        \hat{r}_{i, h}^{\cM}(\hat{c}_h, \gamma_h)\Big|\le \epsilon_r(\cM).
    \end{equation}
    \item It suffices for approximately predicting   common information increment: for any $h\in[H]$, ${\gamma}_h\in\Gamma_h$, $\pi^\prime\in \Pi^{\mathrm{det}}$, and for $\PP_h^\cG(z_{h+1}\given c_h, {\gamma}_h)$ and $\PP_h^{\cM, z}(z_{h+1}\given\hat{c}_h, {\gamma}_h)$, we have
    {
    \begin{equation}\label{def:ais_3}
    \EE_{a_{1:h-1}, o_{1:h}\sim\pi^\prime}^\cG\big\|\PP_h^\cG(\cdot\given c_h, {\gamma}_h) - \PP_h^{\cM, z}(\cdot\given\hat{c}_h, {\gamma}_h)\big\|_1\le \epsilon_z(\cM).
    \end{equation}}
\end{itemize}

\end{definition}
\begin{remark}
	The approximate model $\cM$ defined above can be treated as a \textit{(fully-observable) stochastic game}, where the state space is $\{\hat{\cC}_h\}_{h\in [H+1]}$, $\Gamma$ is the joint action space, 
	{the composition of $\{\PP_{h}^{\cM, z}\}_{h\in [H]}$ and $\{\hat{\phi}_{h+1}\}_{h\in [H]}$ yields the state transition kernel,}   
	and $\hat{r}_{i, h}^{\cM}(\hat{c}_h, \gamma_h)$ is the reward of the $i^{th}$ agent at step $h$ given \emph{state} $\hat{c}_h$ and \emph{joint action} $\gamma_h$.  
\end{remark}

\begin{remark}\label{remark:compare_kao_mao}
Note that related definitions in    \cite{kao2022common,mao2020information,subramanian2022approximate} required the \emph{total variation distance}  between $\PP_h^{\cG}(\cdot\given c_h, {\gamma}_h)$ and $\PP_h^{\cM, z}(\cdot\given\hat{c}_h, {\gamma}_h)$ to be \emph{uniformly} bounded for \emph{all} $c_h$. In fact, this kind of compression may be unnecessary and computationally intractable when it comes to efficient planning. Firstly, some common information $c_h$ may have very low visitation frequency under \emph{any policy} $\pi$, which means that we can allow large variation between true common belief and approximate common belief for these $c_h${, which are inherently less important for the decision-making problem}.
Secondly, even in the single-agent setting, where $c_h=\{a_{1:h-1}, o_{1:h}\}$, the size of such approximate information with errors uniformly bounded for \emph{all} {$\{a_{1:h-1}, o_{1:h}\}$} {may} not be sub-exponential even {under Assumption \ref{observa}},  as shown by Example B.2 in \cite{golowich2022planning}. Therefore, for some kinds of common information, it is actually \emph{not possible} to reduce the order of complexity through the approximate common belief with errors uniformly bounded. 
\end{remark}

 Although we have characterized what conditions the expected approximate common information model $\cM$ should satisfy to well approximate the underlying $\cG$, it is in general unclear how to \emph{construct} such an $\cM$, i.e., mainly how to define $(\{\PP^{\cM, z}_h\}_{h\in[H]}, \hat{r}^\cM)$, even if we are already given certain compression functions. To address this,  in the following, we provide a way to construct $(\{\PP^{\cM, z}_h\}_{h\in[H]}, \hat{r}^\cM)$ from an approximate belief over the  state and the private information $\{\PP_{h}^{\cM, c}(s_h, p_h\given \hat{c}_h)\}_{h\in [H]}$. 
 \begin{definition}[Model-belief consistency]\label{def:consistency}
    We say the {expected approximate} common information model $\cM$ is \emph{consistent with} some belief {$\{\PP_{h}^{\cM, c}(s_h, p_h\given \hat{c}_h)\}_{h\in [H]}$} if it satisfies the following {for all  $i\in [n]$, $h\in [H]$}: 
{\# 
\PP_{h}^{\cM, z}(z_{h+1}\given\hat{c}_h, {\gamma}_h) &= \sum_{ \substack{s_h, p_h, a_h, o_{h+1}:\\ \chi_{h+1}(p_h, a_h, o_{h+1}) = z_{h+1}} }\Big(\PP_h^{\cM, c}(s_h, p_h\given\hat{c}_h) 
\prod_{j=1}^n\gamma_{j, h}(a_{j, h}\given  p_{j, h})\label{consis:t}\times\sum_{s_{h+1}}\TT_{h}(s_{h+1}\given s_h, a_h)\OO_{h+1}(o_{h+1}\given s_{h+1})\Big),\\ 
\hat{r}_{i, h}^{\cM}(\hat{c}_h, \gamma_h)&=\sum_{s_h, p_h, a_h}\PP_h^{\cM, c}(s_h, p_h\given\hat{c}_h) \prod_{j=1}^n\gamma_{j, h}(a_{j, h}\given  p_{j, h})r_{i, h}(s_h, a_h). \label{consis:r}
\#}
\end{definition}
With such an expected approximate common information model, similar to Algorithm \ref{alg:vi}, {we develop a value-iteration-type algorithm (see pseudocode in Algorithm \ref{alg:avi}) running on the model $\cM$ instead of $\cG$, which outputs an approximate NE/CE/CCE, enjoying the following guarantees}. 
{The key benefit of requiring the model $\cM$ to be \emph{consistent} with some belief is that under this condition, the stage game in Algorithm \ref{alg:avi} can be formulated as a \emph{multi-linear} game of \emph{polynomial} size, 
 thus computing its equilibrium is computationally tractable (cf. \Cref{subsec:subroutine}). 
\begin{theorem}\label{thm:struc}
Fix $\epsilon_r, \epsilon_z, \epsilon_e>0$. 
Given any $(\epsilon_r, \epsilon_z)$-expected-approximate common information model $\cM$ for the POSG $\cG$  {under Assumptions \ref{evo} and \ref{str_indi}}. 
Furthermore, if $\cM$ is consistent with some given approximate belief $\{\PP_{h}^{\cM, c}(s_h, p_h\given \hat{c}_h)\}_{h\in [H]}$ (in the sense of \Cref{def:consistency}), then there exists an algorithm, Algorithm \ref{alg:avi}, that can output an $\epsilon$-NE
 if $\cG$ is zero-sum or cooperative, or  $\epsilon$-CE/CCE if $\cG$ is general-sum, where $\epsilon:=2H\epsilon_r + {H^2}\epsilon_z + H\epsilon_e$, 
with time complexity $\max_{h\in[H]}\hat{C}_h\cdot\texttt{poly}(S, A, P_h, H, \frac{1}{\epsilon_e})$.
\end{theorem}

{As a sanity check, by choosing the compression function as the identity mapping, \Cref{thm:struc} recovers \Cref{thm:planning_ind_common_belief}. }

\paragraph{Planning in observable POSGs {without intractable oracles}.} Theorem \ref{thm:struc} {applies to any expected approximate common information model as given in Definition \ref{def:ais},
by substituting the corresponding $\hat{C}_h$. Note that it does not provide a way to \emph{construct} such expected approximate common information models that ensure the computation complexity in the theorem is \emph{(quasi-)polynomial}. 

Next, we show that in several natural and standard  information structure examples, a simple \emph{finite-memory} compression 
can attain the goal of computing $\epsilon$-NE/CE/CCE without computationally intractable oracles, where we refer to \Cref{subsec:finite_memory} for the concrete form of the finite-memory compression. Based on this, we present the corresponding quasi-polynomial time complexities as follows.}
\begin{theorem}\label{thm:plan_cases}
Fix $\epsilon>0$. Under Assumption \ref{observa}, for all the information-sharing structures in \Cref{sec:example}, there exists a quasi-polynomial time {algorithm that can} compute an $\epsilon$-NE {if $\cG$ is  zero-sum or cooperative, and an  $\epsilon$-CE/CCE if $\cG$ is general-sum}.
\end{theorem}

\subsection{Statistically  (quasi-)efficient learning}\label{subsec:subsec-quasi-learn}
Until now, we have been assuming the full knowledge of the model $\cG$ (the transition kernel, emission, and reward functions). In this full-information setting, we are able to construct some model $\cM$ to approximate the true model $\cG$ according to the conditions we identified in Definition \ref{def:ais}. However, when we only have access to the samples drawn from the POSG $\cG$, it is difficult to directly construct such a model $\cM$ due to the lack of the model specification.
 To address this issue, we propose to construct a specific expected approximate common information model that \emph{depends on the policies $\pi^{1:H}$ that generate the data for such a construction, which is denoted by} $\Tilde{\cM}(\pi^{1:H})$.  For such a model, one could \emph{simulate} and \emph{sample} by running policies $\pi^{1:H}$ in the true model $\cG$.  {The choice of $\pi^{1:H}$ will be specified later {to ensure $\tilde{\cM}(\pi^{1:H})$ to be a good approximation of $\cG$}}. 
   
 {Compared to \cite{golowich2022learning}, there are several key technical challenges our analysis needs to address:  firstly, \cite{golowich2022learning} only considered   \emph{finite-memory} approximation for POMDPs, where the sample complexity can be easily characterized by the \emph{length of the finite memory}. In contrast, our goal is to deal with a more general compression scheme such that it can handle different common information structures, for which we need to define a generalized quantity that can characterize the sample complexity under general compression schemes (cf. \Cref{def:L_main}). Secondly,  and more importantly, \cite{golowich2022learning} essentially learned  the transition and reward of the approximate model by simply \emph{enumerating} all possible actions, which corresponds to enumerating all possible prescriptions $\gamma_h\in\Gamma_h$ for each $\hat{c}_h\in\hat{\cC}_h$ to learn $\PP^{\tilde{\cM}(\pi^{1:H}), z}_h(\cdot\given \hat{c}_h, \gamma_h)$ and $\hat{r}_{i, h}^{\tilde{\cM}(\pi^{1:H})}(\hat{c}_h, \gamma_h)$, if one naively applies its algorithm and analyses to our setting. This will lead to an \emph{exponential} sample complexity since even the number of possible \emph{deterministic} prescriptions is $A^{P_h}$ (while all possible \emph{randomized}  prescriptions are even larger and infinitely many). To address this challenge, we identify a \emph{decomposition} on the aforementioned quantities to separately learn the distributions of private information and the next observation. Analyzing such a separate learning procedure requires a careful examination of those $\hat{c}_h\in\hat{\cC}_h$ and $p_h\in\cP_h$ that are rarely visited by $\pi^{h}$.} 
 
 To introduce the aforementioned approximate model  $\Tilde{\cM}(\pi^{1:H})$, we present the following  definition, where the key is to introduce a set of approximate common information-based beliefs  $\{\PP_h^{\pi^h, \cG}(s_h, p_h\given\hat{c}_h)\}_{h\in[H]}$, which is generated by  running a certain policy  $\pi^h\in \Delta(\Pi^{\mathrm{det}})$ under the true model $\cG$.

\begin{definition}[{Policy-dependent approximate common information model}]\label{def:simulation_main} Given a model $\Tilde \cM$ (as in Definition \ref{def:ais}) and $H$ {joint} policies $\pi^{1:H}$, where each $\pi^h\in \Delta(\Pi^{\mathrm{det}})$  for $h\in [H]$, 
 {we say $\Tilde \cM$ is a    \emph{policy-dependent expected approximate {common information} model}}, denoted as  $\Tilde{\cM}(\pi^{1:H})$,  if it is consistent with {the}  \emph{policy-dependent} belief $\{\PP_{h}^{\pi^h, \cG}(s_h, p_h\given \hat{c}_h)\}_{h\in [H]}$ (as per \Cref{def:consistency}). 
\end{definition}

  Now we present the main theorem for learning under an expected approximate common information model $\Tilde{\cM}(\pi^{1:H})$.
A major difference from the analysis for planning in \Cref{sec:planning_ais} is that, we need to \emph{explore} the space of approximate common information, which {is a function of}  a sequence of observations and actions, and we propose to characterize the \emph{length} of the approximate common information as defined below.

\begin{definition}[{Length of approximate common information}]\label{def:L_main}
Given the compression functions $\{\text{Compress}_h\}_{h\in[H+1]}$,  we define the integer $\hat{L}> 0$ as the minimum length such that there exists a mapping $\hat{f}_h:\cA^{\min\{\hat{L}, h\}}\times\cO^{\min\{\hat{L}, h\}}\rightarrow\hat{\cC}_h$ such that for each $h\in [H+1]$ and joint history $\{o_{1:h}, a_{1:h-1}\}$, we have $\hat{f}_h(x_h)=\hat{c}_h$, where $x_{h}=\{a_{\max\{h-\hat{L}, 1\}}, o_{\max\{h-\hat{L}, 1\} + 1}, \cdots, a_{h-1}, o_{h}\}$. 
\end{definition}

Such an  $\hat{L}$ will help characterize our final sample complexity, since we need to do exploration for the steps after $h-\hat{L}$, and $\hat{L}$ characterizes  the cardinality of the space to be explored. 
 With this definition of $\hat{L}$, we develop Algorithm \ref{alg:plam}, {which learns the model $\Tilde{\cM}(\pi^{1:H})$, i.e., mainly learning the two quantities $\PP^{\Tilde{\cM}(\pi^{1:H}), z}$ and $\hat{r}^{\Tilde{\cM}(\pi^{1:H})}$,  
 by executing policies $\pi^{1:H}$  
 in the true model $\cG$}, with the following sample complexity.
 
\begin{theorem}\label{thm:main_thm_learning}
    Suppose the POSG $\cG$ satisfies Assumptions \ref{evo} and  \ref{str_indi}. 
    Given $H$ policies $\pi^{1:H}$, $\tilde{\cM}(\pi^{1:H})$, and $\hat{L}$ as in \Cref{def:L_main}, where each $\pi^h\in \Delta(\Pi^{\mathrm{det}})$,  $\pi^h_{h-\hat{L}:h}= \operatorname{Unif}(\cA)$ 
    for $h\in [H]$.
    Fix the  parameters $\delta_1, \theta_1, \theta_2, \zeta_1, \zeta_2, \epsilon_e>0$ 
     for Algorithm \ref{alg:plam}, and some $\phi>0$, define the approximation error for estimating $\Tilde{\cM}(\pi^{1:H})$ using samples   
     under the policies $\pi^{1:H}$ as     $
     \epsilon_{apx}(\pi^{1:H}, \hat{L}, \zeta_1, \zeta_2, \theta_1, \theta_2, \phi)$. 
Then, Algorithm \ref{alg:plam}, can learn an $\epsilon$-NE if $\cG$ is zero-sum or cooperative,  and an $\epsilon$-CE/CCE if $\cG$ is general-sum, with probability at least $1-\delta_1$, with a sample complexity $N_0= \texttt{poly}(\max_{h\in [H]} P_{h}, \max_{h\in [H]} \hat{C}_{h}, H, A, O, \frac{1}{\zeta_1}, \frac{1}{\zeta_2}, \frac{1}{\theta_1}, \frac{1}{\theta_2})\cdot\log\frac{1}{\delta_1}$, where 
 $
 \epsilon:=H\epsilon_r(\Tilde{\cM}(\pi^{1:H})) +  H^2\epsilon_z(\Tilde{\cM}(\pi^{1:H})) + (H^2+H)\epsilon_{apx}(\pi^{1:H}, \hat{L}, \zeta_1, \zeta_2, \theta_1, \theta_2, \phi) + H\epsilon_e. 
 $
\end{theorem}

A detailed version of the theorem  is in \Cref{subsec:learn}. This meta-theorem establishes a sample complexity guarantee of learning expected approximate common information model $\Tilde{\cM}(\pi^{1:H})$ 
in an online exploration setting, which holds for \emph{any} compression functions and policies $\pi^{1:H}$, whose choices are specified next.  

\paragraph{Sample (quasi-)efficient learning in POSGs {without intractable oracles}.} {Now we apply this  meta-theorem,  and  obtain  quasi-polynomial time and sample complexities for learning the $\epsilon$-NE/CE/CCE, for  several standard information structures.}   

\begin{theorem}\label{thm:learning_example_short}

Under Assumption \ref{observa}, for all the information-sharing structures in \Cref{sec:example}, there exists a multi-agent RL algorithm that 
learns an $\epsilon$-NE if $\cG$ is zero-sum or cooperative, and an $\epsilon$-CE/CCE if $\cG$ is general-sum, with probability at least $1-\delta$,  with both quasi-polynomial time and sample complexities {$(AO)^{C\gamma^{-4}\log\frac{SHO}{\gamma\epsilon}}\log\frac{1}{\delta}$ for some universal constant $C>0$.}\footnote{{Note that throughout the paper, we regard the delay $d$ in  the examples in \Cref{sec:example} as a \emph{constant}. In fact, as shown in the full version of the theorem in \Cref{subsec:learn}, $d$ is allowed to grow \emph{logarithmically} with the horizon $H$ without changing the order of the computational or sample complexities.}} 
\end{theorem}

Due to space constraints, a detailed version of the theorem is presented in \Cref{subsec:learn}, with proof provided in \Cref{sec:proof_learning_ais}. Note that our algorithm is computationally {(quasi-)}efficient, in contrast to the only existing \emph{sample-efficient} MARL algorithm for POSGs in \cite{liu2022sample}, which relied  on computationally intractable oracles.

\section{Finding Team-Optimum  in Dec-POMDPs}
\label{dec-pomdp}

Until now, we have primarily focused  on solving \emph{equilibria} in  POSGs. One notable subclass of POSGs are the Dec-POMDPs, for which a stronger (than equilibrium) solution concept of \emph{team-optimum} (cf. \Cref{def:team}) is usually preferred. 
Our algorithmic framework developed in \Cref{sec:planning_ais} for planning can be readily extended  to computing the team optimal solution, where the only modification is to replace the 
\emph{equilibrium-computation}  subroutine at each step $h$ over the prescription space in  \Cref{alg:avi}  by a \emph{joint-maximization} one  over the prescription space. Specifically, we only need to replace the line 9 of \Cref{alg:avi} by its line 11, i.e., the following step:
\begin{equation}\label{eq:team-Q}
	\left\{\pi_{1, h}(\cdot\given \hat{c}_h, \cdot), \cdots, \pi_{n, h}(\cdot\given \hat{c}_h, \cdot)\right\}\leftarrow\arg\max_{\gamma_{1, h}, \cdots, \gamma_{n, h}} Q_{h}^{\star, \cM}(\hat{c}_h, \gamma_{1, h}, \cdots, \gamma_{n, h}),
\end{equation} 
where we omit the agent index for the  $Q$-function, since  
all the agents share the same $Q$-function for the Dec-POMDP setting. 

 Unfortunately, although we can show in \Cref{prop:linear} that such a $Q$-value is \emph{linear} w.r.t. each $\gamma_{i, h}$, it is not necessarily \emph{concave} w.r.t. $\{\gamma_{1, h}, \cdots, \gamma_{n, h}\}$ \emph{jointly}. Thus, implementing this maximization subroutine can be computationally intractable. In fact, it is an \texttt{NP-hard} problem without additional assumptions.  
\begin{proposition}\label{prop:static-hard}
	Without additional assumptions, even with $n=2$ agents,  solving \Cref{eq:team-Q} is \texttt{NP-hard}. 
\end{proposition}

{Proof of \Cref{prop:static-hard} is deferred to \Cref{supp-sec-5}.} Hence, it seems hopeless to solve \Cref{eq:team-Q} efficiently. Fortunately, many  Dec-POMDPs in real-world applications enjoy certain structures that can be exploited  for efficient computation. Specifically, we identify several (sets of) assumptions below, under which solving \Cref{eq:team-Q} can be  computationally tractable. Note that since we need to do planning in the approximate model $\cM$, which is oftentimes constructed based on the \emph{original model} $\cG$ and some \emph{approximate belief} $\{\PP_{h}^{\cM, c}(s_h, p_h\given \hat{c}_h)\}_{h\in[H]}$, we will necessarily need assumptions on these two quantities, 
for which we refer to as the \textbf{Part (1)} and  \textbf{Part (2)} of the assumptions  below,  respectively.

\paragraph{\textbf{Condition 1: Turn-based structures.}} \textbf{Part (1).} For $\cG$, we assume that at each step $h$, there is only one agent, denoted as $\cth\in[n]$ that can affect the state transition. Hence, the transition dynamics take the forms of $\TT_h:\cS\times\cA_{\cth}\rightarrow \cS$. Meanwhile, since only agent $\cth$ can affect the transition, we assume the increment of the common information $z_{h+1}$ in Assumption \ref{evo} is only a function of $(p_h, a_{\cth, h}, o_{h+1})$, i.e., $z_{h+1}=\chi_{h+1}(p_h, a_{\cth, h}, o_{h+1})$ instead of $\chi_{h+1}(p_h, a_{h}, o_{h+1})$. In other words, since at step $h$, agents other than $\cth$ do not affect the transition, we assume their actions are not shared. \kzedit{For the reward, we additionally assume that the reward function has an additive structure, i.e., $r_h(s_h, a_h)=\sum_{j\in[n]}r_{j, h}(s_h, a_{j, h})$ for some functions $\{r_{j, h}\}_{j\in[n]}$.} \textbf{Part (2).} For the approximate belief, we do not impose any assumption. {Note that such turn-based structures have been   common in the (fully-observable) stochastic game settings \citep{filar2012competitive,bai2020provable}.}

\paragraph{\textbf{Condition 2: Nested information-sharing.}} \textbf{Part (1).} For $\cG$, we do not impose any assumption. \textbf{Part (2).} For the approximate belief, we assume that all the agents form a \emph{hierarchy} according to the private information they possess. Without loss of generality,  we assume for each $i, j\in[n]$ such that for $j< i$, it holds that $p_{j, h}=Y_{h}^{ij}(p_{i, h})$ for some deterministic function $Y_h^{ij}$. More formally, the approximate belief satisfies that $\PP_h^{\cM, c}(p_{j, h}=Y_{h}^{ij}(p_{i, h})\given p_{i, h}, \hat{c}_h)=1$, where $\PP_h^{\cM, c}(p_{j, h}\given p_{i, h}, \hat{c}_h)$ is the \emph{posterior} distribution induced by the joint distribution $\PP_h^{\cM, c}(s_h, p_h\given \hat{c}_h)$. In other words, the $\sigma$-algebra generated by the private information of the $i^{th}$ agent  includes that of the $j^{th}$ agent. {This structure has also been studied in \cite{peralez2024solving} with a heuristic search approach.}

\paragraph{\textbf{Condition 3: Factorized structures.}} \textbf{Part (1).} For $\cG$, we assume that the state $s_h$ at each step $h\in[H]$ can be partitioned into $n$ local states, i.e., $s_h=(s_{1, h}, s_{2, h}, \cdots, s_{n, h})$. Meanwhile, the transition kernel takes the  product form of $\TT_h(s_{h+1}\given s_h, a_h)=\prod_{i=1}^{n}\TT_{i, h}(s_{i, h+1}\given s_{i, h}, a_{i, h})$, the emission also takes the product form of $\OO_{h}(o_h\given s_h)=\prod_{i=1}^n\OO_{i, h}(o_{i, h}\given s_{i, h})$, and the reward function can be decoupled into $n$ terms such that $r_{h}(s_{h}, a_h)=\sum_{i\in[n]}r_{i, h}(s_{i, h}, a_{i, h})$. \textbf{Part (2).} For the approximate belief, we assume the approximate common information and  {its increment} can be factorized so that $\hat{c}_h=(\hat{c}_{1, h}, \cdots, \hat{c}_{n, h})$,  {$z_{i, h+1}=(z_{1, h+1}, \cdots, z_{n, h+1})$}, and their evolutions additionally satisfy that $\hat{c}_{i, h+1}=\hat{\phi}_{i, h+1}(\hat{c}_{i, h}, z_{i, h+1})$,  {$z_{i, h+1} = \chi_{i, h+1}(p_{i, h}, a_{i, h}, o_{i, h+1})$} for some functions $\hat{\phi}_{i, h+1}$ and {$\chi_{i, h+1}$}.
 Correspondingly, the approximate belief needs to satisfy that  
$
\PP^{\cM, c}_h(s_h, p_h\given \hat{c}_h)=\prod_{i=1}^n \PP^{\cM, c}_{i, h}(s_{i, h}, p_{i, h}\given \hat{c}_{i, h}),
$
for some functions $\{\PP_{i, h}^{\cM, c}\}_{i\in[n], h\in[H]}$.

{Under each of these conditions, \Cref{eq:team-Q} can be solved exactly with time complexity $\texttt{poly}(S, A, P_h)$. The key insight into why these conditions suffice is that, they make solving the \emph{joint maximization} in \Cref{eq:team-Q} equivalent to either solving \emph{individual maximization} for each agent,  or \emph{sequential maximization} across agents that can be solved via dynamic programming. Formal statements can be found in  \Cref{lem:single-controller}, \Cref{nested}, and \Cref{factor} for each condition, respectively.} The key insight into why these conditions suffice is that, they make solving the \emph{joint maximization} in \Cref{eq:team-Q} equivalent to either solving \emph{individual maximization} for each agent,  or \emph{sequential maximization} across agents that can be solved via dynamic programming.
Once \Cref{eq:team-Q} can be solved computationally efficiently, computation of the team optimum of Dec-POMDPs becomes tractable, under the same algorithmic framework as \Cref{sec:planning_ais}.
\begin{theorem}\label{thm:plan_cases_dec}
Fix $\epsilon>0$, {and consider a Dec-POMDP $\cG$ satisfying} 
Assumption \ref{observa}, {then} all the examples in \Cref{sec:example} except the one-step delayed sharing case satisfy {either \textbf{Condition 1} or \textbf{Condition 2}} in \Cref{dec-pomdp}. {Hence,}  \Cref{eq:team-Q} can be solved in time complexity $\texttt{poly}(S, A, P_h)$ {for these cases}.
{Correspondingly,} there exists a quasi-polynomial time algorithm that can compute an $\epsilon$-team optimal policy {of $\cG$}. For the one-step delayed sharing {case}, if one additionally assumes that the Dec-POMDP $\cG$ satisfies {\bf Part (1)} of \textbf{Condition 3}, then there also exists a quasi-polynomial time algorithm that can compute an $\epsilon$-team optimal policy of $\cG$,  and moreover,  the time complexity is polynomial (instead of exponential) in  the number of agents $n$. 
\end{theorem}

\paragraph{Extension  to learning settings without model knowledge.} With the planning oracle for Dec-POMDPs developed above, our framework of \emph{learning} in POSGs can be readily extended to learning in Dec-POMDPs accordingly, achieving \emph{both} quasi-polynomial time and sample complexities for learning the approximate \emph{team-optimal} policy. Due to space constraints, we defer the detailed results to \Cref{supp-sec-5}.

\section{Technical Details} 

In this section, we present the proofs for the main results introduced before. More details can be found in the Appendices.

\subsection{Proof of \Cref{thm:struc}} For notational simplicity, we present the main proofs for the NE/CCE case, and the CE case can be derived similarly.
 
\paragraph{Step 1: Evaluating the equilibrium gap of $\hat{\pi}^\star$ under $\cM$.} As we mentioned in \Cref{sec:planning_ais}, \Cref{alg:avi} essentially performs value iteration on $\cM$, where the output policy $\hat{\pi}^\star$ enjoys the standard guarantee of value iteration for (fully-observable) stochastic games at each step $h\in[H]$, with the  \emph{state} being $\hat{c}_h\in\hat{\cC}_h$ and \emph{action} being $\gamma_h\in\Gamma_h$. The formal result is stated as follows.

\begin{lemma}\label{lemma:ne_err_brief}
Fix the input $\cM$ and $\epsilon_e>0$ for \Cref{alg:avi}. The output of the algorithm, i.e., $\hat{\pi}^\star$,  satisfies that for any {$h\in [H+1]$, $c_h\in\cC_h$}, and $\pi_i\in \Pi_i$, $
    {V}^{\pi_i\times\hat{\pi}_{-i}^{\star}, \cM   }_{i, h}(c_h)\le {V}_{i, h}^{\hat{\pi}^\star, \cM   }({c}_h) + (H+1-h)\epsilon_e.
$
\end{lemma}

\paragraph{Step 2: Bounding the value difference between $\cG$ and $\cM$.} Since what we care about in the end is the equilibrium gap in the actual game  $\cG$, we first bound the value difference between $\cM$ and $\cG$ in terms of $\epsilon_z(\cM)$ and $\epsilon_r(\cM)$. 
\begin{lemma}\label{lemma:v_diff}
For any given policy  $\pi^{\prime}\in\Delta(\Pi^{\mathrm{det}})$, $\pi\in\Pi$, and $h\in [H+1]$, we have
$
    \EE_{\pi^{\prime}}^\cG\Big[\Big|V^{\pi,   \cG }_{i, h}(c_h)-{V}_{i, h}^{\pi, \cM   }({c}_h)\Big|\Big]\le (H-h+1)\epsilon_r+ \frac{(H-h+1)(H-h)}{2}\epsilon_z
$.
\end{lemma}
Note that this lemma holds for any $\pi\in\Pi$, thus also $\hat{\pi}^\star$ and its unilaterally deviated policy $\pi_i\times\hat{\pi}^\star_{-i}$, facilitating the following steps. 
\paragraph{Step 3: Evaluating the equilibrium gap of $\hat{\pi}^\star$ under $\cG$.} Now we are ready to evaluate $\hat{\pi}^\star$, the output of \Cref{alg:avi} in $\cG$. We define for each agent $i\in[n]$ the \emph{best response} as
$
    \pi_{i}^{\star} \in \arg\max_{\pi_{i}\in \Pi_i}
    V^{\pi_i\times{\hat{\pi}^{\star}_{-i}}, \cG}_{i, 1}(\emptyset)
$.
Now for any $\pi^\prime\in\Delta(\Pi^{\text{det}})$: 
\begin{align*}
    &\EE_{\pi^{\prime}}^\cG\Big[V^{\pi_i^{\star}\times\hat{\pi}_{-i}^{\star},  \cG }_{i, h}(c_h)-V^{\hat{\pi}^\star,  \cG }_{i, h}(c_h)\Big]\\
    &=\EE_{\pi^{\prime}}^\cG\left[\left(V^{\pi_i^{\star}\times \hat{\pi}_{-i}^{\star},  \cG }_{i, h}(c_h)-{V}_{i, h}^{\hat{\pi}^\star, \cM   }({c}_h)\right)
    +\left({V}_{i, h}^{\hat{\pi}^\star, \cM   }({c}_h) -V^{\hat{\pi}^\star,  \cG }_{i, h}(c_h)\right)\right]\\
    &\le\EE_{\pi^{\prime}}^\cG\left[\left(V^{\pi_i^{\star}\times\hat{\pi}_{-i}^{\star},  \cG }_{i, h}(c_h)-{V}_{i, h}^{\pi_i^\star\times\hat{\pi}_{-i}^\star, \cM   }({c}_h)\right)
    +\left({V}_{i, h}^{\hat{\pi}^\star, \cM }({c}_h) -V^{\hat{\pi}^\star,  \cG }_{i, h}(c_h)\right)\right] \\
    &\qquad+ (H+1-h)\epsilon_e\\
    &\le 2(H-h+1)\epsilon_r + (H-h)(H-h+1)\epsilon_z + (H-h+1)\epsilon_e,
\end{align*}
where the second step is from \Cref{lemma:ne_err_brief} and the third step is by \Cref{lemma:v_diff}. Letting $h=1$, we conclude that 
$
\operatorname{NE/CCE-gap}(\hat{\pi}^\star)\le 2H\epsilon_r + {H^2}\epsilon_z + H\epsilon_e.
$

\paragraph{Step 4: Analyzing computational complexity.} 
Note that \Cref{alg:avi} is of the \emph{double-loop}  type.  \textbf{(1) For the outer-loop:} it enumerates all $\hat{c}_h\in\hat{\cC}_h$ at each $h\in[H]$. \textbf{(2) For the inner-loop:} the main computation comes from computing the $\epsilon_e$-NE/CE/CCE of the game defined by $\{Q_{i, h}^{\star, \cM}(\hat{c}_h, \cdots)\}_{i\in[n]}$. Note that if we treat this game as a normal-form game with the action space being all the deterministic prescriptions, then any normal-form game solvers can be plugged in. However, the corresponding time complexity will suffer from the size of the action space, i.e., $A_{i}^{P_{i,h}}$. Instead, we show that  if we regard each $\gamma_{i, h}$ as a \emph{concatenation}  of simplexes,  i.e., $\gamma_{i, h}\in\Delta(\cA_i)^{P_{i, h}}$, then $Q_{i, h}^{\star, \cM}$ is linear w.r.t. each individual prescription \emph{under our model-belief consistency condition}.
Thus, an $\epsilon_e$-NE/CE/CCE can be solved with time complexity depending only polynomially on the dimension of $\gamma_{i, h}$, which is $A_{i}P_{i, h}$, in contrast to the previous $A_{i}^{P_{i,h}}$. By putting the time complexity for the outer loop and inner loop together, we obtain the final time complexity. \hfill $\square$    

\subsection{Proof of \Cref{thm:plan_cases}}\label{finite-mem-proof} 
We take the one-step delayed sharing case  as an example, and defer the proofs for other information structure examples to \Cref{subsec:finite_memory}.
\paragraph{Step 1: Bounding $\epsilon_r(\cM)$,$\epsilon_z(\cM)$ with the belief error.} As in \Cref{def:consistency}, one can construct $\cM$ from {some {given}} {compression functions and} approximate beliefs of $\{\PP_{h}^{\cM, c}(s_h, p_h\given \hat{c}_h)\}_{h\in[H]}$. Thus, we can relate the model errors of $\cM$, i.e., $\epsilon_r(\cM)$ and $\epsilon_z(\cM)$, with the error of the approximate belief. Specifically, we show the following. 
\begin{lemma}\label{lemma:bounding_z_r_diff_using_c}
Given any belief $\{\PP_{h}^{\cM, c}(s_h, p_h\given \hat{c}_h)\}_{h\in[H]}$ and an associated consistent (in the sense of \Cref{def:consistency})  expected approximate common information model  $\cM$, 
it holds that for any $h\in [H], c_{h}\in\cC_h, {\gamma}_h\in \Gamma_h$:
\begin{align}
    \Big\|\PP_{h}^{ \cG}(\cdot \given {c}_h, {\gamma}_h)-\PP_{h}^{\cM, z}(\cdot \given\hat{c}_h, {\gamma}_h)\Big\|_1&\le \Big\|\PP_{h}^\cG(\cdot, \cdot\given c_{h}) - \PP_{h}^{\cM, c}(\cdot, \cdot\given\hat{c}_h)\Big\|_1,\\ 
    \label{eq:any_r}\Big|\EE^\cG[r_{i, h}(s_h, a_h)\given c_h, {\gamma}_h]-\hat{r}_{i, h}^\cM(\hat{c}_h, {\gamma}_h)\Big|&\le \Big\|\PP_{h}^ \cG(\cdot, \cdot\given c_{h}) - \PP_{h}^{\cM, c}(\cdot, \cdot\given\hat{c}_h)\Big\|_1, 
\end{align}
where we recall that $\hat{c}_h:=\text{Compress}_h(c_h)$, with $\{\text{Compress}_h\}_{h\in[H+1]}$ from $\cM$. 
\end{lemma}
The proof mainly relies on our construction for $\cM$ in \Cref{def:consistency}, and the details can be found in \Cref{subsec:finite_memory}.
\paragraph{Step 2: Compressing common information using finite-memory truncation.} Now it remains to design the compression functions {$\{\text{Compress}_h\}_{h\in[H+1]}$} and define {the associated}  approximate belief{s} $\PP_{h}^{\cM, c}:\hat{\cC}_h\rightarrow\Delta(\cS\times\cP_h)$ for $h\in[H]$. Specifically, the information structure satisfies $c_h=\{a_{1:h-1}, o_{1:h-1}\}$, $p_{i, h} = \{o_{i, h}\}$, $z_{h+1} = \{o_{h}, a_{h}\}$. More importantly, the ground-truth belief can be computed as 
$\PP_h^{\cG}(s_h, p_h\given {c}_h) = \bb_h(a_{1:h-1}, o_{1:h-1})(s_h)\OO_{h}(o_h\given s_h)$, 
where $\bb_h$ denotes the posterior state distribution given $(a_{1:h-1}, o_{1:h-1})$ (cf. \Cref{def:belief-state}). Fix an integer $L>0$, we construct the compression of $c_h$ as $\hat{c}_h = \{a_{h-L:h-1}, o_{h-L+1:h-1}\}$. The approximate belief can be defined similarly as above as 
\[\PP_h^{\cM, c}(s_h, p_h\given \hat{c}_h) = {\bb}^\prime_h(a_{h-L:h-1}, o_{h-L+1:h-1})(s_h)\OO_{h}(o_h\given s_h),
\] 
where $\bb_h^\prime$ denotes the approximate belief state, where one ignores the history before step $h-L$ and performs the belief update via Bayes rule along the trajectory after $h-L$ from a prior distribution of \emph{uniform distribution on the state} (cf. \Cref{def:belief-state}). Now we are ready to verify that Definition \ref{def:ais} is satisfied.

\begin{itemize}
    \item By definition,   $\{\hat{c}_h\}_{h\in[H]}$ satisfies condition \eqref{def:ais_evo}. 
    \item For any $ c_h\in \cC_h$ {and the corresponding $\hat{c}_h$ constructed above}:
    \small
    \$
        &\Big\|\PP_{h}^ \cG(\cdot, \cdot\given c_h) - \PP_{h}^{\cM, c}(\cdot, \cdot\given\hat{c}_h)\Big\|_1 \\
        &=\sum_{s_h, o_{h}}\Big|\bb_h(a_{1:h-1}, o_{1:h-1})(s_h)\OO_{h}(o_h\given s_h)-\bb_h^{\prime}(a_{h-L:h-1}, o_{h-L+1:h-1})(s_h)\OO_{h}(o_h\given s_h)\Big|\\
        &=\Big\|\bb_h(a_{1:h-1}, o_{1:h-1})-\bb_h^{\prime}(a_{h-L:h-1}, o_{h-L+1:h-1})\Big\|_1.
    \$
    \normalsize
    To analyze such an error of using finite-memory-based approximate belief, we rely on the result from \cite{golowich2022planning} that belief update is a contraction under Assumption \ref{observa} so that it \emph{forgets} the misspecified prior (i.e., $\text{Unif}(\cS)$ in $\bb_h^{\prime}(a_{h-L:h-1}, o_{h-L+1:h-1})$) at an exponential rate. Formally, we have the following.  
    \begin{lemma}\label{contraction}
    Suppose that the POSG  satisfies Assumption \ref{observa} with parameter $\gamma$. Let $\epsilon\ge 0$. Fix a policy $\pi^\prime\in\Delta(\Pi^{\mathrm{det}})$ and indices $1\le h-L<h-1\le H$. If $L\ge C\gamma^{-4}\log(\frac{S}{\epsilon})$ for some large enough constant $C$, then the following  holds
\begin{align} 
\EE^\cG_{a_{1:h-1}, o_{1:h}\sim \pi^\prime}\| \bb_{h}(a_{1:h-1}, o_{1:h-1}) - \bb_{h}^{\prime}(a_{h-L:h-1}, o_{h-L+1:h-1})\|_1&\le \epsilon\nonumber.
\end{align}
\end{lemma}

We defer its complete version and corresponding proof details to \Cref{noah_theo}. Therefore, combining \Cref{lemma:bounding_z_r_diff_using_c} and \Cref{contraction}, conditions  \eqref{def:ais_2}, \eqref{def:ais_3} in Definition \ref{def:ais} are satisfied with $\epsilon_r=\epsilon_z=\epsilon$. 
\end{itemize}
Finally, together with \Cref{thm:struc}, by choosing $L=\cO\Big(\gamma^{-4}\log(\frac{SH}{\epsilon})\Big)$, $\epsilon_e=\cO(\epsilon/H)$, we proved that $\hat{\pi}^\star$ is an $\epsilon$-NE/CCE. Meanwhile, it is direct to see that $\hat{C}_h\le(AO)^{L}$ and $P_h\le O$, thus proving the  quasi-polynomial time complexity via \Cref{thm:struc}. \hfill $\square$

\subsection{Proof  of \Cref{thm:main_thm_learning}}\label{proof-learning}
\paragraph{Step 1: Decomposing transitions of $\tilde{\cM}(\pi^{1:H})$.}
Learning $\PP_h^{\Tilde{\cM}(\pi^{1:H}), z}(z_{h+1}\given \hat{c}_h, {\gamma}_h)$ for the model $\Tilde{\cM}(\pi^{1:H})$ is   {equivalent} to {learning} $\PP_h^{\pi^h_{1:h-1},  \cG}(z_{h+1}\given \hat{c}_h, {\gamma}_h)${, given the definition of $\Tilde{\cM}(\pi^{1:H})$ in \Cref{def:simulation_main}.}
As highlighted before, learning $\PP_h^{\Tilde{\cM}(\pi^{1:H}), z}(z_{h+1}\given \hat{c}_h, {\gamma}_h)$ by enumerating all $\hat{c}_h$ and ${\gamma}_h$ is \emph{not}  statistically efficient if naively following that of \cite{golowich2022learning}. To circumvent this issue, we notice
\[\PP_{h}^{\pi^{h}_{1:h-1},  \cG}(z_{h+1}\given\hat{c}_h, {\gamma}_h) = \sum_{\substack{p_h, a_h, o_{h+1}:\\\chi_{h+1}(p_h, a_h, o_{h+1})=z_{h+1}}}\PP_{h}^{\pi^{h}_{1:h-1},  \cG}(p_h, a_h, o_{h+1}\given \hat{c}_h, {\gamma}_h),\]
where we recall $\chi_{h+1}$ from Assumption \ref{evo}. Now, we notice the decomposition:
\[
\PP_{h}^{\pi^{h}_{1:h-1},  \cG}(p_h, a_h, o_{h+1}\given \hat{c}_h, {\gamma}_h) = \PP_{h}^{\pi^h_{1:h-1},  \cG}(p_h\given \hat{c}_h)\gamma_h(a_h\given p_h)\PP_{h}^{\pi^h_{1:h-1},  \cG}(o_{h+1}\given \hat{c}_h, p_{h}, a_{h}),
\]
where we use the shorthand notation $\gamma_h(a_h\given p_h):=\prod_{i=1}^n\gamma_{i, h}(a_{i, h}\given p_{i, h})$, and note that $\PP_{h}^{\pi^h_{1:h-1},  \cG}(o_{h+1}\given \hat{c}_h, p_{h}, a_{h})$ does not depend on $\gamma_h$ anymore. With such a decomposition, it suffices to learn $\PP_{h}^{\pi^{h}_{1:h-1},  \cG}(p_h\given \hat{c}_h)$ and $\PP_{h}^{\pi^h_{1:h-1},  \cG}(o_{h+1}\given \hat{c}_h, p_{h}, a_{h})$.
\paragraph{Step 2: Bounding the statistical error for learning $\PP_{h}^{\pi^{h}_{1:h-1},  \cG}(p_h\given \hat{c}_h)$ and $\PP_{h}^{\pi^h_{1:h-1},  \cG}(o_{h+1}\given \hat{c}_h, p_{h}, a_{h})$.} The accuracy and sample complexity of learning those two conditional probabilities depend on the visitation probability of $\hat{c}_h$ and $p_h$ under $\pi^h$. Therefore, we first handle those $\hat{c}_h$ and $p_h$ with large visitation probability as follows.
\begin{lemma}\label{lemma:estimation-short}
    Fix $\delta_1, \zeta_1,\zeta_2, \theta_1, \theta_2> 0$. Given the compression functions and correspondingly the $\hat{L}$ as per \Cref{def:L_main}, suppose for all $h\in [H]$, $\pi^{h}\in\Delta(\Pi^{\operatorname{det}})$ satisfies that $\pi^{h}_{h-\hat{L}:h}=\text{Unif}(\cA)$, \Cref{alg:construct_mg} with sample complexity $N_0= \texttt{poly}(A, O, $\\$\max_h P_h,\max_h \hat{C}_h, \frac{1}{\zeta_1}, \frac{1}{\zeta_2}, \frac{1}{\theta_1}, \frac{1}{\theta_2}, \log\frac{1}{\delta_1})$ ensures the following holds with probability at least $1 - \delta_1$ for each $h\in[H]$:  
    \begin{itemize}
        \item For $\hat{c}_h\in\hat{\cC}_h$ such that $\PP_h^{\pi^h_{1:h-1}, \cG}(\hat{c}_h)\ge \zeta_1$,  \Cref{alg:construct_mg} can learn \\$\PP_{h}^{\hat{\cM}(\pi^{1:H})}(\cdot\given\hat{c}_h)\in\Delta(\cP_h)$ so that 
        $
        \sum_{p_h}\left|\PP_{h}^{\hat{\cM}(\pi^{1:H})}(p_h\given\hat{c}_h) - \PP_{h}^{\pi^h_{1:h-1},  \cG}(p_h\given\hat{c}_h)\right|\le  \theta_1.
        $
        \item For $(\hat{c}_h, p_h, a_h)\in\hat{\cC}_h\times\cP_h\times\cA$ such that $\PP_h^{\pi^h_{1:h-1}, \cG}(\hat{c}_h, p_h)\ge \zeta_2$, {\Cref{alg:construct_mg} can learn $\PP_{h}^{\hat{\cM}(\pi^{1:H})}(\cdot\given\hat{c}_h, p_h, a_h)\in\Delta(\cO)$ such that}\\ 
        $
        \sum_{o_{h+1}}\left|\PP_{h}^{\hat{\cM}(\pi^{1:H})}(o_{h+1}\given\hat{c}_h, p_h, a_h) - \PP_{h}^{\pi^h_{1:h-1},  \cG}(o_{h+1}\given\hat{c}_h, p_h, a_h)\right|\le \theta_2.
        $

    \end{itemize}
    We refer to the joint of the two bullets above as event $\cE_1$. 
\end{lemma}
\begin{proof}
	We prove the first item, where the second one can be proved similarly. Note that for any trajectory $k$ of Algorithm \ref{alg:construct_mg}, the distribution of $p_h^k$ conditioned on $\hat{c}_h^k$ is exactly $\PP_{h}^{\pi^h_{1:h-1},  \cG}(p_h^k\given\hat{c}_h^k)$.

    Now consider any $\hat{c}_h\in\hat{\cC}_h$ such that $\PP_h^{\pi^h_{1:h-1}, \cG}(\hat{c}_h)\ge \zeta_1$. By the Chernoff bound, with probability at least $1-\exp(-\frac{\zeta_1 N_0}{8})$, there are at least $\frac{\zeta_1 N_0}{2}$ trajectories  {indexed by the set $\cK^1\subseteq[N_0]$}, such that for any $k\in \cK^1$, $\operatorname{Compress}_h(f_h(a_{1:h-1}^k, o_{1:h}^k)) = \hat{c}_h$, where we recall the definition of $f_h$ in \Cref{sec:prelim_model}. By the folklore theorem of learning a discrete probability distribution \citep{canonne2020short}, with probability {at least} $1-p^\prime$, $
        \sum_{p_h}\left|\PP_{h}^{\hat{\cM}(\pi^{1:H})}(p_h\given\hat{c}_h) - \PP_{h}^{\pi^h_{1:h-1},  \cG}(p_h\given\hat{c}_h)\right|\le  \theta_1 
        $ holds as long as
    \begin{equation}\label{eq:learn_dis_brief}
    \frac{\zeta_1 N_0}{2}\ge \frac{C(P_h + \log\frac{1}{p^\prime})}{\theta_1^2},
    \end{equation}
    for some constant $C>1$. By a union bound over all possible $h\in [H]$ and $\hat{c}_h\in \hat{\cC}_h$, the first item holds with probability at least
    $
    1-H\max_h\hat{C}_h\exp(-\frac{\zeta_1 N_0}{8}) - H\max_h\hat{C}_h p^\prime.
    $
    Now set $p^\prime = \frac{\delta_1}{4H\max_h \hat{C}_h}$. It is  direct to verify that \Cref{eq:learn_dis_brief} holds if $N_0\ge \frac{C(\max_h P_h + \log\frac{4H \max_h\hat{C}_h}{\delta_1})}{\zeta_1 \theta_1^2}$. Furthermore, as long as $C$ is sufficiently large, we have that $H\max_h\hat{C}_h\exp(-\frac{\zeta_1 N_0}{8})\le \frac{\delta_1}{4}$. Therefore, we proved that with probability at least $1-\frac{\delta_1}{2}$, the first item holds for all $h\in [H]$  and $\hat{c}_h\in\hat{\cC}_h$ such that $\PP_h^{\pi^h_{1:h-1}, \cG}(\hat{c}_h)\ge \zeta_1$.
\end{proof}

\paragraph{Step 3: Bounding the approximation error of $\hat{\cM}(\pi^{1:H})$ w.r.t. $\tilde{\cM}(\pi^{1:H})$.}  To begin with, with the help of \Cref{lemma:estimation-short}, we are able to handle those $\hat{c}_h\in\hat{\cC}_h$ such that $\PP_h^{\pi^h_{1:h-1}, \cG}(\hat{c}_h)\ge \zeta_1$ as follows.

\begin{lemma}\label{lemma:epsilon_apx-short}
    Given policies $\pi^{1:H}$ such that $\pi^h$ satisfies the same condition as in \Cref{lemma:estimation-short}, under the event $\cE_1$ in Lemma \ref{lemma:estimation-short}, then for any $h\in [H]$, policy $\pi\in\Delta(\Pi^{\mathrm{det}})$, and prescription ${\gamma}_h\in\Gamma_h$, it holds that for any $\hat{c}_h\in\hat{\cC}_h$ with $\PP_h^{\pi^{h}_{1:h-1}}(\hat{c}_h)\ge \zeta_1$
    \[
    \sum_{z_{h+1}}\Big|\PP_{h}^{\Tilde{\cM}(\pi^{1:H}), z}(z_{h+1}\given \hat{c}_h, {\gamma}_h) - \PP_{h}^{\hat{\cM}(\pi^{1:H}), z}(z_{h+1}\given\hat{c}_h, {\gamma}_h)\Big|\\ 
    \le \theta_1 + 2AP_h\frac{\zeta_2}{\zeta_1} + AP_h\theta_2. 
    \]
\end{lemma}
\begin{proof} 
After some algebra, we can bound
\small
\$
&\sum_{p_h, a_h, o_{h+1}}\left|\PP_{h}^{\Tilde{\cM}}(p_h, a_h, o_{h+1}\given \hat{c}_h, {\gamma}_h) - \PP_{h}^{\hat{\cM}}(p_h, a_h,o_{h+1}\given \hat{c}_h, {\gamma}_h)\right| \\
& \le \underbrace{\Big\|\PP_{h}^{\pi^h_{1:h-1},  \cG}(\cdot\given\hat{c}_h) - \PP_{h}^{\hat{\cM}}(\cdot\given\hat{c}_h)\Big\|_1}_{\text{Term I}}  \\
& + \sum_{p_h: \PP_{h}^{\pi^h_{1:h-1},  \cG}(p_h\given\hat{c}_h)\ge \frac{\zeta_2}{\zeta_1}}\PP_{h}^{\pi^h_{1:h-1},  \cG}(p_h\given  \hat{c}_h)\sum_{a_h} \underbrace{\Big\| \PP_{h}^{\pi^h_{1:h-1},  \cG}(\cdot\given\hat{c}_h, p_h, a_h)-\PP_{h}^{\hat{\cM}}(\cdot\given\hat{c}_h, p_h, a_h) \Big\|_1}_{\text{Term II}}\\
& + \sum_{p_h: \PP_{h}^{\pi^h_{1:h-1},  \cG}(p_h\given\hat{c}_h)\le \frac{\zeta_2}{\zeta_1}}\PP_{h}^{\pi^h_{1:h-1},  \cG}(p_h\given  \hat{c}_h)\sum_{a_h} \underbrace{\Big\| \PP_{h}^{\pi^h_{1:h-1},  \cG}(\cdot\given\hat{c}_h, p_h, a_h)-\PP_{h}^{\hat{\cM}}(\cdot\given\hat{c}_h, p_h, a_h) \Big\|_1}_{\text{Term III}},
\$
\normalsize
where under the event $\cE_1$, Term I can be bounded by the first item of \Cref{lemma:estimation-short}. For Term II, since $\PP_{h}^{\pi^h_{1:h-1},  \cG}(p_h\given\hat{c}_h)\ge \frac{\zeta_2}{\zeta_1}$, it implies that $\PP_{h}^{\pi^h_{1:h-1},  \cG}(\hat{c}_h, p_h)\ge \zeta_2$ together with the pre-condition that $\PP_{h}^{\pi^h_{1:h-1},  \cG}(\hat{c}_h)\ge \zeta_1$. This allows us to apply the second item of \Cref{lemma:estimation-short}. For Term III, we directly bound it by $2$. Combining them together, we can conclude
\small
\[
\sum_{p_h, a_h, o_{h+1}}\left|\PP_{h}^{\Tilde{\cM}}(p_h, a_h, o_{h+1}\given \hat{c}_h, {\gamma}_h) - \PP_{h}^{\hat{\cM}}(p_h, a_h,o_{h+1}\given \hat{c}_h, {\gamma}_h)\right| \le \theta_1 + 2AP_h\frac{\zeta_2}{\zeta_1} + AP_h\theta_2.
\]
\normalsize
Noticing that after marginalization, the total variation distance will not increase, we proved our lemma. 
\end{proof}
Until now, we have handled those $\hat{c}_h$ such that $\PP_h^{\pi^{h}_{1:h-1}}(\hat{c}_h)\ge \zeta_1$. For those less visited $\hat{c}_h$, we relate it to certain less-explored \emph{states} at step $h-\hat{L}$, specifically $s_{h-L}$ such that $ \PP_{h-L}^{\pi^h, \cG}(s_{h-L})\le \phi$ as follows. 
\begin{lemma}\label{lemma:low_visitation-short}
    Given compression functions $\{\text{Compress}_h\}_{h\in[H+1]}$ and compute the associated $\hat{L}>0$ as in Definition \ref{def:L_main}.  Fix any $\zeta>0, \phi>0, h\in [H]$. Consider any policies $\pi$, $\pi^\prime\in\Delta(\Pi^{\operatorname{det}})$, such that $\pi^h_{h-\hat{L}:h}=\text{Unif}(\cA)$. Then, we have  
    \[
    \sum_{\hat{c}_h: \PP_h^{\pi^\prime, \cG}(\hat{c}_h)\le \zeta}\PP_h^{\pi, \cG}(\hat{c}_h) \le \frac{A^{2\hat{L}}O^{\hat{L}}\zeta}{\phi} + \bm{1}[h>\hat{L}]\cdot \sum_{s_{h-\hat{L}}: \PP_{h-\hat{L}}^{\pi^\prime, \cG}(s_{h-\hat{L}})\le \phi}\PP_{h-\hat{L}}^{\pi, \cG}(s_{h-\hat{L}}).
    \]
\end{lemma}
This lemma bounds the probability of less-visited $\hat{c}_h$ with that of certain less-visited \emph{state} $s_{h-\hat{L}}$, for which we can leverage 
existing techniques from single-agent RL to minimize by learning a certain \emph{exploratory}  policy $\pi$ later in \Cref{learning-outline}.

Finally, we are ready to evaluate $\epsilon_z(\hat{\cM}(\pi^{1:H}))$. By a triangle inequality, we have
\small
\$
&\epsilon_z\left(\hat{\cM}(\pi^{1:H})\right)
\le \epsilon_z(\tilde{\cM}(\pi^{1:H}))\\
& +\max_{h, \pi\in\Pi^{\text{det}}, \gamma_h\in\Gamma_h}\underbrace{\EE_{\pi}^{ \cG}\bm{1}\Big[\PP_h^{\pi^{h}_{1:h-1}}(\hat{c}_h)\ge \zeta_1\Big]\Big\|\PP_{h}^{\Tilde{\cM}(\pi^{1:H}), z}(\cdot\given \hat{c}_h, {\gamma}_h) - \PP_{h}^{\hat{\cM}(\pi^{1:H}), z}(\cdot\given\hat{c}_h, {\gamma}_h)\Big\|_1}_{\text{Term I}}\\
& + \max_{h, \pi\in\Pi^{\text{det}}, \gamma_h\in\Gamma_h}\underbrace{\EE_{\pi}^{ \cG}\bm{1}\Big[\PP_h^{\pi^{h}_{1:h-1}}(\hat{c}_h)\le \zeta_1\Big]\Big\|\PP_{h}^{\Tilde{\cM}(\pi^{1:H}), z}(\cdot\given \hat{c}_h, {\gamma}_h) - \PP_{h}^{\hat{\cM}(\pi^{1:H}), z}(\cdot\given\hat{c}_h, {\gamma}_h)\Big\|_1}_{\text{Term II}},
\$
\normalsize
where Term I can be bounded by \Cref{lemma:epsilon_apx-short}, and Term II can be bounded by $2\cdot\sum_{\hat{c}_h: \PP_h^{\pi^h_{1:h-1}, \cG}(\hat{c}_h)\le \zeta}\PP_h^{\pi, \cG}(\hat{c}_h)$, which can be further bounded by \Cref{lemma:low_visitation-short}. It is direct to see that Term I and Term II together contribute to the error $\epsilon_{apx}(\pi^{1:H})$ defined in \Cref{thm:main_thm_learning}. $\epsilon_r(\hat{\cM}(\pi^{1:H}))$ can be evaluated similarly.
Now with the help of \Cref{thm:struc}, we proved the optimality in \Cref{thm:main_thm_learning} for planning in $\hat{\cM}(\pi^{1:H})$.
Meanwhile, the sample complexity is $H\times N_0$, thus proving the sample complexity guarantee in \Cref{thm:main_thm_learning}. \hfill $\square$

\subsection{Proof of \Cref{thm:learning_example_short}}\label{learning-outline}
Note that \Cref{thm:main_thm_learning} characterizes the sample complexity for learning an equilibrium for $\cG$ from the model $\tilde{\cM}(\pi^{1:H})$ with approximation errors depending on $\pi^{1:H}$. Therefore, to obtain the final guarantee, one needs to find certain policies $\pi^{1:H}$ to control the corresponding errors in \Cref{thm:main_thm_learning}, i.e., $\epsilon_r(\tilde{\cM}(\pi^{1:H})), \epsilon_z(\tilde{\cM}(\pi^{1:H})), \epsilon_{apx}(\pi^{1:H})$. Note that we have evaluated $\epsilon_{apx}(\pi^{1:H})$ above. For $\epsilon_r(\tilde{\cM}(\pi^{1:H})), \epsilon_z(\tilde{\cM}(\pi^{1:H}))$, similar to the proof for \Cref{thm:plan_cases}, we take the one-step delayed sharing case as an example. 
\paragraph{Step 1: Evaluating $\epsilon_z(\tilde{\cM}(\pi^{1:H}))$ and $\epsilon_r(\tilde{\cM}(\pi^{1:H}))$.} We also use the finite-memory truncation as the compression as before. For any $\pi^{1:H}$, it is direct to verify that  
\small
\[\PP_h^{\Tilde{\cM}(\pi^{1:H}), c}(s_h, p_h\given \hat{c}_h) =\PP_h^{\pi^h,  \cG}(s_h, p_h\given \hat{c}_h) =\Tilde{\bb}_h^{\pi^h}(a_{h-L:h-1}, o_{h-L+1:h-1})(s_h)\OO_{h}(o_h\given s_h),
\]
\normalsize
where $\tilde{\bb}_h^{\pi^h}$ denotes the approximate belief state, where one ignores the history before step $h-L$ and performs the belief update using the Bayes rule after it, from the prior distribution of  $\PP_{h-L}^{\pi^{h}, \cG}(s_{h-L})$ (cf. \Cref{def:belief-state}). If $L\ge C\gamma^{-4}\log (\frac{1}{\epsilon \phi})$, it holds that
\#
\nonumber\epsilon_z&(\tilde{\cM}(\pi^{1:H})) = \max_h\max_{\pi\in\Pi^{\operatorname{det}}, {\gamma}_h} \EE_{\pi}^{ \cG}\left\| \PP_{h}^{\cG}(\cdot\given c_h, {\gamma}_h) - \PP_{h}^{\Tilde{\cM}, z}(\cdot\given \hat{c}_h, {\gamma}_h) \right\|_1\\
\nonumber &\le \max_h\max_{\pi\in\Pi^{\operatorname{det}}, {\gamma}_h}\EE_{\pi}^{{ \cG}}\left\|  \bb_h(a_{1:h-1}, o_{1:h-1})-\Tilde{\bb}_h^{\pi^{h}}(a_{h-L:h-1}, o_{h-L+1:h-1})\right\|_1\\
&\le \epsilon + \max_{h}\max_{\pi\in\Pi^{\operatorname{det}}}\bm{1}[h>L] \cdot 6 \cdot \sum_{s_{h-L}: \PP_{h-L}^{\pi^h, \cG}(s_{h-L})\le \phi}\PP_{h-L}^{\pi, \cG}(s_{h-L}),\nonumber
\# 
where the last step can be proved similarly as \Cref{contraction}. Moreover, $\epsilon_r(\tilde{\cM}(\pi^{1:H}))$ can be evaluated similarly.

\paragraph{Step 2: Minimizing the visitation probability of less-explored states with Barycentric Spanner.} Now we can see that  to control $\epsilon_r(\tilde{\cM}(\pi^{1:H})), \epsilon_z(\tilde{\cM}(\pi^{1:H})), \epsilon_{apx}(\pi^{1:H})$ simultaneously, it suffices to control the quantity, $\sum_{s_{h-L}: \PP_{h-L}^{\pi^h, \cG}(s_{h-L})\le \phi}\PP_{h-L}^{\pi, \cG}(s_{h-L})$. In other words, $\pi^{h}$ should be \emph{exploratory} enough {in the sense that the actual states should be  visited often enough}. It turns out that finding such exploratory policies to minimize this error term can be achieved by the Barycentric-spanner-based techniques \citep{awerbuch2008online}, as also adopted by \cite{golowich2022learning}, using quasi-polynomial sample and computational complexities. By choosing the parameters $\zeta_1$, $\zeta_2$, $\theta_1$, $\theta_2$, and $\phi$ properly, we proved  \Cref{thm:learning_example_short}. \hfill $\square$

\subsection{Proof Outline of \Cref{thm:plan_cases_dec}}

\paragraph{Correctness of the algorithmic framework.} The correctness of our framework follows similarly from the proof of \Cref{thm:struc}. Combining  the fact that $\hat{\pi}^\star$ is an optimal policy of $\cM$ and \Cref{lemma:v_diff}, $\hat{\pi}^\star$ is also an approximate optimal policy of $\cG$.

\paragraph{Computation  analysis.} 
As we mentioned in \Cref{dec-pomdp}, the key of extending our framework to team-optimum-finding in Dec-POMDPs is to implement \Cref{eq:team-Q} in a computationally tractable way {for each $h\in[H]$}. Here we briefly outline how each of the three assumptions can circumvent the hardness in \Cref{prop:static-hard}. \textbf{Condition 1.} In \Cref{lem:single-controller}, we show that, the $Q$-value function \kzedit{can be linearly decomposed into $n$ functions, i.e., $Q_{h}^{\star, \cM}(\hat{c}_h, \gamma_{h})=\sum_{j\in[n]}U_{j, h}(\hat{c}_h, \gamma_{j, h})$, for some functions $\{U_{j, h}\}_{j\in[n]}$, for any $\hat{c}_h\in\hat{\cC}_h$, $\gamma_{h}\in\Gamma_{h}$.} Therefore, \Cref{eq:team-Q} can be solved tractably since \kzedit{each $U_{j, h}$ is indeed a \emph{linear}  function of $\gamma_{j, h}$} with the \emph{concatenation} of simplexes being the constraint.}
\textbf{Condition 2.} With such a nested structure, \Cref{eq:team-Q} can indeed be  solved by a \emph{dynamic programming over the agents}. We consider the following POMDP $\hat{\cP}(n)$ with the horizon length being the number of the agents $n$. The initial state $x_1=(s_h, p_{h})\sim\PP^{\cM, c}_h(s_h, p_{h}\given \hat{c}_h)$. At each step $j\in[n]$ of this POMDP, the observation is $y_j=p_{j, h}$, the $j^{th}$ agent takes the action $a_{j}\in\cA_j$, and the next state transitions to $x_{j+1}=(x_j, a_{j})$. Note that the reward is non-zero only at the last step $n$, where $\hat{r}_n(x_n, a_n)=\EE_{s_{h+1}\sim\TT_h(\cdot\given s_h, a_{1:n}), o_{h+1}\sim\OO_{h+1}(\cdot\given s_{h+1})}[r_{h}(s_h, a_{1:n}) + V_{h+1}^{\star, \cM}(\hat{c}_{h+1})]$. Based on such a  POMDP perspective, we can develop an efficient algorithm for \Cref{eq:team-Q} (cf. \Cref{alg:agent-dp}), where the first for-loop is a standard \emph{backward procedure} of value iteration for the POMDP $\hat{\cP}(n)$ constructed above to compute {its} optimal policy $u_{1:n}^\star$. The second for-loop performs a \emph{forward procedure} of translating $u_{1:n}^\star$ into $\gamma_{1:n, h}^\star$, where $\gamma_{i,h}^\star\in\Gamma_{i, h}$ for each $i\in[n]$ now belongs to the prescription space we hope to optimize over in \Cref{eq:team-Q}. Note that throughout, we regard the number of agents $n$, i.e., the time horizon of $\hat{\cP}(n)$ as a  constant. Hence, the time complexity of such a dynamic programming {for finding the exact optimal policy of} $\hat{\cP}(n)$ is indeed $\texttt{poly}(S, A, P_h)$. 
\textbf{Condition 3.} Due to the factorized structures, in \Cref{factor}, we show that the $Q$-value can be \kzedit{also} decoupled into $n$ terms, such that there exist $n$ functions $\{F_{i, h}\}_{i\in[n]}$ such that $Q_h^{\star, \cM}(\hat{c}_h, \{\gamma_{i, h}\}_{i\in[n]})=\sum_{i\in[n]}F_{i, h}(\hat{c}_{i, h}, \gamma_{i, h})$. Therefore, the maximization over the \emph{joint} $\{\gamma_{i, h}\}_{i\in[n]}$ in  \Cref{eq:team-Q} is equivalent to the \emph{individual}  maximization over each $\gamma_{i, h}$ for $F_{i, h}$, $i\in[n]$, which is again a linear program as we argued before. Thus, \Cref{eq:team-Q} can be also solved with time complexity $\texttt{poly}(S, A, P_h)$. \hfill $\square$

\begin{figure*}[!t]
\centering 
\resizebox{17cm}{!}{
\hspace{-10pt}
\includegraphics[width=0.93\textwidth]{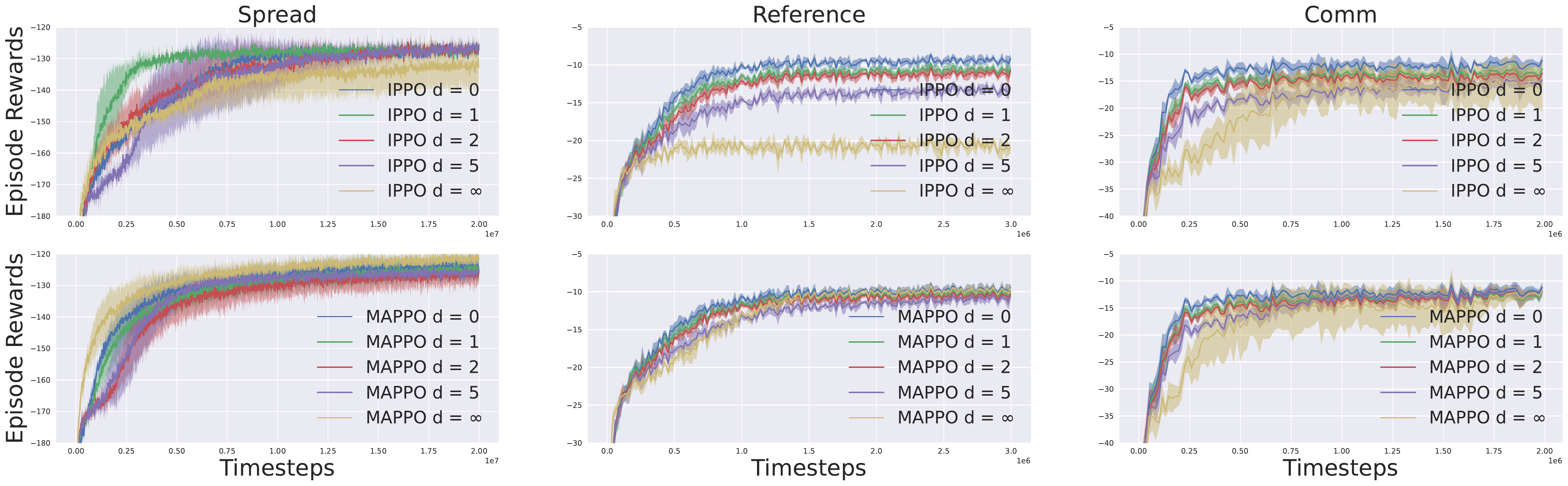}}
\caption{Performance of MAPPO and IPPO in various  delayed-sharing settings.}
\label{fig:mpe}
\end{figure*}

\section{Experimental Results}\label{sec:simulations} 

 For the experiments, we will  both investigate the benefits of \emph{information sharing} as we considered in various empirical MARL environments, and validate the implementability and performance of our proposed approaches on several  odest-scale examples. 

\paragraph{Information sharing improves performance.}  
 We mainly consider three cooperative tasks, the \emph{physical deception} (\emph{Spread}), the \emph{simple reference} (\emph{Reference}), and the \emph{cooperative communication}  (\emph{Comm}) in the popular deep MARL benchmarks, multi-agent particle-world environment (MPE) \citep{lowe2017multi}. 
 We train both the popular centralized-training algorithm MAPPO \citep{yu2021surprising} and the decentralized-training algorithm IPPO \citep{yu2021surprising} with different information-sharing mechanisms by varying the delay from $0$ to $\infty$. 
The rewards during training are shown in Figure \ref{fig:mpe}. It is seen that in all domains (except MAPPO on Spread) with either training paradigms, smaller delays, which correspond to the case of more information sharing,  will lead to faster convergence, higher final performance, and reduced training variance. 

\begin{table}[!t]
\centering
\begin{tabular}{l|l|l|ll|l|l|l}
        & \multicolumn{3}{c}{\textbf{Boxpushing}} &  & \multicolumn{3}{c}{\textbf{Dectiger}} \\
        \hline
Horizon & Ours     & FM-E     & RNN-E    &  & Ours    & FM-E    & RNN-E    \\
\hline
3       & 62.78    & 64.22    & 8.40     &  & 13.06   & -6.0    & -6.0     \\
\hline
4       & 81.44    & 77.80    & 9.10     &  & 20.89   & -4.76   & -7.00    \\
\hline
5       & 98.73    & 96.40    & 21.78    &  & 27.95   & -6.37   & -10.04   \\
\hline
6       & 98.76    & 94.61    & 94.36    &  & 36.03   & -7.99   & -11.90   \\
\hline
7       & 145.35   & 138.44   & 132.70   &  & 37.72   & -7.99   & -13.92 \\
\hline 
\end{tabular}
\caption{Final evaluation of the rewards using our methods, compared with using the methods of FM-E and RNN-E in \cite{mao2020information}.}
\label{table:pa}
\end{table}

\paragraph{Validating implementability and performance.}  
To further validate the tractability of our approaches, we test our learning algorithm on two popular and modest-scale  partially observable benchmarks Dectiger \citep{nair2003taming} and Boxpushing \citep{seuken2012improved}.
We compare our approaches with FM-E and RNN-E, which are also common information-based approaches developed in \cite{mao2020information}. The final rewards are reported in Table \ref{table:pa}. In both domains with various horizons, our methods consistently outperform the baselines.

\section{Concluding Remarks}\label{sec:conclusions}

In this paper, we studied  provable multi-agent RL in  partially observable environments, with both statistical and computational (quasi-)efficiencies.   The key to our results is to identify  the value of \emph{information sharing}, a common practice in empirical MARL and a standard  phenomenon in many multi-agent control systems, in algorithm design and  computation/sample efficiency analysis.  We hope our study may open up the possibilities of leveraging and even designing different  \emph{information structures}, for  developing both statistically and  computationally efficient partially observable MARL algorithms. One open problem and future direction is to develop a fully \emph{decentralized} algorithm and overcome \emph{the curse of multiagents}, such that the sample and computation complexities do not grow exponentially with the number of agents. Another interesting direction is to identify the combination of  certain information-sharing structures and observability assumptions   for more efficient (e.g., polynomial) sample and computation complexity results.  
\vspace{-1mm}

\section*{Acknowledgement} 
The authors would like to thank the anonymous reviewers from ICML 2023  for their helpful comments.  The authors would also like to  thank Noah Golowich and Serdar Y\"uksel for the valuable feedback and discussions. X.L and K.Z.  acknowledge the support from Simons-Berkeley Research Fellowship, Northrop Grumman -- Maryland Seed Grant Program, 
Army Research Office (ARO) grant W911NF-24-1-0085, NSF CAREER Award 2443704, and AFOSR YIP Award FA9550-25-1-0258.  
  
\bibliographystyle{ims}
\bibliography{main}

\appendix

\onecolumn

~\\
\centerline{{\fontsize{14}{14}\selectfont \textbf{Appendices}}}

\vspace{20pt}

\section{Additional Definitions}

\subsection{Belief states}
\label{subsec:belief}

In partially observable environments, each agent cannot know the underlying state but could infer the underlying distribution of states through the  observations and actions. Following the convention in POMDPs, we call such distributions the belief states. Such posterior distributions over  states can be updated whenever the agent receives new observations and actions. Formally, we define the belief update as:
\begin{definition}[Belief state update]\label{def:belief-state}
For each $h\in [H+1]$, the Bayes operator (with respect to the joint observation)  $B_{h}:\Delta(\cS)\times \cO\rightarrow \Delta(\cS)$ is defined for $b\in\Delta(\cS)$, and $y\in\cO$ by:
\[
    B_h(b ; y)(x)=\frac{\mathbb{O}_h(y \mid x) b(x)}{\sum_{z \in \mathcal{S}} \mathbb{O}_h(y \mid z) b(z)}.
\]
Similarly, for each $h\in [H], i \in [n]$, we define the Bayes operator with respect to individual observations $B_{i, h}:\Delta(\cS)\times\cO_i\rightarrow\Delta(\cS)$ by:
\[
    B_{i,h}(b ; y)(x)=\frac{\mathbb{O}_{i, h}(y \mid x) b(x)}{\sum_{z \in \mathcal{S}} \mathbb{O}_{i, h}(y \mid z) b(z)}.
\]
For each $h\in [H]$, the belief update operator $U_h: \Delta(\mathcal{S}) \times \mathcal{A} \times \mathcal{O} \rightarrow \Delta(\mathcal{S})$, is defined by 
\[
    U_h(b ; a, y)=B_{h+1}\left(\mathbb{T}_h(a) \cdot b ; y\right),
\]
where $\mathbb{T}_h(a) \cdot b$ represents the matrix multiplication. We use the notation $\bb_h$ to denote the belief update function, which receives a sequence of actions and observations and outputs a distribution over states at the step $h$. The belief state at step $h=1$ is defined as $\bb_1(\emptyset) = \mu_1$. For any $1\le h\le H$ and any action-observation sequence $(a_{1:h-1}, o_{1:h})$, we inductively define the belief state:
\[
    \bb_{h+1}(a_{1:h}, o_{1:h}) = \mathbb{T}_h(a_h)\cdot \bb_{h}( a_{1:h-1}, o_{1:h}),
\]
\[
    \bb_{h}(a_{1:h-1}, o_{1:h}) = B_{h}(\bb_{h}(a_{1:h-1}, o_{1:h-1}); o_h).
\]
Also, we slightly abuse the notation and define the belief state containing individual observations as
\[
    \bb_{h}(a_{1:h-1}, o_{1:h-1}, o_{i, h}) = B_{i,h}(\bb_{h}(a_{1:h-1}, o_{1:h-1}); o_{i, h}).
\]
We define the approximate belief update using the most recent $L$-step history. For $1\le h \le H$, we follow the notation of \cite{golowich2022planning} and define
\[
    \bb_h^{\mathrm{apx},\cG}(\emptyset;{D})= \begin{cases}\mu_1 & \text { if } h=1 \\ {D} & \text { otherwise },\end{cases}
\]
where ${D}\in\Delta(\cS)$ is the prior for the approximate belief update. 
Then for any $1\le h-L< h\le H$ and any action-observation sequence $(a_{h-L:h-1}, o_{h-L+1:h})$, we inductively define
\[
\bb_{h+1}^{\mathrm{apx}, \cG}(a_{h-L:h}, o_{h-L+1:h};{D}) = \mathbb{T}_h(a_h)\cdot \bb^{\mathrm{apx}, \cG}_{h}(a_{h-L:h-1}, o_{h-L+1:h};{D}),
\]
\[
\bb_{h}^{\mathrm{apx}, \cG}(a_{h-L:h-1}, o_{h-L+1:h};{D}) = B_{h}(\bb^{\mathrm{apx}, \cG}_{h}(a_{h-L:h-1}, o_{h-L+1:h-1};{D}); o_h).
\]
For the remainder of our paper, we shall use the important initialization for the approximate belief, which are defined as $\bb^{\prime}_{h}(\cdot):=\bb_{h}^{\mathrm{apx}, \cG}(\cdot;\operatorname{Unif}(\cS))$.

\end{definition}

\subsection{Additional definitions of value functions and policies}\label{sec:additional_def}

In \Cref{def:g_value_functioin}, we have defined value functions in $\cG$. Similar to the fully-observable settings (MDPs and stochastic games), we can also extend such a definition to the prescription-value function, which corresponds to the action-value function in the fully-observable settings.

\begin{definition}[Prescription-value function {with information sharing}]\label{def:pres_value}
At step $h\in[H]$, given the common information $c_h$, joint policies ${\pi}=\{\pi_{i}\}_{i=1}^{n}\in\Pi$, 
and prescriptions $\{\gamma_{i, h}\}_{i=1}^{n}\in\Gamma_h$, the \emph{prescription-value function  {conditioned on the common information and joint prescription}} of the $i^{th}$ agent is defined as: 
\[
    Q^{{\pi}, \cG}_{i, h}(c_h, \{\gamma_{j, h}\}_{j\in [n]}):= \EE_{\pi}^\cG\big[r_{i, h}(s_h, a_h) + V^{{\pi}, \cG}_{i, h+1}(c_{h+1})\Biggiven c_h, \{\gamma_{j, h}\}_{j\in [n]}\big],
\]
where  prescription $\gamma_{i, h}\in \Gamma_{i, h}$ replaces the partial function $\pi_{i, h}(\cdot\given \omega_{i, h}, c_h, \cdot)$ in the value function. 
\end{definition}

With the expected approximate common information model $\cM$ given in Definition \ref{def:ais}, we can define the value function {and policy} under $\cM$ accordingly  as follows. 

\begin{definition}[Value function and policy under $\cM$]\label{def:M_value_func} Given an expected approximate common information model $\cM$, for any  policy $\pi\in\Pi$, for each $i\in [n], h\in [H]$, we define the value function as
\begin{align}\label{adp}
    V^{{\pi}, \cM}_{i, h}(c_h) &= \EE_{\{\omega_{j, h}\}_{j\in [n]}}\Big[\hat{r}_{i, h}^\cM(\hat{c}_h, {\{\pi_{j, h}(\cdot\given \omega_{j, h}, c_h, \cdot)\}_{j\in [n]}}) +\EE^{\cM}[V^{{\pi}, \cM}_{i, h+1}(c_{h+1})\given \hat{c}_h, {\{\pi_{j, h}(\cdot\given \omega_{j, h}, c_h, \cdot)\}_{j\in [n]}}]\Big].
\end{align}
For any {$c_{H+1}\in \cC_{H+1}$}, we define {$V_{i, H+1}^{\pi, \cM}(c_{H+1}) = 0$}. Furthermore,  for a policy $\hat{\pi}$ whose  $\hat{\pi}_{i,h}: \Omega_{h}\times\cP_{i, h}\times \hat{\cC}_{h}\rightarrow \Delta({\mathcal{A}_i})$ takes \emph{approximate} instead of the exact common information as the input, we define 
\begin{align}\label{adp_2}
    V^{{\hat{\pi}}, \cM}_{i, h}(\hat{c}_h) &= \EE_{\{\omega_{j, h}\}_{j\in [n]}}\Big[\hat{r}_{i, h}^\cM(\hat{c}_h, \{\hat{\pi}_{j, h}(\cdot\given \omega_{j, h}, \hat{c}_h, \cdot)\}_{j\in [n]})+\EE^{\cM}[ V^{\hat{\pi}, \cM}_{i, h+1}(\hat{c}_{h+1})\given \hat{c}_h, \{\hat{\pi}_{j, h}(\cdot\given \omega_{j, h}, \hat{c}_h, \cdot)\}_{j\in [n]}]\Big], 
\end{align}
{where similarly, for each  $\hat{c}_{H+1}\in\hat{\cC}_{H+1}$, we define $V^{\hat{\pi}, \cM}_{i, H+1}(\hat{c}_{H+1}) = 0$. With a slight abuse of notation, sometimes $\hat{\pi}_{i,h}$ may also take $c_h\in\cC_h$ as input and thus $\hat{\pi}\in \Pi$. In this case, when $\cM$ and the corresponding compression  function $\operatorname{Compress}_{h}$ are clear from the context, it means $\hat{\pi}_{i, h}(\cdot\given \cdot, c_h, \cdot) := \hat{\pi}_{i, h}(\cdot\given \cdot, \operatorname{Compress}_h(c_h), \cdot)$. Accordingly, in this case, the definitions of $V_{i, h}^{\hat{\pi}, \cG}(c_h)$ and $V_{i, h}^{\hat{\pi}, \cM}(c_h)$ follows from  Definition \ref{def:g_value_functioin} and Equation \eqref{adp}, respectively. 
} 
\end{definition}

\section{Collection of Algorithm Pseudocodes} \label{sec:alg_app_new}

Here we collect both our planning and learning algorithms as in Algorithms  \ref{alg:vi}, \ref{alg:br}, \ref{alg:avi}, \ref{alg:abr}, \ref{alg:construct_mg}, \ref{alg:plam}, \ref{alg:selection}, \ref{alg:selection-dec}, \ref{alg:learning}. 

\begin{figure*}[!t]
\centering   
\resizebox{16.8cm}{!}{
\hspace{-25pt}
\includegraphics[width=\textwidth]{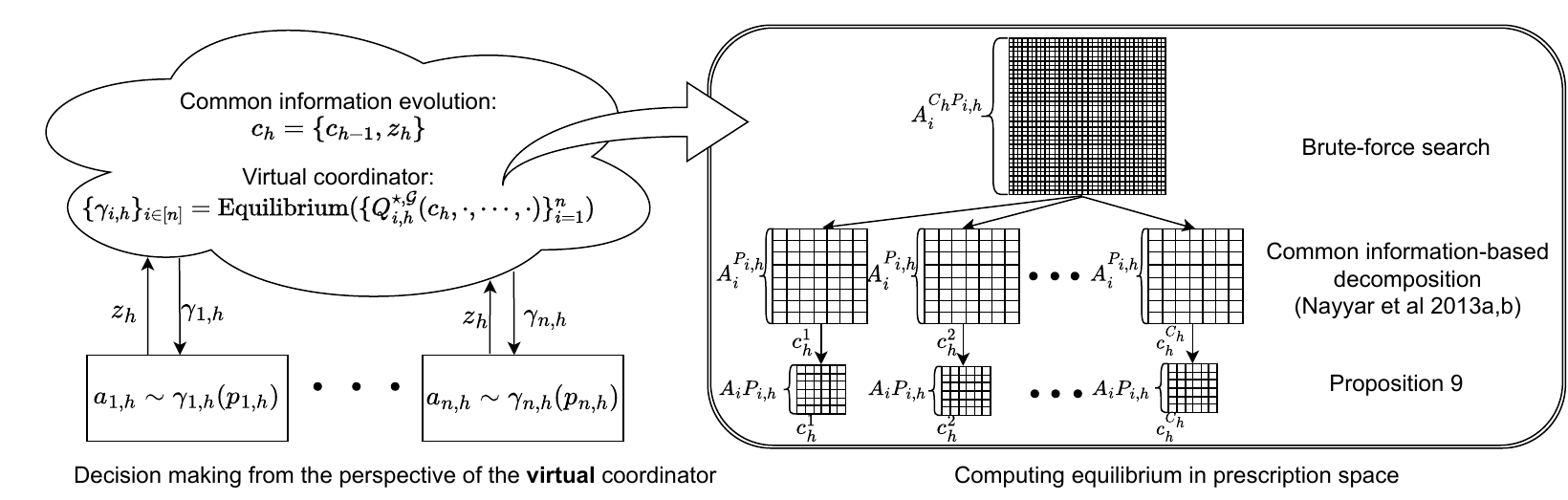}} 
\caption{An overview of our algorithmic framework. The left part of the figure shows that there is a virtual coordinator collecting the information shared among agents. Based on the common information $c_h$, it will compute an equilibrium in the prescription space and assign it to all the agents. The right part shows the computation of equilibrium. Let's take the example of $A_i = 2$, $P_{i, h} = 3$, $C_h=2$. If we search over all deterministic prescriptions, the corresponding matrix game will have the size of $A_i^{C_hP_{i,h}} = 64$. Then,  \cite{nayyar2013common,nayyar2013decentralized} proposed the common information-based decomposition, and solve  $C_h$ number of  games of smaller size. However, in the Dec-POMDP setting, \cite{nayyar2013decentralized} treated each deterministic prescription as an action and the size of each sub-problem will be $A_i^{P_{i, h}} = 8$. Furthermore, Proposition \ref{prop:linear} shows that  we can reformulate each sub-problem as a game {whose payoff is} multi-linear with respect to each agent's prescription{, and whose dimensionality is} $A_iP_{i, h}=6$.} 
\label{fig:framework}
\end{figure*}

\begin{algorithm}[!h]
\caption{Value iteration with common information}\label{alg:vi}
\begin{algorithmic}[1]
\State  \textbf{Input:} $\cG, \epsilon_e$
\For{each $i\in [n]$ and $c_{H+1}$}
    \State $V^{\star,  \cG}_{i, H+1}(c_{H+1}) \leftarrow 0$
\EndFor
\For{$h = H, \cdots, 1$}
    \For{each $c_h$}
        \State Define 
        \$
        Q^{\star,  \cG}_{i, h}(c_h, \gamma_{1, h}, \cdots, \gamma_{n, h}) &:= \EE_{s_h, p_{h}\sim \PP_h^ \cG\left(\cdot, \cdot\given c_h\right)}\EE_{\{a_{j, h}\sim{\gamma}_{j,h}(\cdot\given p_{j,h})\}_{j\in [n]}}\EE_{o_{h+1}\sim \OO_{h+1}^{\top}\TT_{h}(\cdot\given s_h, a_h)}\Big[r_{i, h}(s_h, a_h) + V^{\star,  \cG}_{i, h+1}(c_{h+1})\Big]
        \$
        \State \$\big\{\pi_{1, h}^{\star}(\cdot\given\cdot, c_h, \cdot), \cdots, \pi_{n, h}^{\star}(\cdot\given\cdot, c_h, \cdot)\big\} \leftarrow \operatorname{NE/CE/CCE}(\{Q_{i, h}^{\star,  \cG}(c_h, \cdot, \cdots, \cdot)\}_{i=1}^{n}, \epsilon_e)\$\quad // we refer the implementation to \Cref{subsec:subroutine}
        \For{each $i\in [n]$}
            \[
            V^{\star,  \cG}_{i, h}(c_h) \leftarrow  \EE_{\{\omega_{j,h}\}_{j\in [n]}} \EE^\cG \Big[r_{i, h}(s_h, a_h) + V^{\star,  \cG}_{i, h+1}(c_{h+1})\given c_h, \{\pi^{\star}_{j, h}(\cdot\given\omega_{j, h}, c_h, \cdot)\}_{j\in [n]}\Big]
            \]
        \EndFor
    \EndFor
\EndFor
\State \Return $\pi^\star$ 
\end{algorithmic}
\end{algorithm}

\begin{algorithm}[!h]
\caption{\texttt{BR$(\cG, \pi, i,\epsilon_e)$}: $\epsilon_e$-approximate \textbf{B}est \textbf{R}esponse  for the $i^{th}$ agent under true model $\cG$} \label{alg:br}
\begin{algorithmic}[1]
\State  \textbf{Input:} $\cG, \pi, i,\epsilon_e$
    \State ${V}^{\star,\cG   }_{i, H+1}({c}_{H+1}) \leftarrow 0$ for all ${c}_{H+1}$
\For{$h = H, \cdots, 1$}
    \For{each ${c}_h$}
        \State Define 
        \${Q}^{\star,\cG   }_{i, h}({c}_h, \gamma_{1, h}, \cdots, \gamma_{n, h}) := \EE_{s_h, p_{h}\sim \PP_h^ \cG\left(\cdot, \cdot\given c_h\right)}\EE_{\{a_{j, h}\sim{\gamma}_{j,h}(\cdot\given p_{j,h})\}_{j\in [n]}}
        \EE_{o_{h+1}\sim \OO_{h+1}^{\top}\TT_{h}(\cdot\given s_h, a_h)}\Big[r_{i, h}(s_h, a_h) + V^{\star,  \cG}_{i, h+1}(c_{h+1})\Big]
        \$
        \State \${\pi}_{i, h}^{\star}(\cdot\given\cdot, {c}_h, \cdot) \leftarrow \operatorname{NE/CE/CCE-BR}({Q}_{i, h}^{\star,\cG}({c}_h, \cdot, \cdots, \cdot), \{\pi_{j, h}(\cdot\given \cdot, {c}_h, \cdot)\}_{j\in [n]},i,\epsilon_e)\$ \quad // we refer the implementation to \Cref{subsec:subroutine}
            \State \${V}^{\star,\cG   }_{i, h}({c}_h) \leftarrow &\EE_{\{\omega_{j,h}\}_{j\in [n]}}\EE_{s_h, p_{h}\sim \PP_h^ \cG\left(\cdot, \cdot\given c_h\right)}\EE_{\substack{a_{i, h}\sim \pi_{i, h}^\star(\cdot\given \omega_{i, h}, c_h, p_{i, h}),\\ a_{-i, h}\sim \pi_{-i, h}(\cdot\given \omega_{-i, h}, c_h, p_{-i, h})}}\EE_{o_{h+1}\sim \OO_{h+1}^{\top}\TT_{h}(\cdot\given s_h, a_h)}\Big[r_{i, h}(s_h, a_h) + V^{\star,  \cG}_{i, h+1}(c_{h+1})\Big]\$
    \EndFor
\EndFor
\State \Return ${\pi}_i^\star$
\end{algorithmic}
\end{algorithm}

\begin{algorithm}[!h]
\caption{\texttt{VIACM$(\cM, \epsilon_e)$}: \textbf{V}alue \textbf{I}teration with expected \textbf{A}pproximate {\textbf{C}ommon-information} \textbf{M}odel}\label{alg:avi}
\begin{algorithmic}[1]
\State  \textbf{Input:} $\cM, \epsilon_e$
\For{each $i\in [n]$ and $\hat{c}_{H+1}$} 
    \State ${V}^{\star,\cM   }_{i, H+1}(\hat{c}_{H+1}) \leftarrow 0$
\EndFor
\For{$h = H, \cdots, 1$}
    \For{each $\hat{c}_h$}
        \State Define ${Q}^{\star,\cM   }_{i, h}(\hat{c}_h, \gamma_{1, h}, \cdots, \gamma_{n, h}) := \hat{r}_{i, h}^\cM(\hat{c}_h, \gamma_h) + \EE^\cM\Big[ {V}^{\star,\cM}_{i, h+1}(\hat{c}_{h+1})\mid \hat{c}_h, \{\gamma_{j, h}\}_{j\in [n]}\Big]$ for any $i\in [n]$
        \If{computing the equilibrium}
        \State \$\big\{\hat{\pi}_{1, h}^{\star}(\cdot\given\cdot, \hat{c}_h, \cdot), \cdots, \hat{\pi}_{n, h}^{\star}(\cdot\given\cdot, \hat{c}_h, \cdot)\big\} \leftarrow \operatorname{NE/CE/CCE}(\{{Q}_{i, h}^{\star,\cM   }(\hat{c}_h, \cdot, \cdots, \cdot)\}_{i=1}^{n}, \epsilon_e)\$\qquad // we refer the implementation to \Cref{subsec:subroutine}
        \ElsIf{computing the team-optimum}
        	\State \$\big\{\hat{\pi}_{1, h}^{\star}(\cdot\given \hat{c}_h, \cdot), \cdots, \hat{\pi}_{n, h}^{\star}(\cdot\given \hat{c}_h, \cdot)\big\} \leftarrow \arg\max_{\left\{\gamma_{i, h}\in\Delta(\cA_i)^{\cP_{i,h}}\right\}_{i\in[n]}}({Q}_{1, h}^{\star,\cM}(\hat{c}_h, \gamma_{1, h}, \cdots, \gamma_{n, h}))\$\qquad // we refer the implementation to \Cref{supp-sec-5}
        \EndIf
        \For{each $i\in [n]$}
            \State ${V}^{\star,\cM   }_{i, h}(\hat{c}_h) \leftarrow \EE_{\{\omega_{j, h}\}_{j\in [n]}}\Big[\hat{r}_{i, h}^\cM(\hat{c}_h, \{\hat{\pi}_{j, h}^\star(\cdot\given\omega_{j, h}, \hat{c}_h, \cdot)\}_{j\in [n]}) +\EE^\cM[ {V}^{\star,\cM}_{i, h+1}(\hat{c}_{h+1})\mid \hat{c}_h, \{\hat{\pi}_{j, h}^\star(\cdot\given\omega_{j, h}, \hat{c}_h, \cdot)\}_{j\in [n]}]\Big]$
        \EndFor
    \EndFor
\EndFor
\State \Return $\hat{\pi}^\star$
\end{algorithmic}
\end{algorithm}

\begin{algorithm}[!h]
\caption{\texttt{ABR$(\cM, \hat{\pi}, i,\epsilon_e)$}: $\epsilon_e$-{a}pproximate \textbf{B}est \textbf{R}esponse  for the $i^{th}$ agent under \textbf{A}pproximate common information model $\cM$}\label{alg:abr}
\begin{algorithmic}[1]
\State  \textbf{Input:} $\cM, \hat{\pi}, i,\epsilon_e$
    \State ${V}^{\star,\cM   }_{i, H+1}(\hat{c}_{H+1}) \leftarrow 0$ for all $\hat{c}_{H+1}$ 
\For{$h = H, \cdots, 1$}
    \For{each $\hat{c}_h$}
        \State Define ${Q}^{\star,\cM   }_{i, h}(\hat{c}_h, \gamma_{1, h}, \cdots, \gamma_{n, h}) := \hat{r}_{i, h}^\cM(\hat{c}_h, \gamma_h) + \EE^\cM\Big[ {V}^{\star,\cM}_{i, h+1}(\hat{c}_{h+1})\mid \hat{c}_h, \{\gamma_{j, h}\}_{j\in [n]}\Big]$ for any $i\in [n]$
        \State \$\hat{\pi}_{i, h}^{\star}(\cdot\given\cdot, \hat{c}_h, \cdot) \leftarrow \operatorname{NE/CE/CCE-BR}({Q}_{i, h}^{\star,\cM}(\hat{c}_h, \cdot, \cdots, \cdot), \{\hat{\pi}_{j, h}(\cdot\given \cdot, \hat{c}_h, \cdot)\}_{j\in [n]},i,\epsilon_e)\$\quad // we refer the implementation to \Cref{subsec:subroutine}
            \State ${V}^{\star,\cM   }_{i, h}(\hat{c}_h) \leftarrow \EE_{\{\omega_{j, h}\}_{j\in [n]}}\Big[\hat{r}_{i, h}^\cM(\hat{c}_h, \{\hat{\pi}_{i, h}^\star(\cdot\given\omega_{i, h}, \hat{c}_h, \cdot),\hat{\pi}_{-i, h}(\cdot\given\omega_{-i, h}, \hat{c}_h, \cdot)\}) + \EE^\cM[{V}^{\star,\cM}_{i, h+1}(\hat{c}_{h+1})\mid \hat{c}_h, \{\hat{\pi}_{i, h}^\star(\cdot\given\omega_{i, h}, \hat{c}_h, \cdot),\hat{\pi}_{-i, h}(\cdot\given\omega_{-i, h}, \hat{c}_h, \cdot)\}]\Big]$
    \EndFor
\EndFor
\State \Return $\hat{\pi}_i^\star$
\end{algorithmic}
\end{algorithm}

\begin{algorithm}[!h]
\caption{\texttt{{LEE}}$(\pi^{1:H}, \{\hat{\cC}_h\}_{h\in[H+1]},\{\hat{\phi}_{h+1}\}_{h\in[H]}, \Gamma, \zeta_1, \zeta_2, \theta_1, \theta_2, \delta_1)$:  \textbf{L}earning \textbf{E}mpirical \textbf{E}stimator $\hat{\cM}(\pi^{1:H})$ of $\Tilde{\cM}(\pi^{1:H})$}\label{alg:construct_mg}
\begin{algorithmic}[1]

\State  \textbf{Input:} $\pi^{1:H}, \{\hat{\cC}_h\}_{h\in[H+1]},\{\hat{\phi}_{h+1}\}_{h\in[H]}, \Gamma, \zeta_1, \zeta_2, \theta_1, \theta_2, \delta_1$
\For{$1\le h\le H$}
	\State Define $N_0$ as in Equation \eqref{eq:n0}.
    \State Draw $N_0$ independent trajectories by executing the policy $\pi^{h}$, and denote the $k^{th}$ trajectory by $(a_{1:H-1}^k, o_{1:H}^k, r_{1:H}^k)$, for $k\in [N_0]$, where $N_0$ is specified in \Cref{thm:main_thm_learning}.
    \For{each $\hat{c}_h\in \hat{\cC}_h$}
        \State Define $\varphi(p_h):= |\{k:\operatorname{Compress}_h(f_h(a_{1:h-1}^k, o_{1:h}^k)) = \hat{c}_h,\text{ and } g_{h}(a_{1:h-1}^k, o_{1:h}^k) = p_h\}|$.
        \State Set $\PP_{h}^{\hat{\cM}(\pi^{1:H})}(p_h\given \hat{c}_h) := \frac{\varphi(p_h)}{\sum_{p_h^\prime}\varphi(p_h^\prime)}$ for all $p_h\in  \cP_h$.
    \EndFor
    \For{each $\hat{c}_h\in \hat{\cC}_h$, $p_h\in  \cP_h$, $a_h\in \cA$}
            \State Define $\psi(o_{h+1}):= |\{k:\operatorname{Compress}_h(f_h(a_{1:h-1}^k, o_{1:h}^k)) = \hat{c}_h, g_{h}(a_{1:h-1}^k, o_{1:h}^k) = p_h, a_{h}^k=a_h, \text{ and }o_{h+1}^k=o_{h+1}\}|$.
        \State Set $\PP_{h}^{\hat{\cM}(\pi^{1:H})}(o_{h+1}\given \hat{c}_h, p_h, a_h) := \frac{\psi(o_{h+1})}{\sum_{o_{h+1}^\prime}\psi(o_{h+1}^\prime)}$ for all $o_{h+1}\in \cO$.

        \State Define $\kappa(\hat{c}_h, p_h, a_h):= \{k:\operatorname{Compress}_h(f_h(a_{1:h-1}^k, o_{1:h}^k)) = \hat{c}_h, g_{h}(a_{1:h-1}^k, o_{1:h}^k) = p_h, a_{h}^k=a_h, \text{ and }o_{h+1}^k=o_{h+1}\}$.
        \State Set $\hat{r}_{i, h}^{\hat{\cM}(\pi^{1:H})}(\hat{c}_h, p_h, a_h) := \frac{\sum_{k\in\kappa(\hat{c}_h, p_h, a_h)} r_{i, h}^k}{|\kappa(\hat{c}_h, p_h, a_h)|}$ for all $i\in[n]$.
    \EndFor
\EndFor
\State Define for any $h\in [H]$, $\hat{c}_h\in\hat{\cC}_h$, $\gamma_h\in\Gamma_h$, $o_{h+1}\in\cO_{h+1}$, $z_{h+1}\in\cZ_{h+1}$: 
	\#
 \label{eq:trans-es}&\PP_{h}^{\hat{\cM}(\pi^{1:H}), z}(z_{h+1}\given \hat{c}_h, \gamma_h)\leftarrow \sum_{\substack{p_h, a_h, o_{h+1}}}\bm{1}[\chi_{h+1}(p_h, a_h, o_{h+1})=z_{h+1}]
	\times\PP_{h}^{\hat{\cM}(\pi^{1:H})}(p_h\given \hat{c}_h)\gamma_{h}(a_h\given p_h)\PP_{h}^{\hat{\cM}(\pi^{1:H})}(o_{h+1}\given \hat{c}_h, p_{h}, a_{h})\\
	\label{eq:reward-es}&\hat{r}_{i, h}^{\hat{\cM}(\pi^{1:H})} (\hat{c}_h, \gamma_h)\leftarrow \sum_{p_h, a_h}\PP_{h}^{\hat{\cM}(\pi^{1:H})}(p_h\given \hat{c}_h)\gamma_h(a_h\given p_h)\hat{r}_{i, h}^{\hat{\cM}(\pi^{1:H})}(\hat{c}_h, p_h, a_h),
	\#
	where we recall $\gamma_h(a_h\given p_h):=\prod_{j=1}^n\gamma_{j, h}(a_{j, h}\given p_{j, h})$.
\State \Return $\hat{\cM}(\pi^{1:H}):=(\{\hat{\cC}_h\}_{h\in[H+1]},\{\hat{\phi}_{h+1}\}_{h\in[H]}, \{\PP_{h}^{\hat{\cM}(\pi^{1:H}), z}\}_{h\in [H]}, \Gamma,\hat{r})$
\end{algorithmic}
\end{algorithm}

\begin{algorithm}[!h]
\caption{\texttt{{Plam}}$(\pi^{1:H}, \{\hat{\cC}_h\}_{h\in[H+1]},\{\hat{\phi}_{h+1}\}_{h\in[H]}, \Gamma, \zeta_1, \zeta_2, \theta_1, \theta_2, \delta_1, \epsilon_e)$:  \textbf{P}lanning in \textbf{l}earned \textbf{a}pproximate \textbf{m}odel}\label{alg:plam}
\begin{algorithmic}[1]

\State  \textbf{Input:} $\pi^{1:H}, \{\hat{\cC}_h\}_{h\in[H+1]},\{\hat{\phi}_{h+1}\}_{h\in[H]}, \Gamma, \zeta_1, \zeta_2, \theta_1, \theta_2, \delta_1, \epsilon_e$
\State $\hat{\cM}(\pi^{1:H})\leftarrow \texttt{Construct}(\pi^{1:H}, \{\hat{\cC}_h\}_{h\in[H+1]},\{\hat{\phi}_{h+1}\}_{h\in[H]}, \Gamma, \zeta_1, \zeta_2, \theta_1, \theta_2, \delta_1)$\quad // i.e., Algorithm \ref{alg:construct_mg}
\State $\pi^{\star}\leftarrow \texttt{VIACM}(\hat{\cM}(\pi^{1:H}), \epsilon_e)$\quad // i.e., Algorithm \ref{alg:avi}
\State \Return $\left\{\pi^{\star}, \hat{\cM}(\pi^{1:H})\right\}$
\end{algorithmic}
\end{algorithm}

\begin{algorithm}[!h]
\caption{\texttt{PoS}$(\{\hat{\cM}(\pi^{1:H, m})\}_{m\in [K]}, \{\pi^{\star, j}\}_{j\in[K]}, \epsilon_e, N_2)$: \textbf{Po}licy \textbf{S}election}\label{alg:selection}
\begin{algorithmic}[1]
\State  \textbf{Input:} $\{\hat{\cM}(\pi^{1:H, j})\}_{j\in [K]}, \{\pi^{\star, j}\}_{j\in[K]}, \epsilon_e, N_2$
\For{$i\in [n], j\in [K], m\in [K]$}
    \State $\pi_{i}^{\star, j, m}\leftarrow \texttt{ABR}(\hat{\cM}(\pi^{1:H, m}), \pi^{\star, j}, i,\epsilon_e)$\quad // i.e., Algorithm \ref{alg:abr}
\EndFor
\For{$j \in [K]$}
    \State Execute $\pi^{\star, j}$ for $N_2$ trajectories and let the mean accumulated reward for the $i^{th}$ agent be $R_{i}^{j}$
\EndFor
\For{$i\in [n], j\in [K], m\in [K]$}
    \State Execute {$\pi_{i}^{\star, j, m}\odot \pi_{-i}^{\star, j}$} for $N_2$ trajectories and let the mean accumulated reward for the $i^{th}$ agent  be $R_{i}^{j, m}$
\EndFor
\State $\hat{j}\leftarrow \arg\min_{j\in[K]} \left(\max_{i\in[n]}\max_{m\in [K]} (R_{i}^{j, m} - R_{i}^j)\right)$
\State \Return $\pi^{\star, \hat{j}}$
\end{algorithmic}
\end{algorithm}

\begin{algorithm}[!h]
\caption{\texttt{PoS-Dec}$(\{\pi^{\star, j}\}_{j\in[K]}, N_2)$: \textbf{Po}licy \textbf{S}election for \textbf{Dec}-POMDP}\label{alg:selection-dec}
\begin{algorithmic}[1]
\State  \textbf{Input:} $\{\pi^{\star, j}\}_{j\in[K]}, N_2$
\For{$j \in [K]$}
    \State Execute $\pi^{\star, j}$ for $N_2$ trajectories and let the mean accumulated reward be $R^{j}$
\EndFor
\State $\hat{j}\leftarrow \arg\max_{j\in[K]} R^j$
\State \Return $\pi^{\star, \hat{j}}$
\end{algorithmic}
\end{algorithm}

\begin{algorithm}[!h]
\caption{\texttt{LACI}$(\cG, \{\hat{\cC}_h\}_{h\in[H+1]},\{\hat{\phi}_{h+1}\}_{h\in[H]}, \Gamma, \hat{L}, \epsilon, \delta_2, \zeta_1, \zeta_2, \theta_1, \theta_2, \delta_1, $ $N_2, \epsilon_e)$: \textbf{L}earning with   \textbf{A}pproximate \textbf{C}ommon \textbf{I}nformation}\label{alg:learning}
\begin{algorithmic}[1]
\State  \textbf{Input:} $\cG, \{\hat{\cC}_h\}_{h\in[H+1]},\{\hat{\phi}_{h+1}\}_{h\in[H]}, \Gamma, \hat{L}, \epsilon, \delta_2, \zeta_1, \zeta_2, \theta_1, \theta_2, \delta_1, N_2, \epsilon_e$
\State $\{\pi^{1:H, j}\}_{j=1}^K\leftarrow \texttt{BaSeCAMP}( \cG, \hat{L}, \epsilon, \delta_2)$\quad // i.e., Algorithm 3 of \cite{golowich2022learning} 
\For{$j\in [K]$} 
\State 
$\left\{\pi^{\star, j}, \hat{\cM}(\pi^{1:H, j})\right\}\leftarrow \texttt{Plam}(\pi^{1:H, j},\{\hat{\cC}_h\}_{h\in[H+1]},\{\hat{\phi}_{h+1}\}_{h\in[H]}, \Gamma, \zeta_1, \zeta_2,\theta_1, \theta_2, \delta_1, \epsilon_e)$\quad // i.e., Algorithm \ref{alg:plam}

\EndFor
\If{learning the equilibrium}

\State $\pi^{\star, \hat{j}}\leftarrow \texttt{PoS}(\{\hat{\cM}(\pi^{1:H, j})\}_{j=1}^{K}, \{\pi^{\star, j}\}_{j=1}^K,\epsilon_e, N_2)$\quad // i.e., Algorithm \ref{alg:selection}
\ElsIf{learning the team-optimum}
	\State $\pi^{\star, \hat{j}}\leftarrow \texttt{PoS-Dec}(\{\pi^{\star, j}\}_{j=1}^K, N_2)$\quad // i.e., Algorithm \ref{alg:selection-dec}
\EndIf
\State \Return $\pi^{\star, \hat{j}}$
\end{algorithmic}
\end{algorithm}

\begin{algorithm}[!h]
\caption{\texttt{ADPNIS$(\PP_h^{\cM, c}(\cdot,\cdot\given\hat{c}_h))$}: \textbf{A}gent-based \textbf{D}ynamic \textbf{P}rogramming under \textbf{N}ested \text{I}nformation-\textbf{S}haring}\label{alg:agent-dp}
\begin{algorithmic}[1]
\State  \textbf{Input:} $\PP_h^{\cM, c}(\cdot,\cdot\given\hat{c}_h)$
\State Initialize $V_{n+1}(p_{1:n+1,h}, a_{1:n, h})\leftarrow\EE_{s_h\sim\PP^{\cM, c}_h(\cdot\given \hat{c}_h, p_{h}), s_{h+1}\sim\TT_h(\cdot\given s_h, a_h)}$ $\EE_{o_{h+1}\sim\OO_{h+1}(\cdot\given s_{h+1})}\Big[r_{h}(s_h, a_h) + V_{h+1}^{\star, \cM}(\hat{c}_{h+1})\Big]$ for any $p_{h}\in\cP_h$, $a_h\in\cA$
\For{$i = n, \cdots, 1$}
    \For{each $p_{1:i, h}\in\times_{j=1}^i\cP_{j, h}$, $a_{1:i-1, h}\in\times_{j=1}^{i-1}\cA_j$}
        \State \$u_{i}^\star(p_{1:i, h}, a_{1:i-1, h})\leftarrow\arg\max_{a_{i, h}\in\cA_i} \EE_{p_{i+1, h}\sim \PP^{\cM, c}_h(\cdot\given \hat{c}_h, p_{1:i, h})} V_{i+1}(p_{1:(i+1), h}, a_{1:i, h})\$
        \State \$V_{i}(p_{1:i, h}, a_{1:i-1, h})\leftarrow \max_{a_{i, h}\in\cA_i} \EE_{p_{i+1, h}\sim \PP^{\cM, c}_h(\cdot\given \hat{c}_h, p_{1:i, h})} V_{i+1}(p_{1:(i+1), h}, a_{1:i, h})\$
    \EndFor
\EndFor

\For{$i = 1, \cdots, n$}
    \For{each $p_{i, h}\in\cP_{i, h}$}
    	\For{each $j=1, \cdots, i$}
    		\State $p_{j, h}\leftarrow Y_{h}^{ij}(p_{i, h})$
    		\State $a_{j, h}^\star\leftarrow u_{j}^\star(p_{1:j, h}, a_{1:j-1}^\star)$
    	\EndFor
        \State $\gamma_{i, h}^\star(p_{i, h})\leftarrow a_{i, h}^\star$
    \EndFor
\EndFor

\State \Return $\{\gamma_{i, h}^\star\}_{i\in[n]}$
\end{algorithmic}
\end{algorithm}

\section{Full Versions  of the Results}

\subsection{Planning}\label{subsec:plan}

Now we state the full version of Theorem \ref{thm:plan_cases} regarding the instantiations of Theorem \ref{thm:struc}.

\begin{theorem}\label{thm:plan_cases_full}
Fix $\epsilon>0$. Suppose there exists an  $(\epsilon_r, \epsilon_z)$-expected-approximate common information model $\cM$ consistent with some given approximate belief $\{\PP_{h}^{\cM, c}(s_h, p_h\given \hat{c}_h)\}_{h\in [H]}$ for the POSG $\cG$ {under Assumptions \ref{evo} and \ref{str_indi}} such that $\max\{\epsilon_z(\cM), \epsilon_r(\cM)\}\le \cO(\epsilon)$ and $\max_h\hat{C}_hP_h$ is quasi-polynomial of the problem instance size, then there exists a quasi-polynomial time algorithm that can compute an $\epsilon$-NE if $\cG$ is zero-sum or cooperative, and an  $\epsilon$-CE/CCE if $\cG$.

 In particular, under Assumption \ref{observa}, examples in \Cref{sec:example} satisfy  all such conditions. Therefore, there exists a quasi-polynomial time algorithm computing $\epsilon$-NE if $\cG$ is zero-sum or cooperative and $\epsilon$-CE/CCE if $\cG$ is general-sum,  with the following information-sharing structures {and time complexities}, where we recall $\gamma$ is the constant  in Assumption \ref{observa}:
 \begin{itemize}
    \item \textbf{One-step delayed information sharing:}    $(AO)^{C\gamma^{-4}\log\frac{SH}{\epsilon}}$ for some universal constant $C>0$.
    \item \textbf{State controlled by one controller with asymmetric $d=\operatorname{poly}(\log H)$-step delayed sharing sharing:} $(AO)^{C(\gamma^{-4}\log\frac{SH}{\epsilon} + d)}$ for some constant $C>0$.
        \item \textbf{Information sharing with one-directional-one-step delay:}\\ $(AO)^{C\gamma^{-4}\log\frac{SH}{\epsilon}}$ for some universal constant $C>0$.
    \item \textbf{Uncontrolled state process with $d=\operatorname{poly}(\log H)$-step delayed sharing:}  $(AO)^{C(\gamma^{-4}\log\frac{SH}{\epsilon} + d)}$ for some universal constant $C>0$.
    \item \textbf{Symmetric information game:}  $(AO)^{C\gamma^{-4}\log\frac{SH}{\epsilon}}$ for some universal constant $C>0$.
\end{itemize}
\end{theorem}

\subsection{Learning}\label{subsec:learn}
Here we state the full version of Theorem \ref{thm:main_thm_learning} regarding the sample efficiency of learning and approximate common information model.

\begin{theorem}\label{thm:main_learning_full}
	Suppose the POSG $\cG$ satisfies Assumptions \ref{evo} and  \ref{str_indi}. Given any compression functions of common information, $\operatorname{Compress}_h:\cC_h\rightarrow \hat{\cC}_h$ for $h\in [H+1]$, we can compute $\hat{L}$ as defined in Definition \ref{def:L_main}. 
    Then, given any $H$ policies $\pi^{1:H}$, where $\pi^h\in \Delta(\Pi^{\mathrm{det}})$,  $\pi^h_{h-\hat{L}:h}= \operatorname{Unif}(\cA)$ 
    for $h\in [H]$, we can construct a policy-dependent expected approximate common information model $\Tilde{\cM}(\pi^{1:H})$, whose compression functions are $\{\operatorname{Compress}_h\}_{h\in[H+1]}$. We write  $\epsilon_r(\pi^{1:H}):=\epsilon_r(\Tilde{\cM}(\pi^{1:H}))$ and $\epsilon_z(\pi^{1:H}):=\epsilon_z(\Tilde{\cM}(\pi^{1:H}))$ for short. 
    Fix some parameters $\delta_1, \theta_1, \theta_2, \zeta_1, \zeta_2>0$ for Algorithm \ref{alg:construct_mg}, $\epsilon_e>0$ for Algorithm \ref{alg:avi}, and $\phi>0$, define the approximation error for estimating $\Tilde{\cM}(\pi^{1:H})$ using samples under the policies $\pi^{1:H}$ as: 
    \begin{align} \label{equ:def_eps_apx}
    \epsilon_{apx}&(\pi^{1:H}, \hat{L}, \zeta_1, \zeta_2, \theta_1, \theta_2, \phi) = \theta_1 + 2A\max_h P_h\frac{\zeta_2}{\zeta_1} \\
    &+ A\max_h P_h\theta_2 + \frac{A^{2\hat{L}}O^{\hat{L}}\zeta_1}{\phi} + \max_h\max_{\pi\in \Pi^{\mathrm{det}}} \bm{1}[h>\hat{L}]\cdot 2\cdot d_{\cS, h-\hat{L}}^{{\pi, \cG }}\left(\cU_{\phi, h-\hat{L}}^\cG(\pi^h)\right),\nonumber
\end{align}
{where for any policy $\pi^\prime\in\Delta(\Pi^{\mathrm{det}}), h\in [H]$, we define $d_{\cS, h}^{\pi^\prime, \cG}(s) := \PP_{h}^{\pi^\prime, \cG}(s_h = s)$, $d_{\cS, h}^{\pi^\prime, \cG}(A) := \sum_{s\in A}d_{\cS, h}^{\pi^\prime, \cG}(s)$ for any $A\subseteq \cS$, $\cU_{\phi, h}^{\cG}(\pi^\prime):=\{s\in \cS: d_{\cS,h}^{{\pi^\prime, \cG}}(s)< \phi\}$, representing under-explored states under the policy $\pi^\prime$.}
Then, Algorithm \ref{alg:construct_mg} can learn an model $\hat{\cM}(\pi^{1:H})$ with the sample complexity  
    \begin{equation}\label{eq:n0}
    N_0= \max \left\{\frac{C(\max_h P_h + \log\frac{4H \max_h\hat{C}_h}{\delta_1})}{\zeta_1 \theta_1^2}, \frac{CA(O + \log\frac{4H\max_h(\hat{C}_hP_h)A}{\delta_1})}{\zeta_2\theta_2^2}\right\},
    \end{equation} 
    for some universal constant $C>0$, such that with probability at least $1-\delta_1$, for any policy $\pi\in\Pi$, and $i\in [n]$:
    \$
   \Big|V_{i, 1}^{\pi, \cG}(\emptyset) - V_{i, 1}^{\pi, \hat{\cM}(\pi^{1:H})}(\emptyset)\Big|\leq  H\cdot\epsilon_r(\pi^{1:H}) + \frac{H^2}{2}\epsilon_z(\pi^{1:H}) + \left(\frac{H^2}{2} + H\right)\epsilon_{apx}(\pi^{1:H}, \hat{L}, \zeta_1, \zeta_2, \theta_1, \theta_2, \phi).
    \$
    
    Under such a high probability event, the policy output of Algorithm \ref{alg:avi} on $\hat{\cM}(\pi^{1:H})$ {is an $\epsilon$-NE if $\cG$ zero-sum or cooperative and $\epsilon$-CE/CCE if $\cG$ is general-sum, where 
 \[
 \epsilon:=H\epsilon_r(\pi^{1:H}, \hat{r}) +  H^2\epsilon_z(\pi^{1:H}) + (H^2+H)\epsilon_{apx}(\pi^{1:H}, \hat{L}, \zeta_1, \zeta_2, \theta_1, \theta_2, \phi) + H\epsilon_e. 
 \]}
\end{theorem}

We state the full version of Theorem \ref{thm:learning_example_short} regarding the instantiation of Theorem \ref{thm:main_thm_learning} in the following. 

\begin{theorem}
{Fix $\epsilon, \delta>0$. Suppose the POSG $\cG$ satisfies Assumptions \ref{evo} and \ref{str_indi}. If there exist some  compression functions of common information, $\operatorname{Compress}_h:\cC_h\rightarrow \hat{\cC}_h$ for $h\in [H+1]$, $\pi^{1:H}$, and $\Tilde{\cM}(\pi^{1:H})$ satisfying the conditions in Theorem \ref{thm:main_thm_learning}, and there exists 
some parameters $\delta_1, \theta_1, \theta_2, \zeta_1, \zeta_2>0$ for Algorithm \ref{alg:construct_mg}, $\epsilon_e>0$ for Algorithm \ref{alg:avi}, and some $\phi>0$, such that 
\[
\max\left\{\epsilon_z(\pi^{1:H}), \epsilon_r(\pi^{1:H}), \epsilon_{apx}(\pi^{1:H}, \hat{L}, \zeta_1, \zeta_2, \theta_1, \theta_2, \phi)\right\}\le \cO(\epsilon)
\]
and $N_0 = \operatorname{poly}(\max_{h\in [H]} P_{h}, \max_{h\in [H]} \hat{C}_{h}, H, A, O, \frac{1}{\zeta_1}, \frac{1}{\zeta_2}, \frac{1}{\theta_1}, \frac{1}{\theta_2})\cdot\log\frac{1}{\delta_1}$ is quasi-polynomial of the problem instance size, then Algorithm \ref{alg:construct_mg}, {together with Algorithm \ref{alg:avi}}, can 
output an {$\epsilon$-NE if $\cG$ is zero-sum or cooperative, and an $\epsilon$-CE/CCE if $\cG$ is general-sum}, with probability at least $1-\delta$, using quasi-polynomial time and samples, where $\hat{L}$ is defined as in Definition \ref{def:L_main}.}

In particular, under Assumption \ref{observa}, examples in \Cref{sec:example} satisfy  such all such conditions. 
Then, there exists a {multi-agent RL} algorithm (Algorithm \ref{alg:learning}) {that,  with probability at least $1-\delta$, learns an} $\epsilon$-NE if $\cG$ is zero-sum or cooperative, and $\epsilon$-CE/CCE if $\cG$ is general-sum, with the following information-sharing structures {and corresponding sample and time complexities:}
\begin{itemize}
    \item \textbf{One-step delayed information sharing:} $(AO)^{C\gamma^{-4}\log\frac{SHO}{\gamma\epsilon}}\log\frac{1}{\delta}$ for some universal constant $C>0$.
    {\item \textbf{State controlled by one controller with asymmetric $d=\operatorname{poly}(\log H)$-step delayed sharing sharing:} $(AO)^{C(\gamma^{-4}\log\frac{SHO}{\gamma\epsilon} + d)}\log\frac{1}{\delta}$ for some constant $C>0$.}
    \item \textbf{Information sharing with one-directional-one-step delay:} \\$(AO)^{C\gamma^{-4}\log\frac{SHO}{\gamma\epsilon}}\log\frac{1}{\delta}$ for some universal constant $C>0$.
    \item \textbf{Uncontrolled state process with $d=\operatorname{poly}(\log H)$-step delayed sharing:} $(AO)^{C(\gamma^{-4}\log\frac{SHO}{\gamma\epsilon} + d)}\log\frac{1}{\delta}$ for some universal constant $C>0$. 
    \item \textbf{Symmetric information game:} $(AO)^{C\gamma^{-4}\log\frac{SHO}{\gamma\epsilon}}\log\frac{1}{\delta}$ for some universal constant $C>0$.
\end{itemize}
\end{theorem}

\section{Technical Details and Omitted Proofs}

\subsection{Missing details in \Cref{subsec:hardness_main_body}}\label{subsec:hardness_proof}
Before proving \Cref{posg_hardness_1}, we present some hardness results for solving the stronger solution concepts of team-optimal policy in Dec-POMDPs to further justify the necessity of some favorable information-sharing structures. 
\begin{proposition}\label{prop1}
With 1-step delayed information-sharing structure and {A}ssumption \ref{observa}, computing the team optimal policy in Dec-POMDPs with $n=2$ is \texttt{NP-hard}.
\end{proposition}

To prove Proposition \ref{prop1}, we will firstly consider Dec-POMDPs with $H=1$ and then connect the 1-step Dec-POMDP with Dec-POMDPs that have \emph{1-step delayed sharing}. We will show the reduction from \textit{Team Decision Problem} \citep{tsitsiklis1985complexity}:
\begin{problem}[Team decision problem]\label{prob:team}
Given finite sets $\cY_1$, $\cY_2$, $\cU_1$, $\cU_2$, a rational probability function $p: \cY_1\times\cY_2\rightarrow \QQ$ 
	and an integer cost function $c:\cY_1\times\cY_2\times\cU_1\times\cU_2\rightarrow \NN$, find decision rules $\gamma_i: \cY_i\rightarrow\cU_i, i=1,2$, which minimize the expected cost:
\[
    J(\gamma_1, \gamma_2)=\sum_{y_1\in\cY_1}\sum_{y_2\in \cY_2}c(y_1, y_2, \gamma_1(y_1), \gamma_2(y_2))p(y_1, y_2).
\]
\end{problem}

\begin{proposition}\label{one-step}
Without any information sharing, computing jointly team optimal policies in Dec-POMDP with $H=1$, $n=2$ is \texttt{NP-hard}.
\end{proposition}
\begin{proof}
We can notice that the team decision problem is quite similar to our two-agent one-step Dec-POMDP. The only difference in Dec-POMDP is that the joint observations are sampled given the initial state, which is again sampled from $\mu_1$. Now we will show how to reduce the team decision problem to a Dec-POMDP. To begin with, we define $c_{\max}=\max_{y_1, y_2, u_1, u_2} c(y_1, y_2, u_1, u_2)$. For any team decision problem, we can construct the following Dec-POMDP:
\begin{itemize}
    \item $\cA_i = \cU_i, i=1, 2$;
    \item $\cO_i = \cY_i, i=1, 2$;
    \item $\cS = \cO_1\times\cO_2$.
    \item $\OO(o_{1, h}, o_{2, h}\given s_h) = 1$ if $s_h=(o_{1, h}, o_{2, h})$, else $0$, for $h\in\{1, 2\}$; 
    \item $r_1(s_1, a_1) = 1-c(y_1, y_2, u_1, u_2)/c_{\max}$, where $s_1=(y_1, y_2)$;
    \item $\mu_1(s_1) = p(y_1, y_2)$, where $s_1=(y_1, y_2)$.
\end{itemize}
Based on the construction, computing the optimal policies $\{\pi_{1, 1}^{\star}, \pi_{2, 1}^{\star}\}$ under the no-information-sharing structure in the reduced Dec-POMDP problem will give us the optimal policies $\{\gamma_1^{\star}, \gamma_2^{\star}\}$ in the original team decision problem. Concretely, we can construct the optimal policy for the team decision problem as $\gamma_{i}^{\star}(y_i) = \pi_{i, 1}^\star(o_{i, 1})$, where $o_{i, 1} = y_i$. Given the \texttt{NP-hardness} of the team decision problem shown in \cite{tsitsiklis1985complexity}, solving this corresponding Dec-POMDP without information sharing is also \texttt{NP-hard}.
\end{proof}

This result directly implies the hardness of Dec-POMDPs with 1-step delayed sharing structure:
\begin{proposition}\label{one-delay}
With 1-step delayed information-sharing structure, computing joint{ly team} optimal policies in Dec-POMDPs with $n=2$ is at least \texttt{NP-hard}.
\end{proposition}
\begin{proof}
Since there exists 1-step delay for the common information to be shared, when the Dec-POMDPs have only 1-step, there is no shared common information among agents. Therefore, based on the proof of  Proposition \ref{one-step}{, which concerns exactly such a case, }
computing joint optimal policies in Dec-POMDPs with $n=2$ is also at least \texttt{NP-hard}.
\end{proof}
Finally, we are ready to prove Proposition \ref{prop1}. 

\begin{prevproof}{prop1}
Similar to the proof of  Proposition \ref{one-delay}, it suffices to show that the proposition holds for Dec-POMDPs, with $H=1$ and without information sharing. Note that in the proof of Proposition  \ref{one-step}, the constructed Dec-POMDPs have the state space defined as the joint observation space (the Cartesian product of the individual observation spaces), and the observation emission is actually a one-to-one mapping from state space to joint observation space. Correspondingly, $\OO_h$ is indeed an identity matrix. Therefore, we have $\left\|\mathbb{O}_{h}^{\top} b-\mathbb{O}_{h}^{\top} b^{\prime}\right\|_{1} = \left\|b-b^{\prime}\right\|_{1}$, for any $b, b^\prime\in\Delta(\cS)$, verifying that $\gamma=1$.
\end{prevproof}
Now, let us restate and prove our hardness results regarding NE/CE/CCE in Proposition \ref{posg_hardness_1} as the following two propositions.
\begin{proposition}\label{posg_hardness_8}
For zero-sum or cooperative POSGs with any kind of information-sharing  structure {(including the fully-sharing   structure)}, computing $\epsilon$-NE/CE/CCE is \texttt{PSPACE-hard}. 
\end{proposition}
\begin{proof}
The proof leverages the known results of the hardness of solving POMDPs. Given any instance of POMDPs, one could add a dummy agent with only one dummy observation and one available action, which does not affect the transition, and use any desired information-sharing strategy. Since this dummy agent only has one action and therefore it has only one policy. And the reward could be identical to the original agent for cooperative games or the opposite of that for zero-sum games. Therefore, $\epsilon$-NE/CE/CCE in this constructed POSG with the desired information-sharing strategy gives the $\epsilon$-optimal policy in the original POMDP. Given the known \texttt{PSPACE-hardness} of POMDPs  \citep{papadimitriou1987complexity,lusena2001nonapproximability}, we conclude our proof.
\end{proof}
\begin{proposition}
For zero-sum or cooperative POSGs satisfying Assumption \ref{observa} {without information sharing}, computing $\epsilon$-NE/CE/CCE is \texttt{PSPACE-hard}. 
\end{proposition}
\begin{proof}
Similar to the proof of Proposition \ref{posg_hardness_8}, given any instance of a POMDP, we could add a dummy agent with only one available action, and the observation of the dummy agent is exactly the underlying state. Formally, given an instance of POMDP $\cP = (\cS^{\cP}, \cA^\cP, \cO^\cP, \{\OO^\cP_h\}_{h\in[H+1]}, \{\TT_h^\cP\}_{h\in [H]}, r^{\cP})$, we construct the POSG $\cG$ as follows:
\begin{itemize}
	\item $\cS = \cS^\cP$;
	\item $\cA_1 = \cA^\cP$, and $\cA_2 = \{\emptyset\}$;
	\item $\cO_1 = \cO^\cP$, and $\cO_2 = \cS^\cP$;
	\item For any $h\in [H+1]$, $o_{1, h}\in\cO_1$, $o_{2,h}\in\cO_2$, $s_h\in\cS$, it holds that 
	 \[
	 \OO_{h}(o_{1, h}, o_{2, h}\given s_h)=\begin{cases}\OO_h^\cP(o_{1, h}\given s_h) & \text { if } o_{2, h}=s_h \\ 0 & \text { otherwise }\end{cases};
	 \]
	\item For any $h\in [H]$, $a_{1,h}\in\cA_1$, $a_{2, h}\in\cA_2$, $s_{h},s_{h+1}\in\cS$, it holds that $\TT_{h}(s_{h+1}\given s_h, a_{1, h}, a_{2, h}) = \TT_{h}^\cP(s_{h+1}\given s_h, a_{1, h})$; 
	\item For the reward, we use the reward from the original POMDP.
\end{itemize}
Now we are ready to verify that the joint observation emission satisfies Assumption \ref{observa} with $\gamma=1$. Consider any $b, b^\prime\in\Delta(\cS)$, denote $b-b^\prime = (\delta_s)_{s\in\cS}^\top$ as the column vector. For any $h\in[H+1]$, it holds that 
	\$
	\|\OO_{h}^\top(b-b^\prime)\|_1 &= \sum_{o_{1, h}, o_{2, h}}\Big|\sum_{s\in\cS}\OO_{h}(o_{1, h}, o_{2, h}\given s)\delta_s\Big| = \sum_{o_{1, h}, s}|\OO^{\cP}_h(o_{1, h}\given s)\delta_s|= \sum_{s}| \delta_{s}| = \|b-b^\prime\|_1,
	\$
	which verifies that $\gamma = 1$ for our constructed POSG.
 Computing $\epsilon$-NE/CE/CCE in such a 1-observable POSG immediately gives us the $\epsilon$-optimal policy in the original POMDP. Furthermore, note that $\gamma\le 1$ for any possible emission, therefore, the conclusion also holds for any $\gamma$-observable POSG, which proves our conclusion. 
\end{proof}

Finally, we provide the proof for Lemma \ref{lemma:cardi} regarding usually how large $C_hP_h$ is. 

\begin{prevproof}{lemma:cardi}
	Fix any $h\in [H+1]$. If each agent has perfect recall, then it holds that for any joint history $\{o_1, a_1, o_2, \cdots, a_{h-1}, o_h\}\in \cO^{h}\times \cA^{h-1}$, there exists some 
$c_h\in \cC_h$ and $p_h\in \cP_h$ such that $\{c_h, p_h\} = \{o_1, a_1, o_2, \cdots, a_{h-1}, o_h\}$, {which can be found by the functions $f_h$ and $g_h$ introduced after Assumption \ref{evo}}.  Therefore, we conclude that $\cO^{h}\times \cA^{h-1}\subseteq\cC_h\times\cP_h$, implying that $C_hP_h\ge (OA)^{h-1}$.
\end{prevproof}

\subsection{Missing details in \Cref{subsec:plan-exact}}\label{subsec:subroutine}
Similar to the value iteration algorithm in Markov games \citep{shapley1953stochastic}, which solves a normal-form game at each step, we utilize a similar value iteration framework. 
 Specifically, under Assumption \ref{str_indi}, we can have the Bellman equation as follows
\$
V^{{\pi},  \cG }_{i, h}(c_h) = \EE_{\{\omega_{j,h}\}_{j\in[n]}}\EE_{\substack{s_h, p_{h}\sim \PP_h^ \cG\left(\cdot, \cdot\given c_h\right)}}\EE_{\substack{\{a_{j, h}\sim\pi_{j, h}(\cdot\given\omega_{j,h}, c_h, p_{j, h})\}_{j\in [n]}\\o_{h+1}\sim \OO_{h+1}^{\top}\TT_{h}(\cdot\given s_h, a_h)}}\Big[r_{i, h}(s_h, a_h) + V^{{\pi},  \cG }_{i, h+1}(c_{h+1})\Big].
\$

With Assumption \ref{str_indi}, we are ready to present our Algorithm \ref{alg:vi} based on value iteration in the common information space, which runs in a backward way, enumerating all possible $c_h$ at each step $h$ and computing the corresponding equilibrium in the prescription space.
\paragraph{Implementing the equilibrium subroutine at each step.}
Now we will discuss the three equilibrium or best response (BR) subroutines at each step $h\in[H]$, where NE or NE-BR is used for zero-sum or cooperative games, and CE/CCE (or CE/CCE-BR) is used for general-sum games for computational  tractability. To find efficient implementation for these subroutines, we need the following important properties on the prescription-value function. 

\begin{proposition}\label{prop:linear}
$Q^{\star,  \cG}_{i, h}(c_h, \gamma_{1, h}, \cdots, \gamma_{n, h})$ defined in Algorithm \ref{alg:vi} is linear with respect to each $\gamma_{i, h}$. More specifically, we have:
\#\label{grad}
    \frac{\partial Q^{\star,  \cG}_{i, h}(c_h, \gamma_{1, h}, \cdots, \gamma_{n, h})}{\partial \gamma_{i, h}(a_{i, h}\given p_{i, h})}& =  
    \sum_{s_h^{\prime}, p^{\prime}_{-i, h}}\sum_{a^{\prime}_{-i, h}}\PP_h^ \cG(s^{\prime}_h, p_{i, h}, p^{\prime}_{-i, h}\given c_h)\gamma_{-i, h}(a^{\prime}_{-i, h}\given p^{\prime}_{-i, h})\\ 
    &\qquad\quad\times\left( \sum_{o_{h+1}, s_{h+1}^\prime}\OO_{h+1}(o_{h+1}|s_{h+1}^\prime)\TT_{h}(s_{h+1}^\prime|s_h^{\prime}, a_h)\left[r_{i, h}(s_h, a_h) + V^{\star,  \cG}_{i, h+1}(c_{h+1})\right]\right).\nonumber 
\#
\end{proposition}
\begin{proof}

The partial derivative can be easily verified by  algebraic manipulations and the definition of $Q_{i, h}^{\star, \cG}$. From Equation \eqref{grad}, we could notice  that $\gamma_{i, h}$ does not appear on the RHS, which proves $Q^{\star,  \cG}_{i, h}(c_h, \gamma_{1, h}, \cdots, \gamma_{n, h})$ is linear with respect to $\gamma_{i, h}$.
\end{proof}
{With such kind of linear structures, we are ready to introduce how to implement those oracles efficiently.}
\begin{itemize}
    \item The NE subroutine will give us the approximate NE $\{\gamma^{\star}_{1, h}, \cdots, \gamma^{\star}_{n, h}\}$ up to some error $\epsilon_e$, which satisfies:
\[
    Q^{\star,  \cG}_{i, h}(c_h, \gamma^{\star}_{i, h}, \gamma^{\star}_{-i, h}) \ge \max_{\gamma_{i, h}\in \Delta(\cA_i)^{P_{i, h}}}Q^{\star,  \cG}_{i, h}(c_h, \gamma_{i, h}, \gamma^{\star}_{-i, h}) - \epsilon_e, \qquad\forall i\in [n].
\]
This NE subroutine will be intractable for general-sum games even with only two agents  \citep{daskalakis2009complexity,chen2009settling}. However, for cooperative games and zero-sum games, this NE subroutine can be implemented efficiently. At first look, this can be done by formulating it as a normal-form game, where each agent has the corresponding action space $\cA_{i}^{P_{i, h}}$. However, this could not be tractable since the action space is indeed exponentially large. Fortunately, for cooperative games and two-agent  zero-sum games, we could utilize the linear (concave) structure, {where $\gamma_{i, h}$ is a vector of dimension $A_iP_{i, h}$} to develop an efficient algorithm to compute $\epsilon_e$-NE using standard no-external-regret  {or specifically gradient-play} algorithms \citep{daskalakis2011near, zhang2021gradient,leonardos2022global,ding2022independent,mao2022improving}, which will run in $\texttt{poly}(S, A, P_h, \frac{1}{\epsilon_e})$ time. To further illustrate how we avoid the dependence of $\cA_{i}^{P_{i, h}}$, we refer to Figure \ref{fig:framework}. Similarly, the best response (BR) subroutine for NE, denoted as the NE-BR 
	 subroutine, is defined as follows: it  outputs the approximate best response $\gamma_{i, h}^\star$ for the $i^{th}$ agent given $\{\gamma_{j, h}\}_{j\in [n]}$ up to some error $\epsilon_e$, which satisfies:
	\[
	Q^{\star,  \cG}_{i, h}(c_h, \gamma^{\star}_{i, h}, \gamma_{-i, h}) \ge \max_{\gamma_{i, h}^\prime\in \Delta(\cA_i)^{P_{i, h}}}Q^{\star,  \cG}_{i, h}(c_h, \gamma_{i, h}^\prime, \gamma_{-i, h}) - \epsilon_e.
	\]
	Its implementation is straightforward  by linear programming since $Q_{i, h}^{\star, \cG}$ is linear with respect to each agent's  prescription.
    \item The CCE subroutine will give us the approximate CCE, {a uniform mixture of} $\{\gamma^{\star, t}_{1, h}, \cdots, \gamma^{\star, t}_{n, h}\}_{t=1}^T$  up to some error $\epsilon_e$, which satisfy for any $i\in[n]$:
\[
    \frac{1}{T}\sum_{t=1}^T Q^{\star,  \cG}_{i, h}(c_h, \gamma^{\star, t}_{i, h}, \gamma^{\star, t}_{-i, h}) \ge \max_{\gamma_{i, h}\in \Delta(\cA_i)^{P_{i, h}}}\frac{1}{T}\sum_{t=1}^T Q^{\star,  \cG}_{i, h}(c_h, \gamma_{i, h}, \gamma^{\star, t}_{-i, h}) - \epsilon_e.
\]
This subroutine can be implemented using standard no-external-regret learning algorithm as in \cite{gordon2008no,farina2022near} with $\texttt{poly}(S, A, P_h, \frac{1}{\epsilon_e})$ time.

	Similarly, the CCE-BR subroutine    can be defined as follows: it  outputs the best response $\gamma_{i, h}^\star$ of the $i^{th}$ agent, given $\{\gamma_{1, h}^t, \cdots, \gamma_{n, h}^t\}_{t=1}^{T}$ up to some error $\epsilon_e$, which satisfies:
	\[
	\frac{1}{T}\sum_{t=1}^T Q^{\star,  \cG}_{i, h}(c_h, \gamma^{\star}_{i, h}, \gamma^{t}_{-i, h}) \ge \max_{\gamma_{i, h}^\prime\in \Delta(\cA_i)^{P_{i, h}}}\frac{1}{T}\sum_{t=1}^T Q^{\star,  \cG}_{i, h}(c_h, \gamma_{i, h}^\prime, \gamma^{t}_{-i, h}) - \epsilon_e.
	\]
	The implementation of CCE-BR is the same as CCE except that only the $i^{th}$ agent runs the no-external-regret algorithm and other agents remain fixed. Once we get the sequence $\{\gamma_{i, h}^{\star, t}\}_{t=1}^T$ from the no-external-regret algorithm, we can take $\gamma_{i, h}^\star = \frac{1}{T}\sum_{t=1}^T\gamma_{i, h}^{\star, t}$ since $Q_{i, h}^{\star, \cG}$ is linear with respect to each agent's prescription. 
    \item The CE subroutine  will give us the approximate CE $\{\gamma^{\star, t}_{1, h}, \cdots, \gamma^{\star, t}_{n, h}\}_{t=1}^T$ up to some error $\epsilon_e$, which satisfy for any $i\in[n]$:
\[
    \frac{1}{T}\sum_{t=1}^T Q^{\star,  \cG}_{i, h}(c_h, \gamma^{\star, t}_{i, h}, \gamma^{\star, t}_{-i, h}) \ge \max_{u_{i, h}}\frac{1}{T}\sum_{t=1}^T Q^{\star,  \cG}_{i, h}(c_h, u_{i, h}\diamond \gamma_{i, h}^{\star, t}, \gamma^{\star, t}_{-i, h}) - \epsilon_e.
\]
Here $u_{i, h}=\{u_{i, h, p_{i, h}}\}_{p_{i, h}}$ is the strategy modification, where $u_{i, h, p_{i, h}}: \cA_i\rightarrow \cA_i$ will modify the action $a_{i, h}$ to $u_{i, h, p_{i, h}}(a_{i, h})$ given the private information $p_{i, h}$. It is easy to see that the composition of $u_{i, h}$ with any prescription $\gamma_{i, h}$ is equivalent to $(u_{i, h}\diamond \gamma_{i, h})(a_{i, h}\given p_{i, h}):=\sum_{u_{i, h, p_{i, h}}(a_{i, h}^\prime) = a_{i, h}} \gamma_{i, h}(a_{i, h}^\prime\given p_{i, h})$. One can verify that $u_{i, h}\diamond \gamma_{i, h} = U\cdot \gamma_{i, h}$, for some matrix $U\in \RR^{A_{i}P_{i, h}\times A_{i}P_{i, h}}$ (in a block diagonal form). Therefore, the composition of $u_{i, h}$ and $\gamma_{i, h}$ is indeed a linear transformation. Now, as 
the function $Q_{i, h}^\star(c_{h}, \gamma_{1, h}, \cdots, \gamma_{n, h})$ is concave (in fact, linear) with respect to each $\gamma_{i, h}$, one can run the \emph{no-linear-regret} algorithm as in \cite{gordon2008no}, such that the time-averaged policy will give us the approximate CE. In particular, such a guarantee can be achieved by running the swap-regret minimization algorithm in \cite{blum2007learning} separately for each $(i, p_{i, h})$ and the corresponding time complexity will be of $\texttt{poly}(S, A, P_h, \frac{1}{\epsilon_e})$.

	The CE-BR subroutine can be defined as follows:  it will output the best strategy modification $u_{i, h}^\star$ of the $i^{th}$ agent,  given $\{\gamma^{t}_{1, h}, \cdots, \gamma^{t}_{n, h}\}_{t=1}^T$ up to some error $\epsilon_e$, which satisfies:
	\[
	\frac{1}{T}\sum_{t=1}^T Q^{\star,  \cG}_{i, h}(c_h, u_{i, h}^\star\diamond\gamma^{t}_{i, h}, \gamma^{t}_{-i, h}) \ge \max_{u_{i, h}}\frac{1}{T}\sum_{t=1}^T Q^{\star,  \cG}_{i, h}(c_h, u_{i, h}\diamond \gamma_{i, h}^{t}, \gamma^{t}_{-i, h}) - \epsilon_e.
	\]
	For notational convenience, we shall slightly abuse the notation, writing $\gamma_{i, h}^{\star, t}:=u_{i, h}^{\star}\diamond \gamma_{i, h}^t$ for any $t\in [T]$ and we assume our CE-BR subroutine returns $\{u_{i, h}^\star\diamond \gamma_{i, h}^t\}_{t\in[T]}$ instead of $u_{i, h}^\star$. Its implementation still follows from that of CE except that only the agent  $i$ runs the \emph{no-linear-regret} algorithm.
\end{itemize}

\subsection{Proof of \Cref{thm:struc}}

To prove Theorem \ref{thm:planning_ind_common_belief}, we prove our main theorem, Theorem \ref{thm:struc}, which is a generalized version. We will first bound the sub-optimality of the planning algorithm on $\cM$ at each step $h$ through the following two lemmas.
\begin{lemma}\label{lemma:ne_err}
Fix the input $\cM$ and $\epsilon_e>0$ for Algorithm \ref{alg:avi}. For any {$h\in [H+1]$, $c_h\in\cC_h$},  and $\pi_i\in \Pi_i$, for computing approximate  NE/CCE, the output of Algorithm \ref{alg:avi}, $\hat{\pi}^\star$, satisfies that
\[
    {V}^{\pi_i\times\hat{\pi}_{-i}^{\star}, \cM   }_{i, h}(c_h)\le {V}_{i, h}^{\hat{\pi}^\star, \cM   }({c}_h) + (H+1-h)\epsilon_e.
\]
\end{lemma}

\begin{proof}	
Obviously, the proposition holds for $h=H+1$. Note that $\pi_i$ does not share the randomness with $\hat{\pi}_{-i}^\star$. In other words, the following $\omega_{i, h}^\prime$ is independent of $\omega_{-i, h}$. Then, we have that
\begin{align}
    &{V}^{\pi_i\times\hat{\pi}_{-i}^{\star}, \cM}_{i, h}(c_h)\nonumber\\
    &= \EE^\cM_{\omega_{i, h}^\prime, \{\omega_{j, h}\}_{j\in [n]}}[\hat{r}_{i, h}^\cM + {V}^{\pi_i\times\hat{\pi}_{-i}^{\star}, \cM }_{i, h+1}({c}_{h+1})\mid \hat{c}_h, \{\pi_{i, h}(\cdot\given\omega_{i, h}^\prime, c_h, \cdot), \hat{\pi}_{-i, h}^\star(\cdot\given\omega_{-i, h}, \hat{c}_h, \cdot)\}]\nonumber\\
    &\le \EE^\cM_{\omega_{i, h}^\prime, \{\omega_{j, h}\}_{j\in [n]}}[\hat{r}_{i, h}^\cM + {V}^{\hat{\pi}^\star, \cM}_{i, h+1}({c}_{h+1})\mid \hat{c}_h, \{\pi_{i, h}(\cdot\given\omega_{i, h}^\prime, c_h, \cdot), \hat{\pi}_{-i, h}^\star(\cdot\given\omega_{-i, h}, \hat{c}_h, \cdot)\}] \label{induct}\\
    &\qquad + (H-h)\epsilon_e\nonumber\\
    \nonumber&=\EE_{\omega_{i, h}^\prime}\EE_{\{\omega_{j, h}\}_{j\in [n]}}Q_{i, h}^{\hat{\pi}_i^\star\times\hat{\pi}_{-i}^\star, \cM}(c_h, \pi_{i, h}(\cdot\given\omega_{i, h}^\prime, c_h, \cdot), \hat{\pi}_{-i, h}^{\star}(\cdot\given\omega_{-i, h}, \hat{c}_h, \cdot)) + (H-h)\epsilon_e\\
    \label{BNE}&\le \EE_{\omega_{i, h}^\prime}\EE_{\{\omega_{j, h}\}_{j\in [n]}}Q_{i, h}^{\hat{\pi}_i^\star\times\hat{\pi}_{-i}^\star, \cM}(c_h, \hat{\pi}^\star_{i, h}(\cdot\given\omega_{i, h}, c_h, \cdot), \hat{\pi}_{-i, h}^{\star}(\cdot\given\omega_{-i, h}, \hat{c}_h, \cdot))  + (H-h+1)\epsilon_e\\
    \nonumber&={V}_{i, h}^{\hat{\pi}^\star, \cM}({c}_h) + (H-h+1)\epsilon_e,
\end{align}
where \Cref{induct} comes from the inductive hypothesis,  Equation \eqref{BNE} holds since $\hat{\pi}^{\star}_h(\cdot\mid \cdot, \hat{c}_h,\cdot)$ is an $\epsilon_e$-NE/CCE for the stage game and ${V}_{i, h+1}^{\hat{\pi}^\star, \cM }({c}_{h+1}) ={V}_{i, h+1}^{\star, \cM   }(\hat{c}_{h+1})$ through a simple induction argument  using Definition \ref{def:M_value_func}.
\end{proof}

\begin{corollary}\label{corr:ne_br}
	Fix the input $\cM$, $i\in[n]$, $\hat{\pi}$ whose $\hat{\pi}_{j, h}: \Omega_h\times\cP_{j, h}\times\hat{\cC}_h\rightarrow\Delta(\cA_j)$ for $j\in [n]$  takes only approximate common information instead of the exact common information as the input, and $\epsilon_e>0$ for Algorithm \ref{alg:abr}. For any {$h\in [H+1]$, $c_h\in\cC_h$},  and $\pi_i\in \Pi_i$, the output of Algorithm \ref{alg:abr}, $\hat{\pi}_i^\star$ satisfies that
	\[
    {V}^{\pi_i\times\hat{\pi}_{-i}, \cM}_{i, h}(c_h)\le {V}_{i, h}^{\hat{\pi}_i^\star\times\hat{\pi}_{-i}, \cM   }({c}_h) + (H+1-h)\epsilon_e.
	\]
\end{corollary}
\begin{proof}
	Since Algorithm \ref{alg:abr} simply replaces the equilibrium oracle at each stage of Algorithm \ref{alg:avi} by a best-response oracle, its proof follows directly from the proof of Lemma \ref{lemma:ne_err}.
\end{proof}

\begin{lemma}\label{lemma:ce_err}
For any $h\in [H+1]$,  $c_h\in\cC_h$, and $\phi_i\in \Phi_i$, for computing {approximate} CE, the output of Algorithm \ref{alg:avi}, $\hat{\pi}^\star$ satisfies that
\[
    {V}^{(\phi_i\diamond\hat{\pi}_i^\star)\odot\hat{\pi}_{-i}^{\star}, \cM   }_{i, h}(c_h)\le {V}_{i, h}^{\hat{\pi}^\star, \cM  }({c}_h) + (H-h+1)\epsilon_e.
\]
\end{lemma}
\begin{proof}
It is direct to see that the lemma holds for the step $H+1$. For step $h\in[H]$, it holds that 
 \begin{align}
    &{V}^{(\phi_i\diamond\hat{\pi}_i^\star)\odot\hat{\pi}_{-i}^{\star}, \cM   }_{i, h}(c_h) 
    = \EE^\cM_{\{\omega_{j, h}\}_{j\in [n]}}\Big[\hat{r}_{i, h}^\cM\nonumber + {V}^{(\phi_i\diamond\hat{\pi}_i^\star)\odot \hat{\pi}_{-i}^{\star}, \cM   }_{i, h+1}({c}_{h+1})\mid \hat{c}_h, \{\phi_{i, h, c_h}\diamond\hat{\pi}^\star_{i, h}(\cdot\given\omega_{i, h}, \hat{c}_h, \cdot), \hat{\pi}_{-i, h}^\star(\cdot\given\omega_{-i, h}, \hat{c}_h, \cdot)\}\Big]\nonumber \\
    \label{induct_2}&\quad\le \EE^\cM_{\{\omega_{j, h}\}_{j\in [n]}}\Big[\hat{r}_{i, h}^\cM + {V}^{\hat{\pi}^\star, \cM}_{i, h+1}({c}_{h+1})\mid \hat{c}_h, \{\phi_{i, h, c_h}\diamond\hat{\pi}^\star_{i, h}(\cdot\given\omega_{i, h}, \hat{c}_h, \cdot), \hat{\pi}_{-i, h}^\star(\cdot\given\omega_{-i, h}, \hat{c}_h, \cdot)\Big]  + (H-h)\epsilon_e \\
    \label{BNE_2}&\quad\le \EE^\cM_{\{\omega_{j, h}\}_{j\in [n]}}\Big[\hat{r}_{i, h}^\cM + {V}^{\hat{\pi}^\star, \cM}_{i, h+1}({c}_{h+1})\mid \hat{c}_h, \{\hat{\pi}_{i, h}^\star(\cdot\given\omega_{i, h}, \hat{c}_h, \cdot), \hat{\pi}_{-i, h}^\star(\cdot\given\omega_{-i, h}, \hat{c}_h, \cdot)\}\Big]  + (H-h)\epsilon_e \\
    \nonumber&\quad={V}_{i, h}^{\hat{\pi}^\star, \cM}({c}_h) + (H-h+1)\epsilon_e, 
\end{align}
where Equation \eqref{induct_2} comes from the inductive hypothesis, Equation \eqref{BNE_2} holds since ${V}_{i, h+1}^{\hat{\pi}^\star, \cM   }({c}_{h+1}) ={V}_{i, h+1}^{\hat{\pi}^\star, \cM   }(\hat{c}_{h+1})$, and $\hat{\pi}^{\star}_h(\cdot\mid \cdot, \hat{c}_h,\cdot)$ is an $\epsilon_e$-CE for the stage game.
\end{proof}

\begin{corollary}\label{corr:ce_br}
	Fix the input $\cM$, $i\in[n]$, $\hat{\pi}$ whose $\hat{\pi}_{j, h}: \Omega_h\times\cP_{j, h}\times\hat{\cC}_h\rightarrow\Delta(\cA_j)$ for $j\in [n]$  takes only approximate common information instead of the exact common information as the input, and $\epsilon_e>0$ for Algorithm \ref{alg:abr}. For any {$h\in [H+1]$, $c_h\in\cC_h$},  and $\phi_i\in \Phi_i$, the output of Algorithm \ref{alg:abr}, $\hat{\pi}_i^\star$ satisfies that
	\[
    {V}^{(\phi_i\diamond\hat{\pi}_i)\odot\hat{\pi}_{-i}, \cM}_{i, h}(c_h)\le {V}_{i, h}^{\hat{\pi}_i^\star\odot\hat{\pi}_{-i}, \cM   }({c}_h) + (H+1-h)\epsilon_e.
	\]
\end{corollary}
\begin{proof}
	Similar to the proof of Corollary \ref{corr:ne_br}, its proof follows directly from Lemma \ref{lemma:ce_err}.
\end{proof}

Now we prove \Cref{lemma:v_diff}, showing the difference between the approximate value functions and true value functions under the same set of policies. 
\begin{prevproof}{lemma:v_diff}
	
Note that it suffices to consider any policy  $\pi^\prime\in\Pi^{\mathrm{det}}$ instead of $\pi^\prime\in\Delta(\Pi^{\mathrm{det}})$. Obviously, the proposition holds for $h=H+1$. For step $h\in[H]$, we have 
\$
    &\EE_{a_{1:h-1}, o_{1:h}\sim \pi^{\prime}}^\cG[|V^{\pi, \cG }_{i, h}(c_h)-{V}_{i, h}^{\pi, \cM}({c}_h)|]\\ 
    &\le \EE_{a_{1:h-1}, o_{1:h}\sim \pi^{\prime}}^\cG\Big[\Big|\EE_{\{\omega_{j, h}\}_{j\in [n]}}\EE^ \cG[r_{i, h}(s_h, a_h)\mid c_h, \{\pi_{j, h}(\cdot\given\omega_{j, h}, c_h, \cdot)\}_{j=1}^n] -\EE_{\{\omega_{j, h}\}_{j\in [n]}}\hat{r}_{i, h}^\cM(\hat{c}_h, \{\pi_{j, h}(\cdot\given\omega_{j, h}, c_h, \cdot)\}_{j=1}^n) \Big|\Big]\\
    &\quad+\EE_{a_{1:h-1}, o_{1:h}\sim \pi^{\prime}}^\cG
    \Big[\Big|\EE_{\{\omega_{j, h}\}_{j\in [n]}}\EE_{z_{h+1}\sim \PP_{h}^\cG(\cdot\given c_h, \{\pi_{j, h}(\cdot\given\omega_{j, h}, c_h, \cdot)\}_{j=1}^n)}[V_{i, h+1}^{\pi,  \cG }(\{c_h, z_{h+1}\})]\\
    &\qquad\quad-\EE_{\{\omega_{j, h}\}_{j\in [n]}}\EE_{z_{h+1}\sim \PP^{\cM, z}_{h}{(\cdot\given\hat{c}_h, \{\pi_{j, h}(\cdot\given\omega_{j, h}, c_h, \cdot)\}_{j=1}^n)}}[{V}_{i, h+1}^{\pi, \cM   }(\{c_h, z_{h+1}\})]\Big|\Big]\\
    &\le \epsilon_r+ (H-h)\EE_{a_{1:h-1}, o_{1:h}\sim \pi^\prime}^\cG\EE_{\{\omega_{j, h}\}_{j\in [n]}}\Big\|\PP_{h}^ \cG{(\cdot\given{c}_h, \{\pi_{j, h}(\cdot\given\omega_{j, h}, c_h, \cdot)\}_{j=1}^n)} - \PP_{h}^{\cM, z}{(\cdot\given\hat{c}_h, \{\pi_{j, h}(\cdot\given\omega_{j, h}, c_h, \cdot)\}_{j=1}^n)}\Big\|_1 
    \\&\qquad\quad+ \EE_{a_{1:h}, o_{1:h+1}\sim \bar{\pi}}^\cG\Big[\Big|{V}_{i, h+1}^{\pi, \cM   }(c_{h+1})-{V}_{i, h+1}^{\pi,  \cG }(c_{h+1})\Big|\Big]\\
    &\le\epsilon_r + (H-h)\epsilon_z + (H-h)\epsilon_r + \frac{(H-h)(H-h-1)}{2}\epsilon_z
    \\&\le (H-h+1)\epsilon_r + \frac{(H-h)(H-h+1)}{2}\epsilon_z,
\$
where $\bar{\pi}\in\Delta(\Pi^{\mathrm{det}})$ is the policy following $\pi^\prime$ from step $1$ to $h-1$ and $\pi$ from step $h$ to $H$, thus completing the proof. 
\end{prevproof}
Finally, we are ready to prove our main theorem, Theorem \ref{thm:struc}. Before that, we need to show that the equilibrium subroutine at each $h\in[H]$ in $\cM$ is also computationally tractable. Specifically, similar to \Cref{prop:linear}, we can show $Q^{\star, \cM}_{i, h}$ is also linear w.r.t. each $\gamma_{i, h}$ for $i\in[n]$. Hence, the algorithms developed to implement the equilibrium subroutine for $\cG$ in \Cref{subsec:subroutine} are directly applicable for $\cM$. 
\begin{proposition}\label{prop:linear-M} Given $\cM$ that is consistent with the approximate belief $\{\PP_h^{\cM, c}(s_h, p_h\given\hat{c}_h)\}_{h\in[H]}$, we have
$Q^{\star,  \cM}_{i, h}(\hat{c}_h, \gamma_{1, h}, \cdots, \gamma_{n, h})$ defined in Algorithm \ref{alg:avi} is linear with respect to each $\gamma_{i, h}$. More specifically, we have:
\#\label{grad-M}
    \frac{\partial Q^{\star,  \cM}_{i, h}(\hat{c}_h, \gamma_{1, h}, \cdots, \gamma_{n, h})}{\partial \gamma_{i, h}(a_{i, h}\given p_{i, h})} &=  
    \sum_{s_h^{\prime}, p^{\prime}_{-i, h}}\sum_{a^{\prime}_{-i, h}}\PP_h^{\cM, c}(s^{\prime}_h, p_{i, h}, p^{\prime}_{-i, h}\given \hat{c}_h)\gamma_{-i, h}(a^{\prime}_{-i, h}\given p^{\prime}_{-i, h})\\ 
    &\qquad\quad\times\left( \sum_{o_{h+1}, s_{h+1}^\prime}\OO_{h+1}(o_{h+1}|s_{h+1}^\prime)\TT_{h}(s_{h+1}^\prime|s_h^{\prime}, a_h)\left[r_{i, h}(s_h, a_h) + V^{\star,  \cM}_{i, h+1}(\hat{c}_{h+1})\right]\right).\nonumber 
\#
\end{proposition}

\begin{proof}

The partial derivative can be easily verified by  algebraic manipulations and the definition of $Q_{i, h}^{\star, \cM}$. From Equation \eqref{grad-M}, we could notice  that $\gamma_{i, h}$ does not appear on the RHS, which proves $Q^{\star,  \cM}_{i, h}(\hat{c}_h, \gamma_{1, h}, \cdots, \gamma_{n, h})$ is linear with respect to $\gamma_{i, h}$.
\end{proof}

\begin{prevproof}{thm:struc}
	
For computing NE/CCE, we define for each agent $i\in[n]$ 
\[
    \pi_{i}^{\star} \in \arg\max_{\pi_{i}\in \Pi_i}
    V^{\pi_i\times{\hat{\pi}^{\star}_{-i}}, \cG}_{i, 1}(\emptyset).
\]
Now note that
\begin{align*}
    &\EE_{a_{1:h-1}, o_{1:h}\sim \pi^{\prime}}[V^{\pi_i^{\star}\times\hat{\pi}_{-i}^{\star},  \cG }_{i, h}(c_h)-V^{\hat{\pi}^\star,  \cG }_{i, h}(c_h)]\\
    &=\EE_{a_{1:h-1}, o_{1:h}\sim \pi^{\prime}}\left[\left(V^{\pi_i^{\star}\times \hat{\pi}_{-i}^{\star},  \cG }_{i, h}(c_h)-{V}_{i, h}^{\hat{\pi}^\star, \cM   }({c}_h)\right)
    +\left({V}_{i, h}^{\hat{\pi}^\star, \cM   }({c}_h) -V^{\hat{\pi}^\star,  \cG }_{i, h}(c_h)\right)\right]\\
    &\le\EE_{a_{1:h-1}, o_{1:h}\sim \pi^{\prime}}\left[\left(V^{\pi_i^{\star}\times\hat{\pi}_{-i}^{\star},  \cG }_{i, h}(c_h)-{V}_{i, h}^{\pi_i^\star\times\hat{\pi}_{-i}^\star, \cM   }({c}_h)\right)
    +\left({V}_{i, h}^{\hat{\pi}^\star, \cM   }({c}_h) -V^{\hat{\pi}^\star,  \cG }_{i, h}(c_h)\right)\right]  + (H+1-h)\epsilon_e\\
    &\le 2(H-h+1)\epsilon_r + (H-h)(H-h+1)\epsilon_z + (H-h+1)\epsilon_e.
\end{align*}
Let $h=1$, and note that $c_1=\emptyset$, we get
\[
    V^{\pi_i^{\star}\times \hat{\pi}_{-i}^{\star},  \cG }_{i, 1}(\emptyset)-V^{\hat{\pi}^\star,  \cG }_{i, 1}(\emptyset)\le 2H\epsilon_r + H^2\epsilon_z + H\epsilon_e.
\]
By the definition of $\pi_{i}^{\star}$, we conclude
\begin{equation*}
\operatorname{NE/CCE-gap}(\hat{\pi}^\star)\le 2H\epsilon_r + {H^2}\epsilon_z + H\epsilon_e.
\end{equation*}
For computing CE, define
\[
    \phi_i^\star \in \arg\max_{\phi_{i}}
    V^{(\phi_i\diamond\hat{\pi}_i^\star)\odot{\hat{\pi}^{\star}_{-i}}, \cG}_{i, 1}(\emptyset).
\]
Now note that
\$
    &\EE_{a_{1:h-1}, o_{1:h}\sim \pi^{\prime}}[V^{(\phi_i^\star\diamond \hat{\pi}_i^\star)\odot\hat{\pi}_{-i}^{\star},  \cG }_{i, h}(c_h)-V^{\hat{\pi}^\star,  \cG }_{i, h}(c_h)]\\
    &=\EE_{a_{1:h-1}, o_{1:h}\sim \pi^{\prime}}\left[\left(V^{(\phi_i^\star\diamond \hat{\pi}_i^\star)\odot \hat{\pi}_{-i}^{\star},  \cG }_{i, h}(c_h)-{V}_{i, h}^{\hat{\pi}^\star, \cM   }({c}_h)\right)
    +\left({V}_{i, h}^{\hat{\pi}^\star, \cM   }({c}_h) -V^{\hat{\pi}^\star,  \cG }_{i, h}(c_h)\right)\right]\\
    &\le\EE_{a_{1:h-1}, o_{1:h}\sim \pi^{\prime}}\Big[\left(V^{(\phi_i^\star\diamond \hat{\pi}_i^\star)\odot\hat{\pi}_{-i}^{\star},  \cG }_{i, h}(c_h)-{V}_{i, h}^{(\phi_i^\star\diamond \hat{\pi}_i^\star)\odot\hat{\pi}_{-i}^\star, {\cM}   }({c}_h)\right)\Big]\\
    &\qquad+\EE_{a_{1:h-1}, o_{1:h}\sim \pi^{\prime}}\Big[\left({V}_{i, h}^{\hat{\pi}^\star, {\cM}   }({c}_h) -V^{\hat{\pi}^\star,  \cG }_{i, h}(c_h)\right)\Big]+ (H+1-h)\epsilon_e\\
    &\le 2(H-h+1)\epsilon_r + (H-h)(H-h+1)\epsilon_z + (H-h+1)\epsilon_e.
\$
Let $h=1$, and note that $c_1=\emptyset$, we get
\[
    V^{(\phi_i^\star\diamond \hat{\pi}_i^\star)\odot \hat{\pi}_{-i}^{\star},  \cG }_{i, 1}(\emptyset)-V^{\hat{\pi}^\star,  \cG }_{i, 1}(\emptyset)\le 2H\epsilon_r + H^2\epsilon_z + H\epsilon_e.
\]
By the definition of $\phi_i^\star$, we conclude
\[
\operatorname{CE-gap}(\hat{\pi}^\star)\le 2H\epsilon_r + {H^2}\epsilon_z + H\epsilon_e.
\]
The last step is the analysis of the computational  complexity. A major difference from the exact common-information setting is that it is unclear whether there exist efficient NE/CE/CCE subroutines at each step $h$. However, if $\cM$ is consistent with some approximate belief $\{\PP_{h}^{\cM, c}(s_h, p_h\given \hat{c}_h)\}_{h\in [H]}$, by Proposition \ref{prop:linear-M}, we conclude the NE subroutine for zero-sum or cooperative games and CE/CCE subroutine for general-sum games can be also implemented efficiently with the computational complexity of $\texttt{poly}(S, A, P_h, \frac{1}{\epsilon_e})$. Hence the overall computational  complexity of the Algorithm \ref{alg:avi} is \\$H\max_h\hat{C}_h\texttt{poly}(S, A, P_h, \frac{1}{\epsilon_e})$, where $\hat{C}_h$ comes from the loop at each step $h$.
\end{prevproof}

Finally, we are ready to prove Theorem \ref{thm:planning_ind_common_belief} as a special case.
\begin{prevproof}{thm:planning_ind_common_belief}
 we can leverage the reduction in \cite{nayyar2013common}   that reduces $\cG$ to an \emph{exact} common information model $\cM(\cG)$ such that $\epsilon_{z}(\cM(\cG))=\epsilon_r(\cM(\cG)) = 0$, where in this $\cM(\cG)$, we have $\hat{c}_h = c_h$ for any $h\in [H+1], c_{h}\in \cC_{h}$, and $\cM(\cG)$ is consistent with $\{\PP_h^{\cG}(s_h, p_h\given c_h)\}_{h\in [H]}$. Therefore, by applying \Cref{thm:struc}, we conclude the proof.
 \end{prevproof}
\subsection{Proof of \Cref{thm:plan_cases}}\label{subsec:finite_memory}

Theorem \ref{thm:struc} provides a structural result for the optimality of NE/CE/CCE policy computed with approximate common information in the underlying POSG, when the approximate common information satisfies the condition in Definition  \ref{def:ais}. However, it is not clear how to construct such approximate common information and how high the induced computational complexity is. Here we will show when the joint observation is informative enough, specifically satisfying Assumption \ref{observa}, we could simply use \emph{finite-memory truncation} to compress the common information, and indeed, the corresponding most recent $L$ steps of history is a kind of approximate common information. Importantly, we need the a series of following result showing that the most recent history is enough to predict the latent state of the POSG (with information sharing).

\begin{lemma}[Lemma 4.9 in \cite{golowich2022planning}]\label{lemma:contract}
Suppose the POSG satisfies Assumption \ref{observa}, $b, b^\prime\in \Delta(\cS)$ with $b \ll b^{\prime}$, and fix any $h\in [H]$. Then
\[
    \mathbb{E}_{y \sim \mathbb{O}_h^{\top} b}\left[\sqrt{\exp \left(\frac{D_2\left(B_h(b ; y) \| B_h\left(b^{\prime} ; y\right)\right)}{4}\right)-1}\right] \leq\left(1-\gamma^4 / 2^{40}\right) \cdot \sqrt{\exp \left(\frac{D_2\left(b \| b^{\prime}\right)}{4}\right)-1},
\]
where we recall the definition of $B_h$ in \Cref{subsec:belief}.
\end{lemma}
This lemma states that once the emission $\OO_h$ satisfies the condition in Assumption \ref{observa}, the Bayes operator $B_{h}$ is a contraction in expectation. Since the individual emission $\OO_{i, h}$ does not necessarily satisfy Assumption \ref{observa}, the individual Bayes operator $B_{i, h}$ satisfies a weaker result. We first state a more  general  lemma as follows. 

\begin{lemma}\label{lemma:poster}
    Given two finite domains $X, Y$, and the conditional probability $q(y\given x)$ for $x\in X, y\in Y$. Define the posterior update $F^q(P;y):\Delta(X)\rightarrow \Delta(X)$ for $P\in \Delta(X), y\in Y$ as
    \begin{equation}\label{eq:poster}
    F^q(P;y)(x) = \frac{P(x)q(y\given x)}{\sum_{x^\prime\in X} P(x^\prime)q(y\given x^\prime)}.
    \end{equation}
    Then for any $\delta_1, \delta_2\in \Delta(X)$ such that $\delta_1\ll \delta_2$, it holds that
    \[
    \EE_{x\sim \delta_1, y\sim q(\cdot\given x)}\sqrt{\exp\left(\frac{D_2(F^q(\delta_1;y)||F^q(\delta_2; y))}{4}\right) - 1}\le \sqrt{\exp\left(\frac{D_2(\delta_1||\delta_2)}{4}\right) - 1}.
    \]
\end{lemma}
\begin{proof}

    This is a direct consequence of the proof of Lemma 4.9 in \cite{golowich2022planning} by allowing $\gamma=0$ since here we do not assume any observability on $q$.
\end{proof}
\begin{corollary}\label{corr:non_expansive}
Suppose $b, b^\prime\in \Delta(\cS)$ with $b\ll b^\prime$, and fix any $h\in [H], i\in [n]$. Then 
\[
    \mathbb{E}_{y \sim \mathbb{O}_{i, h}^{\top} b}\left[\sqrt{\exp \left(\frac{D_2\left(B_{i, h}(b ; y) \| B_{i, h}\left(b^{\prime} ; y\right)\right)}{4}\right)-1}\right] \leq \sqrt{\exp \left(\frac{D_2\left(b \| b^{\prime}\right)}{4}\right)-1}.
\]
\end{corollary}
\begin{lemma}[Lemma 4.8 in \cite{golowich2022planning}]\label{2renyi}
Consider probability distributions $P$, $Q$. Then 
\[
    \|P-Q\|_1 \leq 4 \cdot \sqrt{\exp \left(D_2(P \| Q) / 4\right)-1}.
\]
\end{lemma}

\begin{theorem}[Adapted from Theorem 4.7 in \cite{golowich2022planning}]\label{noah_theo}
There is a constant $C\ge 1$ so that the following holds. Suppose that the POSG  satisfies Assumption \ref{observa} with parameter $\gamma$. Let $\epsilon\ge 0$. Fix a policy $\pi^\prime\in\Delta(\Pi^{\mathrm{det}})$ and indices $1\le h-L<h-1\le H$. If $L\ge C\gamma^{-4}\log(\frac{S}{\epsilon})$, then the following set of propositions hold
\begin{align}
    \label{eq:contract_1}\EE^\cG_{a_{1:h-1}, o_{1:h}\sim \pi^\prime}\| \bb_{h}(a_{1:h-1}, o_{1:h}) - \bb_{h}^{\prime}(a_{h-L:h-1}, o_{h-L+1:h})\|_1&\le \epsilon,\\
    \label{eq:contract_2}\EE^\cG_{a_{1:h-1}, o_{1:h}\sim \pi^\prime}\| \bb_{h}(a_{1:h-1}, o_{1:h-1}) - \bb_{h}^{\prime}(a_{h-L:h-1}, o_{h-L+1:h-1})\|_1&\le \epsilon,\\
\label{eq:contract_3}\EE^\cG_{a_{1:h-1}, o_{1:h}\sim \pi^\prime}\| \bb_{h}( a_{1:h-1}, o_{1:h-1}, o_{1, h}) - \bb_{h}^{\prime}(a_{h-L:h-1}, o_{h-L+1:h-1}, o_{1, h})\|_1&\le \epsilon.
\end{align}
Furthermore, for any finite domain $Y$, conditional probability $q(y\given s)$ and the posterior update operator $F^q:\Delta(\cS)\rightarrow\Delta(\cS)$ as defined in Lemma \ref{lemma:poster}, it holds that
\#
\EE^\cG_{\pi^\prime}\EE_{y\sim q\cdot \bb_{h}(a_{1:h-1}, o_{1:h})}\| F^q(\bb_{h}(a_{1:h-1}, o_{1:h}); y) - F^q(\bb_{h}^{\prime}(a_{h-L:h-1}, o_{h-L+1:h}); y)\|_1\le \epsilon.\label{eq:contract_5}
\#
\end{theorem}
\begin{proof}

 \Cref{eq:contract_1} is from Theorem 4.7 in \cite{golowich2022planning}. For the remaining, it suffices to only consider $\pi^\prime\in\Pi^{\text{det}}$. We prove \Cref{eq:contract_2} first.
Note that if $h-L\le 1$, then we have $\bb_h(a_{1:h-1}, o_{1:h-1}) = \bb_{h}^\prime(a_{h-L:h-1}, o_{h-L+1:h-1})$. The proposition holds trivially. Now let us consider $h>L+1$. Fix some history $(a_{1:h-L-1}, o_{1:h-L})$. We condition on this history throughout the proof. For $0\le t\le L$, define the random variables
\begin{align*}
b_{h-L+t}&=\bb_{h-L+t}\left(a_{1: h-L+t-1}, o_{1: h-L+t-1}\right), \\
b_{h-L+t}^{\prime}&=\bb_{h-L+t}^{\prime}\left(a_{h-L: h-L+t-1}, o_{h-L+1: h-L+t-1}\right), \\
Y_t&=\sqrt{\exp \left(\frac{D_2\left(b_{h-L+t} \| b_{h-L+t}^{\prime}\right)}{4}\right)-1}.
\end{align*}
Then $D_2(b_{h-L}||b_{h-L}^\prime)=\log\EE_{x\sim b_h}\frac{b_{h}(x)}{b_{h}^\prime(x)}\le \log(S)$ since $b_{h-L}^{\prime} = \bb_{h-L}^\prime(\emptyset) = \operatorname{Unif}(\cS)$, so we have  
\[
Y_0\le \sqrt{\exp(D_2(b_{h-L}||b_{h-L}^\prime))}\le S.
\]
Moreover, for any $0\le t\le L-1$, by denoting the shorthand notation of the matrix $A:=\TT_{h-L+t}(a_{h-L+t})$, we have :
\$
    &\EE_{a_{h-L: h-L+t}, o_{h-L+1: h-L+t}\sim \pi^\prime}Y_{t+1} \\
    &=\EE_{ a_{h-L: h-L+t-1}, o_{h-L+1: h-L+t}\sim \pi^\prime}\EE_{a_{h-L+t}\sim \pi^\prime(\cdot\given a_{1: h-L+t-1}, o_{1: h-L+t})}\\
    &\left[\sqrt{\exp\left(\frac{D_2(A\cdot B_{h-L+t}(b_{h-L+t}; o_{h-L+t})||A\cdot B_{h-L+t}(b_{h-L+t}^\prime; o_{h-L+t}))}{4}\right)-1}\right] \\
    & \le\EE_{\substack{(a_{h-L: h-L+t-1},\\ o_{h-L+1: h-L+t-1})\sim \pi^\prime}
    }\EE_{o_{h-L+t} \sim \OO_{h-L+t}^\top b_{h-L+t}} \left[\sqrt{\exp\left(\frac{D_2( B_{h-L+t}(b_{h-L+t}; o_{h-L+t})|| B_{h-L+t}(b_{h-L+t}^\prime; o_{h-L+t}))}{4}\right)-1}\right]\\
    &\le \left(1-\frac{\gamma^4}{2^{40}}\right)\EE_{a_{h-L: h-L+t-1}, o_{h-L+1: h-L+t-1}\sim \pi^\prime} Y_t,
\$
where the second last step comes from the data processing inequality and the last step comes from Lemma \ref{lemma:contract}. By induction and the choice of $L$, we have that
\begin{equation}\label{eq:contract_4}
    \EE_{o_{h-L: h-1}, a_{h-L: h-1}\sim \pi^\prime}\sqrt{\exp \left(\frac{D_2\left(b_{h} \| b_{h}^{\prime}\right)}{4}\right)-1}\le \left(1-\frac{\gamma^4}{2^{40}}\right)^L S\le \frac{\epsilon}{4}.
\end{equation}
It follows from Lemma \ref{2renyi} that
\[
    \EE_{a_{h-L: h-1}, o_{h-L+1: h-1}\sim \pi^\prime}||b_{h}-b_{h}^\prime||_1\le \epsilon.
\]

 Equation \eqref{eq:contract_2} follows from Equation \eqref{eq:contract_4} and Lemma \ref{lemma:contract}. Equation \eqref{eq:contract_3} follows from Equation \eqref{eq:contract_4} and Corollary \ref{corr:non_expansive}. Equation \eqref{eq:contract_5} follows from Equation \eqref{eq:contract_4} and Lemma \ref{lemma:poster}.
\end{proof}

Before instantiating the  information structure in particular cases, we prove\\ \Cref{lemma:bounding_z_r_diff_using_c} first, which is a more sufficient condition for our Definition \ref{def:ais}. 

\begin{prevproof}
{lemma:bounding_z_r_diff_using_c}
By \Cref{def:consistency}, it holds that
\[
\left|\EE^\cG[r_{i, h}(s_h, a_h)\given c_h, {\gamma}_h]-\hat{r}_{i, h}^\cM(\hat{c}_h, {\gamma}_h)\right|\le \sum_{s_h, a_h}\left|\PP_{h}^{{ \cG}}(s_h, a_h\given {c}_h, {\gamma}_h) - \PP_{h}^{\cM, o}(s_h, a_h\given \hat{c}_h, {\gamma}_h)\right|.
\] 
Therefore, it suffices to bound the right-hand side to order to prove Equation \eqref{eq:any_r}. Now, note that for any $c_h\in\cC_h, {\gamma}_h\in\Gamma_h$:
\$
   &\sum_{s_h, p_h, a_h, s_{h+1}, o_{h+1}}\Big|\PP_{h}^{ \cG}(s_h, s_{h+1}, p_h, a_h, o_{h+1}\given c_h, {\gamma}_h)-\PP_{h}^{\cM}(s_h, s_{h+1}, p_h, a_h, o_{h+1}\given\hat{c}_h, {\gamma}_h)\Big|\\
   &=\sum_{s_h, p_h, a_h, s_{h+1}, o_{h+1}}\bigg|\PP_{h}^ \cG(s_h, p_h\given c_h)\prod_{j=1}^n\gamma_{j, h}(a_{j, h}\given p_{j, h})\TT_{h}(s_{h+1}\given s_h, a_h)\OO_{h+1}(o_{h+1}\given s_{h+1})\\
   &\qquad\qquad-\PP_{h}^{\cM, c}(s_h, p_h\given\hat{c}_h)\prod_{j=1}^n\gamma_{j, h}(a_{j, h}\given p_{j, h})\TT_{h}(s_{h+1}\given s_h, a_h)\OO_{h+1}(o_{h+1}\given s_{h+1})\bigg|\\
   &=\sum_{s_h, p_h}\Big|\PP_{h}^ \cG(s_h,p_h \given c_{h}) - \PP_{h}^{\cM, c}(s_h,p_h\given \hat{c}_h)\Big|.
\$
Finally, since after marginalization, the total variation will not increase, we conclude that
\$
&\sum_{z_{h+1}}\Big|\PP_{h}^ \cG(z_{h+1}\given c_h, {\gamma}_h) - \PP_{h}^{\cM, z}(z_{h+1}\given\hat{c}_h, {\gamma}_h)\Big|\\  
&\quad\le\sum_{s_h, p_h, a_h, s_{h+1}, o_{h+1}}\Big| \PP_{h}^{ \cG}(s_h, s_{h+1}, p_h, a_h, o_{h+1}\given c_h, {\gamma}_h)-\PP_{h}^{\cM}(s_h, s_{h+1}, p_h, a_h, o_{h+1}\given \hat{c}_h, {\gamma}_h)\Big|,\\
&\sum_{s_h, a_h}\Big|\PP_{h}^\cG(s_h, a_h\given c_h, {\gamma}_h) - \PP_{h}^{\cM}(s_h, a_h\given\hat{c}_h, {\gamma}_h)\Big|\\
&\quad\le\sum_{s_h, p_h, a_h, s_{h+1}, o_{h+1}}\Big|\PP_{h}^{\cG}(s_h, s_{h+1}, p_h, a_h, o_{h+1}\given c_h, {\gamma}_h)-\PP_{h}^{\cM}(s_h, s_{h+1}, p_h, a_h, o_{h+1}\given \hat{c}_h, {\gamma}_h)\Big|, 
\$
which proved the lemma.
\end{prevproof}

Therefore, in the following discussion, we only need to define $\hat{c}_h$ and the corresponding belief \\$\{\PP_h^{\cM, c}(s_h, p_h\given \hat{c}_h)\}_{h\in [H]}$. The definition of $\PP_{h}^{\cM, z}(z_{h+1}\given\hat{c}_h, {\gamma}_h)$ and $\hat{r}_{i, h}^\cM(\hat{c}_h, {\gamma}_h)$ will follow from the consistency condition \eqref{consis:t} and \eqref{consis:r}. Now we will show when $\cG$ satisfies our Assumptions \ref{evo}, \ref{observa}, \ref{str_indi}, how we can construct approximate common information with history truncation that satisfies Definition \ref{def:ais}.

\paragraph{One-step delayed information-sharing.} In this case, the information structure has $c_h=\{a_{1:h-1}, o_{1:h-1}\}$, $p_{i, h} = \{o_{i, h}\}$, $z_{h+1} = \{o_{h}, a_{h}\}$, {and $\PP_h^{\cG}(s_h, p_h\given {c}_h) = {\bb}_h(a_{1:h-1}, o_{1:h-1})(s_h)\OO_{h}(o_h\given s_h)$, which verifies Assumption \ref{str_indi}}. Fix $L>0$, we define the approximate common information as $\hat{c}_h = \{a_{h-L:h-1}, o_{h-L+1:h-1}\}$. Furthermore, define the common information conditioned belief as $\PP_h^{\cM, c}(s_h, p_h\given \hat{c}_h) = {\bb}^\prime_h(a_{h-L:h-1}, o_{h-L+1:h-1})(s_h)\OO_{h}(o_h\given s_h)$. Now we are ready to verify that it satisfies Definition \ref{def:ais}.
\begin{itemize}
    \item Obviously, it satisfies condition \eqref{def:ais_evo}. 
    \item Note that for any $ c_h\in \cC_h$ {and the corresponding $\hat{c}_h$ constructed above}:
    \$
        &\|\PP_{h}^ \cG(\cdot, \cdot\given c_h) - \PP_{h}^{\cM, c}(\cdot, \cdot\given\hat{c}_h)\|_1 \\
        &\quad=\sum_{s_h, o_{h}}\Big|\bb_h(a_{1:h-1}, o_{1:h-1})(s_h)\OO_{h}(o_h\given s_h)-\bb_h^{\prime}(a_{h-L:h-1}, o_{h-L+1:h-1})(s_h)\OO_{h}(o_h\given s_h)\Big|\\
        &\quad=\|\bb_h(a_{1:h-1}, o_{1:h-1})-\bb_h^{\prime}(a_{h-L:h-1}, o_{h-L+1:h-1})\|_1.
    \$
Therefore, by setting $L\ge C\gamma^{-4}\log(\frac{S}{\epsilon})$, according to Equation \eqref{eq:contract_1} in Theorem \ref{noah_theo}, we conclude that for any $ \pi^\prime\in\Pi^{\mathrm{det}}, h\in [H]$: 
\begin{align*}
    &\EE_{a_{1:h-1}, o_{1:h}\sim \pi^\prime}^\cG\|\PP_{h}^ \cG(\cdot, \cdot\given c_h) - \PP_{h}^{\cM, c}(\cdot, \cdot\given\hat{c}_h)\|_1 \\
    &\quad\le \EE_{a_{1:h-1}, o_{1:h}\sim \pi^\prime}\|\bb_h(a_{1:h-1}, o_{1:h-1})-\bb_h^{\prime}(a_{h-L:h-1}, o_{h-L+1:h-1})\|_1\le \epsilon.
\end{align*}
Therefore, conditions  \eqref{def:ais_2}, \eqref{def:ais_3} in Definition \ref{def:ais} are satisfied {using Lemma \ref{lemma:bounding_z_r_diff_using_c}} with $\epsilon_r=\epsilon_z=\epsilon$. 
\end{itemize}

Formally, we have the following theorem:
\begin{theorem}
Let $\epsilon, \gamma>0$. Algorithm \ref{alg:vi} given a $\gamma$-observable POSG of one-step delayed information sharing \xynew{computes an $\epsilon$-NE if the POSG is zero-sum or cooperative, and an $\epsilon$-CE/CCE if the POSG is general-sum with} time complexity $H(AO)^{C\gamma^{-4}\log\frac{SH}{\epsilon}}\texttt{poly}(S, A, O, H, \frac{1}{\epsilon})$ for some universal constant $C>0$.
\end{theorem}
\begin{proof}
It is direct to see that $\hat{C}_h \le (AO)^L$ and $P_h\le O$, the polynomial dependence on $S$, $H$, $A$, and $O$ comes from computing $\PP_{h}^{\cM, c}(s_h, p_h\given \hat{c}_h)$ {and the equilibrium computation subroutines}.
\end{proof}

\paragraph{State controlled by one controller with asymmetric delay sharing.} The information structure is given as $c_{h} = \{o_{1, 1:h}, o_{2, 1:h-d}, a_{1, 1:h-1}\}$, $p_{1, h} = \emptyset$, $p_{2, h} = \{o_{2, h-d+1:h}\}$, $z_{h+1} = \{o_{1, h+1}, o_{2, h-d+1}, a_{1, h}\}$. It is a bit less straightforward to verify Assumption \ref{str_indi}. We do so by explicitly computing $\PP_{h}^ \cG(s_h, p_h\given c_h)$  as follows. Denote $\tau_{h-d} = \{a_{1, 1:h-d-1}, o_{1:h-d}\}$, $f_{a} = \{a_{1, h-d:h-1}\}$, $f_{o} = \{o_{1, h-d+1:h}\}$. Now $\PP_h^{ \cG}(s_h, p_h\given c_h) = \sum_{s_{h-d}}\PP^ \cG_h(s_h, p_h\given s_{h-d}, f_a, f_o) \PP^\cG_h(s_{h-d}\given \tau_{h-d}, f_a, f_o)$. It is direct to see that $\PP_h^\cG(s_h, p_h\given s_{h-d}, f_a, f_o)$ does not depend on the policy. 
For $\PP^ \cG_h(s_{h-d}\given \tau_{h-d}, f_a, f_o)$, the following holds
\[
\PP^\cG_h(s_{h-d}\given \tau_{h-d}, f_a, f_o) = \frac{\PP^\cG_h(s_{h-d}, f_a, f_o\given \tau_{h-d})}{\sum_{s_{h-d}^\prime}\PP^\cG_h(s_{h-d}^\prime, f_a, f_o\given \tau_{h-d})}.
\]
Now note that
\$
&\PP^\cG_h(s_{h-d}, f_a, f_o\given \tau_{h-d}) \\
&\quad= \bb_{h-d}(a_{1, 1:h-d-1}, o_{1:h-d})(s_{h-d})\PP^\cG_h(a_{1, h-d}\given \tau_{h-d}) \PP^\cG_h(o_{1, h-d+1}\given s_{h-d}, a_{1, h-d})\cdots\PP^{\cG}_h(o_{1, h}\given s_{h-d}, a_{1, h-d:h-1}).
\$
Now let us use the notation $P_h(f_{o}\given s_{h-d}, f_a):= \prod_{t=1}^{d}\PP^\cG_h(o_{1, h-d+t}\given s_{h-d}, a_{1, h-d:h-d+t-1})$. Then it holds that $\sum_{f_o}P_h(f_o\given s_{h-d}, f_a) = 1$, which suggests that the notation $P_h(f_{o}\given s_{s-d}, f_a)$ can be understood as a conditional probability. With such notation, we have
\$
\PP^\cG_h(s_{h-d}\given \tau_{h-d}, f_a, f_o) &= \frac{\bb_{h-d}(a_{1, 1:h-d-1}, o_{1:h-d})(s_{h-d})P_h(f_o\given s_{h-d}, f_a)}{\sum_{s_{h-d}^\prime}\bb_{h-d}(a_{1, 1:h-d-1}, o_{1:h-d})(s_{h-d}^\prime)P_h(f_o\given s_{h-d}^\prime, f_a)} \\
&= F^{P_h(\cdot\given \cdot, f_a)}(\bb_{h-d}(a_{1, 1:h-d-1}, o_{1:h-d}); f_o)(s_{h-d}),
\$ 
where we recall the definition of $F$ in  Lemma \ref{lemma:poster}. Finally, we compute:
\$
\PP_h^{ \cG}(s_h, p_h\given c_h) = \sum_{s_{h-d}}\PP^ \cG_h(s_h, p_h\given s_{h-d}, f_a, f_o)F^{P_h(\cdot\given \cdot, f_a)}(\bb_{h-d}(a_{1, 1:h-d-1}, o_{1:h-d}); f_o)(s_{h-d}).
\$
It is easy to see that this expression does not depend on the policy executed, thus verifying Assumption \ref{str_indi}. Now for some fixed $L> 0$, we construct the approximate common information \\$\hat{c}_h :=\{o_{1, h-d-L+1:h}, o_{2, h-d-L+1:h-d}, a_{1, h-d-L:h-1}\}$ and correspondingly:
\#\label{eq:complex_belief}
	\PP_{h}^{\cM, c}(s_h, p_{h}\given\hat{c}_h) \nonumber=\sum_{s_{h-d}}\PP^\cG_h(s_h, p_h\given s_{h-d}, f_a, f_o)F^{P_h(\cdot\given \cdot, f_a)}(\bb_{h-d}^\prime(a_{1, h-d-L:h-d-1}, o_{h-d-L+1:h-d}); f_o)(s_{h-d}).
\#

To verify Definition \ref{def:ais}:\begin{itemize}
    \item Obviously, it satisfies the condition \eqref{def:ais_evo}.
    \item For any $c_h\in \cC_h$  {and the corresponding $\hat{c}_h$ constructed above}:
    \$
    &\|\PP_{h}^{ \cG}(\cdot, \cdot\given c_h) -\PP_h^{\cM, c}(\cdot, \cdot\given \hat{c}_h) \|_1\\
    &\quad\le \Big\|F^{P(\cdot\given \cdot, f_a)}(\bb_{h-d}(a_{1, 1:h-d-1}, o_{1:h-d}); f_o) - F^{P(\cdot\given \cdot, f_a)}(\bb_{h-d}^\prime(a_{1, h-d-L:h-d-1}, o_{h-d-L+1:h-d}); f_o)\Big\|_1.
    \$
    Finally, for any policy $\pi^\prime\in \Pi^{\mathrm{det}}$ taking expectations over $\tau_{h-d}, f_a, f_o$, we conclude that as long as $L\ge C\gamma^{-4}\log\frac{S}{\epsilon}$ using Equation \ref{eq:contract_5} of Theorem \ref{noah_theo}, we have 
    \[
    \EE_{a_{1:h-1}, o_{1:h}\sim \pi^\prime}^ \cG\|\PP_{h}^{ \cG}(\cdot, \cdot\given c_h) -\PP_h^{\cM, c}(\cdot, \cdot\given \hat{c}_h) \|_1\le\epsilon.
    \]
    Therefore, conditions  \eqref{def:ais_2},  \eqref{def:ais_3} in  \Cref{def:ais}  are satisfied {using Lemma \ref{lemma:bounding_z_r_diff_using_c}} with $\epsilon_r=\epsilon_z=\epsilon$. 
\end{itemize}

Formally, we have the following theorem:
\begin{theorem}
Let $\epsilon, \gamma>0$. Algorithm \ref{alg:vi} given a $\gamma$-observable POSG of state controlled by one controller with asymmetric delay sharing \xynew{computes an $\epsilon$-NE if the POSG is zero-sum or cooperative, and an $\epsilon$-CE/CCE if the POSG is general-sum with} time complexity $H(AO)^{C(\gamma^{-4}\log\frac{SH}{\epsilon} + d)}\texttt{poly}(S, A, O, H, \frac{1}{\epsilon})$ for some universal constant $C>0$.
\end{theorem}
\begin{proof}
It follows from the fact that $\hat{C}_h \le (AO)^{L+d}$ and $P_h\le O_2^d$. The polynomial dependence on $S$, $H$, $A$, and $O$ comes from computing $\PP_{h}^{\cM, c}(s_h, p_h\given \hat{c}_h)$ {and the equilibrium computation subroutines.} 
\end{proof}

\paragraph{Information sharing with one-directional-one-step delay.} For this case, we have 
$c_h = \{a_{1:h-1}, o_{1:h-1}, o_{1, h} \}$, $p_{1, h} = \emptyset$, $p_{2, h} = \{o_{2, h}\}$, $z_{h+1} = \{o_{1, h+1}, o_{2, h}, a_h\}$, {and $\PP_{h}^{\cG}(s_h, p_{h}\given {c}_h) = \bb_{h}(a_{1:h-1}, o_{1:h-1}, o_{1, h})(s_h)\PP_h(o_{2, h}\given s_h, o_{1, h})$}, where $\PP_h(o_{2, h}\given s_h, o_{1, h}) = \frac{\OO_{h}(o_{1, h}, o_{2, h}\given s_h)}{\sum_{o_{2, h}^\prime}\OO_h (o_{1, h}, o_{2, h}^\prime \given s_h)}$, thus verifying Assumption \ref{str_indi}. Fix $L>0$, we construct the approximate common information as $\hat{c}_h = \{a_{h-L:h-1}, o_{h-L+1:h-1}, o_{1, h}\}$. Furthermore, we define the belief as 
\[
\PP_{h}^{\cM, c}(s_h, p_{h}\given \hat{c}_h) = \bb_{h}^{\prime}(a_{h-L:h-1}, o_{h-L+1:h-1}, o_{1, h})(s_h)\PP_h(o_{2, h}\given s_h, o_{1, h}).
\]
Now we are ready to verify that Definition  $\ref{def:ais}$ is satisfied.
\begin{itemize}
    \item Obviously, the condition \eqref{def:ais_evo} is satisfied.
    \item Note that for any $ c_h\in\cC_h$ {and the corresponding $\hat{c}_h$ constructed above}:
    \begin{align*}
        &\Big\|\PP_{h}^\cG(\cdot, \cdot\given c_h) - \PP_{h}^{\cM, c}(\cdot, \cdot\given\hat{c}_h)\Big\|_1 \\   
        &\quad=\sum_{s_h, o_{2,h}}\Big|\bb_{h}(a_{1:h-1}, o_{1:h-1}, o_{1, h} )(s_h)\PP_h(o_{2, h}\given s_h, o_{1, h})-\bb_{h}^{\prime}(a_{h-L:h-1}, o_{h-L+1:h-1}, o_{1, h} )(s_h)\PP_h(o_{2, h}\given s_h, o_{1, h})\Big|\\
        &\quad=\|\bb_{h}(a_{1:h-1}, o_{1:h-1}, o_{1, h} )-\bb_{h}^{\prime}(a_{h-L:h-1}, o_{h-L+1:h-1}, o_{1, h} )\|_1.
    \end{align*}
        Therefore, by setting $L\ge C\gamma^{-4}\log(\frac{S}{\epsilon})$, according to \eqref{eq:contract_3} in Theorem \ref{noah_theo}, we conclude that for any $ \pi^\prime\in\Pi^{\mathrm{det}}$:
    \$
        &\EE_{a_{1:h-1}, o_{1:h}\sim \pi^\prime}^\cG\|\PP_{h}^ \cG(\cdot, \cdot\given c_h) - \PP_{h}^{\cM, c}(\cdot, \cdot\given\hat{c}_h)\|_1\\
        &\quad\le \EE_{a_{1:h-1}, o_{1:h}\sim \pi^\prime}^\cG\Big\|\bb_{h}(a_{1:h-1}, o_{1:h-1}, o_{1, h} )-\bb_{h}^{\prime}(a_{h-L:h-1}, o_{h-L+1:h-1}, o_{1, h} )\Big\|_1\le \epsilon.
    \$
    Therefore, conditions \eqref{def:ais_2}, \eqref{def:ais_3} in \Cref{def:ais} are satisfied using {Lemma \ref{lemma:bounding_z_r_diff_using_c}} with $\epsilon_r=\epsilon_z=\epsilon$.
\end{itemize}

Formally, we have the following theorem:
\begin{theorem}
Let $\epsilon, \gamma>0$. Algorithm \ref{alg:vi} given a $\gamma$-observable POSG of information sharing with one-directional-one-step delay \xynew{computes an $\epsilon$-NE if the POSG is zero-sum or cooperative, and an $\epsilon$-CE/CCE if the POSG is general-sum with} time complexity $H(AO)^{C\gamma^{-4}\log\frac{SH}{\epsilon}}\texttt{poly}(S, A, O,H, \frac{1}{\epsilon})$ for some universal constant $C>0$.
\end{theorem}
\begin{proof}
It is direct to see that $\hat{C}_h \le (AO)^L$ and $P_h\le O_2$. The polynomial dependence on $S$, $H$, $A$, and $O$ comes from computing $\PP_{h}^{\cM, c}(s_h, p_h\given \hat{c}_h)$ and the equilibrium computation subroutines.
\end{proof}
\paragraph{Uncontrolled state process with delayed sharing.}
As long as the state transition does not depend on the actions, Assumption \ref{str_indi} is satisfied.  {To be more concrete, we have 
\[
\PP_h^{\cG}(s_h, p_h\given {c}_h) = \sum_{s_{h-d}}\bb_{h-d}(o_{h-d-L+1:h-d})(s_{h-d})\PP_h^\cG(s_{h}, o_{h-d+1:h}\given s_{h-d}),
\]
which verifies Assumption \ref{str_indi}, where in the notation for $\bb_{h-d}$, we omit the actions since they do not affect transitions.} For generality, we consider the $d$-step delayed sharing information structure, where $d\ge 0$ and not necessarily $d=1$,  as in the one-step delayed information sharing structure. The information structure satisfies $c_{h} = \{o_{1:h-d}\}$, $p_{i, h} = \{o_{i, h-d+1:h}\}$, and $z_{h+1} = \{o_{h-d+1}\}$. Fix a $L>0$, the approximate common information is $\hat{c}_h = \{o_{h-d-L+1:h-d}\}$, the corresponding belief is $\PP_h^{\cM, c}(s_h, p_h\given \hat{c}_h) = \sum_{s_{h-d}}\bb_{h-d}^\prime(o_{h-d-L+1:h-d})(s_{h-d})\PP_h^\cG(s_{h}, o_{h-d+1:h}\given s_{h-d})$. Now we are ready to verify Definition \ref{def:ais}.
\begin{itemize}
    \item Obviously, the condition \eqref{def:ais_evo} is satisfied.
    \item Note that for any $c_h$ {and the corresponding $\hat{c}_h$ constructed above}:
    \$
        &\Big\|\PP_{h}^ \cG(\cdot, \cdot\given c_h) - \PP_{h}^{\cM, c}(\cdot, \cdot\given\hat{c}_h)\Big\|_1 \\
        &=\sum_{s_h, o_{h-d+1:h}}\bigg|\sum_{s_{h-d}}\bb_{h-d}(o_{1:h-d})(s_{h-d})\PP^\cG_h(s_{h}, o_{h-d+1:h}\given s_{h-d})- \sum_{s_{h-d}}\bb_{h-d}^\prime(o_{h-d-L+1:h-d})(s_{h-d})\PP^\cG_h(s_{h}, o_{h-d+1:h}\given  s_{h-d})\bigg|\\
        &=\sum_{s_h, o_{h-d+1:h}}\bigg|\sum_{s_{h-d}}(\bb_{h-d}(o_{1:h-d})(s_{h-d})-\bb_{h-d}^\prime(o_{h-d-L+1:h-d})(s_{h-d}))\PP_h^\cG(s_{h}, o_{h-d+1:h}\given s_{h-d})\bigg|\\
        &\le \|\bb_{h-d}(o_{1:h-d})-\bb_{h-d}^\prime(o_{h-d-L+1:h-d})\|_1,
    \$
    where for the last step, we use Lemma \ref{trick} {(proved later)}. Therefore, by setting $L\ge C\gamma^{-4}\log(\frac{S}{\epsilon})$, according to Equation \eqref{eq:contract_2}   in Theorem \ref{noah_theo}, we conclude that for any $\pi^\prime\in\Pi^{\mathrm{det}}$:
    \$
        \EE_{\pi^\prime}^\cG\big\|\PP_{h}^ \cG(\cdot, \cdot\given c_h) - \PP_{h}^{\cM, c}(\cdot, \cdot\given\hat{c}_h)\big\|_1  \le\EE_{\pi^\prime}^\cG\big\|\bb_{h-d}(o_{1:h-d})-\bb_{h-d}^\prime(o_{h-d-L+1:h-d})\big\|_1\le \epsilon.
    \$
    This verifies the conditions  \eqref{def:ais_2}, \eqref{def:ais_3} in \Cref{def:ais} using Lemma \ref{lemma:bounding_z_r_diff_using_c} with $\epsilon_r=\epsilon_z=\epsilon$.
\end{itemize}

Finally, to guarantee that  $\hat{\pi}^\star$ is  an $\epsilon$-NE{/CE/CCE}, according to our Theorem \ref{thm:struc}, one needs $L\ge C\gamma^{-4}\log (\frac{SH}{\epsilon})$. Formally, we have the following theorem:
\begin{theorem}
Let $\epsilon, \gamma>0$. Algorithm \ref{alg:vi} given a $\gamma$-observable POSG of uncontrolled state process \xynew{computes an $\epsilon$-NE if the POSG is zero-sum or cooperative, and an $\epsilon$-CE/CCE if the POSG is general-sum with} time complexity $H(O)^{C\gamma^{-4}\log\frac{SH}{\epsilon}}\texttt{poly}(S, A, O^d, H, \frac{1}{\epsilon})$ for some universal constant $C>0$.
\end{theorem}
\begin{proof}
It is direct to see that $\hat{C}_h \le O^L$ and $P_h=O^d$. The polynomial dependence on $S$, $A$, $H$, and $O^d$ comes from computing $\PP_{h}^{\cM, c}(s_h, p_h\given \hat{c}_h)$ and the equilibrium computation  subroutines.
\end{proof}

\paragraph{Symmetric information game.} For symmetric information games, it has the following information structure:  $c_{h} = \{a_{1:h-1}, o_{1:h}\}$, $p_{i, h} = \emptyset$, $z_{h+1} = \{a_{h}, o_{h+1}\}$, {and $\PP_{h}^{\cG}(s_h, p_{h}\given {c}_h) =\bb_{h}(a_{1:h-1}, o_{1:h})(s_h)$, verifying Assumption \ref{str_indi}}. Fix $L>0$, we construct the approximate common information as $\hat{c}_{h} = \{a_{h-L:h-1}, o_{h-L+1:h}\}$. Furthermore, we define the belief $\PP_{h}^{\cM, c}(s_h, p_{h}\given \hat{c}_h) =\bb_{h}^\prime(a_{h-L:h-1}, o_{h-L+1:h})(s_h)$. Now we are ready to verify Definition \ref{def:ais}.  
\begin{itemize}
    \item Obviously, it satisfies the condition \eqref{def:ais_evo}.
    \item Note that for any $ c_h\in\cC_h$ {and the corresponding $\hat{c}_h$ constructed above}:
    \[
        \big\|\PP_{h}^ \cG(\cdot, \cdot\given c_h)-\PP_{h}^{\cM, c}(\cdot, \cdot\given\hat{c}_h)\big\|_1=\|\bb_h(a_{1:h-1}, o_{1:h})- \bb_h^{\prime}(a_{h-L, h-1}, o_{h-L+1:h})\|_1.
    \]
    Therefore, by setting $L\ge C\gamma^{-4}\log(\frac{S}{\epsilon})$, according to \eqref{eq:contract_2} in Theorem \ref{noah_theo}, we conclude that for any $\pi^\prime\in\Pi^{\mathrm{det}}$:
    \$
            \EE_{a_{1:h-1}, o_{1:h}\sim \pi^\prime}^\cG\big\|\PP_{h}^ \cG(\cdot, \cdot\given c_h)-\PP_{h}^{\cM, c}(\cdot, \cdot\given\hat{c}_h)\big\|_1=\|\bb_h(a_{1:h-1}, o_{1:h})- \bb_h^{\prime}(a_{h-L, h-1}, o_{h-L+1:h})\|_1\le \epsilon.
    \$
    Therefore, the conditions  \eqref{def:ais_2} and \eqref{def:ais_3} in \Cref{def:ais} are satisfied with $\epsilon_r=\epsilon_z=\epsilon$ using Lemma \ref{lemma:bounding_z_r_diff_using_c}.
\end{itemize}

Finally, to guarantee $\hat{\pi}^\star$ is an $\epsilon$-NE/CE/CCE, according to Theorem \ref{thm:struc}, one needs $L\ge C\gamma^{-4}\log (\frac{SH}{\epsilon})$. Formally, we have the following theorem:
\begin{theorem}
Let $\epsilon, \gamma>0$. Algorithm \ref{alg:vi} given a $\gamma$-observable POSG of symmetric information \xynew{computes an $\epsilon$-NE if the POSG is zero-sum or cooperative, and an $\epsilon$-CE/CCE if the POSG is general-sum with} time complexity $H(AO)^{C\gamma^{-4}\log\frac{SH}{\epsilon}}\texttt{poly}(S, A, H, O, \frac{1}{\epsilon})$ for some universal constant $C>0$.
\end{theorem}
\begin{proof}
It is direct to see that $\hat{C}_h = (AO)^L$ and $P_h=1$, the polynomial dependence on $S$, $H$, $A$, and $O$ comes from computing $\PP_{h}^{\cM, c}(s_h, p_h\given \hat{c}_h)$ and {the equilibrium computation subroutines.}
\end{proof}

{We conclude the section by proving the following lemma.} 

\begin{lemma}\label{trick}
For any given sequence $\{x_i\}_{i=1}^{m}$ and $\{\{y_{i, j}\}_{i=1}^{m}\}_{j=1}^{n}$ such that $\sum_{j=1}^{n}|y_{i, j}| = 1$ , $\forall i\in [m]$. The following holds
\[
    \sum_{j=1}^{n}\bigg|\sum_{i=1}^{m}x_i y_{i, j}\bigg|\le \sum_{i=1}^{m}|x_i|.
\]
\end{lemma}
\begin{proof}
Let $\bx = (x_1, \cdots, x_m)^{\top}$, $\by_j=(y_{1, j}, \cdots, y_{m, j})^\top$, and $\bY = (\by_1, \cdots, \by_n)$. Therefore, we have
\[
   \sum_{j=1}^{n}\bigg|\sum_{i=1}^{m}x_i y_{i, j}\bigg|=\sum_{j=1}^{n}|\bx^{\top}\by_j|=||\bY^{\top}\bx||_1\le ||\bY^\top||_1||\bx||_1. 
\]
Note that  $||\bY^\top||_1 = ||\bY||_{\infty} = \max_{i}\sum_{j=1}^{n}|y_{i, j}|=1$. Therefore, we have 
$\sum_{j=1}^{n}|\sum_{i=1}^{m}x_i y_{i, j}| \le \sum_{i=1}^{m}|x_i|,$ and conclude the proof. 
\end{proof}

\subsection{Proof of \Cref{thm:main_thm_learning}}\label{sec:proof_learning_ais}

Note that our previous planning algorithms require the knowledge of the true model (transition dynamics and rewards) of the POSG $ \cG$,  which avoids the issue of strategic explorations. For learning NE/CE/CCE in $ \cG$, one could {potentially} treat $ \cG$ as a (fully-observable) Markov game on the state space of $c_h$, {and use black-box algorithms for learning Markov games}. However, this formulation could be neither computationally nor sample efficient because of the typical large space of common information. Therefore, we have to learn NE/CE/CCE in {the approximate model} $\cM$ with the state space of $\hat{c}_h$ in Definition \ref{def:ais}. However, the key problem is that we can only sample according to the model of $ \cG$ instead of $\cM$. As we highlighted in \Cref{subsec:subsec-quasi-learn} of our main paper, to circumvent this issue, inspired by the idea of \cite{golowich2022learning}, one solution is to construct $\Tilde{\cM}(\pi^{1:H})$ using a sequence of $H$ policies $\pi^{1:H}$ according to \Cref{def:simulation_main}, where each $\pi^h\in\Delta(\Pi^{\mathrm{det}})$. 
Formally, Proposition \ref{prop:simulation} verifies that $\Tilde{\cM}(\pi^{1:H})$ constructed according to \Cref{def:simulation_main} can be simulated by executing policies $\pi^{h}$ at each step $h$ in the underlying true model $\cG$.
\begin{proposition}\label{prop:simulation}
    Given $\Tilde{\cM}(\pi^{1:H})$ as in Definition \ref{def:simulation_main}, it holds that for any $i\in[n]$, $h\in[H]$, $\hat{c}_h\in\hat{\cC}_h$, $\gamma_h\in\Gamma_h$, $o_{h+1}\in\cO$, $z_{h+1}\in\cZ_{h+1}$:
    \$
    \PP_h^{\Tilde{\cM}(\pi^{1:H}), z}(z_{h+1}\given\hat{c}_h, {\gamma}_h) &= \PP_h^{\pi^h_{1:h-1}, \cG}(z_{h+1}\given\hat{c}_h, {\gamma}_h),\\
    \hat{r}_{i, h}^{\Tilde{\cM}(\pi^{1:H})}(\hat{c}_h, \gamma_h) &= \EE^{\cG}_{\pi^h_{1:h-1}}[r_{i, h}(s_h, a_h)\given \hat{c}_h, \gamma_h]. 
    \$
\end{proposition}

\begin{proof}
	Note for $\PP_h^{\pi^h_{1:h-1}, \cG}(z_{h+1}\given\hat{c}_h, {\gamma}_h)$, it holds that
	\$
	&\PP_h^{\pi^h_{1:h-1}, \cG}(z_{h+1}\given\hat{c}_h, {\gamma}_h)\\
	&\quad=\sum_{\substack{p_h, a_h, o_{h+1}:\\\chi_{h+1}(p_h, a_h, o_{h+1})=z_{h+1}}}\PP_h^{\pi^h_{1:h-1}, \cG}(p_h, a_h, o_{h+1}\given\hat{c}_h, {\gamma}_h)\\
	&\quad=\sum_{\substack{s_h, p_h, a_h, o_{h+1}:\\\chi_{h+1}(p_h, a_h, o_{h+1})=z_{h+1}}}\Big(\PP_h^{\pi^h_{1:h-1}, \cG}(s_h, p_h, \given\hat{c}_h)\gamma_{h}(a_h\given p_h)\times\sum_{s_{h+1}}\TT_h(s_{h+1}\given s_h, a_h)\OO_{h+1}(o_{h+1}\given s_{h+1})\Big),
	\$
	where we recall the shorthand notation $\gamma_h(a_h\given p_h):=\prod_{j\in[n]}\gamma_{j, h}(a_{j, h}\given p_{j, h})$. Now by \Cref{def:simulation_main}, we have  $\PP_h^{\pi^h_{1:h-1}, \cG}(s_h, p_h, \given\hat{c}_h)=\PP_h^{\tilde{\cM}(\pi^{1:H}), c}(s_h, p_h, \given \hat{c}_h)$. Combined with Equation \eqref{consis:t} of \Cref{def:consistency}, we conclude $\PP_h^{\Tilde{\cM}(\pi^{1:H}), z}(z_{h+1}\given\hat{c}_h, {\gamma}_h) = \PP_h^{\pi^h_{1:h-1}, \cG}(z_{h+1}\given\hat{c}_h, {\gamma}_h)$. At the same time, we can prove $\hat{r}_{i, h}^{\Tilde{\cM}(\pi^{1:H})}(\hat{c}_h, \gamma_h) = \EE^{\cG}_{\pi^h_{1:h-1}}[r_{i, h}(s_h, a_h)\given \hat{c}_h, \gamma_h]$ holds by the same derivation.
	\end{proof}
Therefore, different from a generic $\cM$ in Definition \ref{def:ais}, to which we do not have algorithmic access, such a delicately designed transition dynamic and reward function allow us to actually simulate $\Tilde{\cM}(\pi^{1:H})$ by executing policies $\pi^{1:H}$ in $ \cG$.

The next question is how to explore the \emph{state space $\{\hat{\cC}_h\}_{h\in[H+1]}$}. It turns out that when such a state $\hat{c}_h$ {comes from} a sequence of observations and actions, a uniform policy can be used to explore the state space \citep{efroni2022provable, golowich2022learning}. Formally, define the under-explored set of $\hat{c}_h$ and $\hat{c}_h\cup p_h$ under some policy $\pi$ as follows.

\begin{definition}
{Fix $\hat{L}>0$ as given in Definition \ref{def:L_main}. }For each $h\in [H]$, $\zeta>0$, and a joint policy $\pi\in\Delta(\Pi^{\text{det}})$, define the set $\cC_{h,\zeta}^{\text{low}}(\pi)\subseteq\hat{\cC}_h$ as
\[
\cC_{h,\zeta}^{\text{low}}(\pi):=\left\{\hat{c}_h\in \hat{\cC}_h: d_{\cC, h}^{\pi, \cG}(\hat{c}_h)< \zeta\right\},
\]
and the set $\cV_{h, \zeta}^{\text{low}}(\pi)\subseteq \cV_h:=\hat{\cC}_h\times \cP_h$ as
\[
\cV_{h,\zeta}^{\text{low}}(\pi):=\left\{v_h\in {\cV}_h: d_{\cV, h}^{ \pi, \cG}(v_h)< \zeta\right\},
\]
and the set $\cX_{h,\zeta}^{\text{low}}(\pi)\subseteq \cX_h:=\cA^{\min\{h, \hat{L}\}} \times \cO^{\min\{h, \hat{L}\}}$ as
\[
\cX_{h,\zeta}^{\text{low}}(\pi):=\left\{x_h\in {\cX}_h: d_{\cX, h}^{\pi, \cG}(x_h)< \zeta\right\},
\]
where $d_{\cC, h}^{ \pi, \cG}(\hat{c}_h) := \PP_{h}^{\pi, \cG}(\hat{c}_h)$, $d_{\cV, h}^{\pi, \cG}(v_h) := \PP_{h}^{\pi, \cG}(v_h)$, and $d_{\cX, h}^{\pi, \cG}(x_h) := \PP_{h}^{\pi, \cG}(x_h)$. 
\end{definition}

Now we shall relate the under-explored set of $\hat{c}_h$   with the under-explored set of $s_{h^\prime}$ for some $h^\prime\in [H]$. Firstly, for any $\phi>0$, define the under-explored states under some policy $\pi\in\Delta(\Pi^{\text{det}})$ as
\[
    \cU_{\phi, h}^ \cG(\pi):=\{s\in \cS: d_{\cS,h}^{\pi, \cG}(s)< \phi\}.
\]
Then the following lemma holds.

\begin{lemma}\label{lemma:low_visitation}
    {Fix $\hat{L}>0$ as given in Definition \ref{def:L_main}.} Fix any $\zeta>0, \phi>0, h\in [H]$. Consider any policies $\pi$, $\pi^\prime\in\Delta(\Pi^{\text{det}})$, such that $\pi^\prime$ takes uniformly random actions at each step from $\max\{h-\hat{L}, 1\}$ to $h$, each chosen independently of all previous states, actions, and observations. Then, we have  
    \[
    \label{eq:low_c}d_{\cC, h}^{ \pi, \cG}(\cC_{h,\zeta}^{\text{low}}(\pi^\prime))\le \frac{A^{2\hat{L}}O^{\hat{L}}\zeta}{\phi} + \bm{1}[h>\hat{L}]\cdot d_{\cS, h-\hat{L}}^{ \pi, \cG}(\cU_{\phi, h-\hat{L}}^ \cG(\pi^\prime)).
    \]
\end{lemma}
\begin{proof}
   Note that we have for each $\hat{c}_h\in \hat{\cC}_h$
    \[
    d_{\cC, h}^{ \pi, \cG}(\hat{c}_h) = \sum_{x_h: \hat{f}_h(x_h) = \hat{c}_h}d_{\cX, h}^{\pi, \cG}(x_h)
    \]
   {where we recall the definition of $\hat{f}_h$ and $x_h$ from Definition \ref{def:L_main}.} Therefore, we have
    \$
    \sum_{\hat{c}_h\not\in \cC_{h,\zeta}^{\text{low}}(\pi^\prime)}d_{\cC, h}^{ \pi, \cG}(\hat{c}_h) = \sum_{\substack{\hat{c}_h\not\in \cC_{h,\zeta}^{\text{low}}(\pi^\prime)\\x_h: \hat{f}_h(x_h)= \hat{c}_h}}d_{\cX, h}^{\pi, \cG}(x_h)= \sum_{x_h:\hat{f}_h(x_h)\not\in \cC_{h,\zeta}^{\text{low}}(\pi^\prime)}d_{\cX, h}^{ \pi, \cG}(x_h)\ge \sum_{x_h\not\in \cX_{h,\zeta}^{\text{low}}(\pi^\prime)}d_{\cX, h}^{ \pi, \cG}(x_h),
    \$
    where the last step comes from the fact that $x_h\not\in \cX_{h,\zeta}^{\text{low}}(\pi^\prime)$ implies $\hat{f}_h(x_h)\not\in \cC_{h,\zeta}^{\text{low}}(\pi^\prime)$. This leads to that
    \[
    d_{\cC, h}^{ \pi, \cG}(\cC_{h,\zeta}^{\text{low}}(\pi^\prime))\le d_{\cX, h}^{ \pi, \cG}(\cX_{h,\zeta}^{\text{low}}(\pi^\prime))\le \frac{A^{2\hat{L}}O^{\hat{L}}\zeta}{\phi} + \bm{1}[h>\hat{L}]\cdot d_{\cS, h-\hat{L}}^{ \pi, \cG}(\cU_{\phi, h-\hat{L}}^ \cG(\pi^\prime)),
    \]
    where in the second inequality, we use Lemma 10.4 of \cite{golowich2022learning}.  
\end{proof}

The next step is to learn {$\PP_h^{\Tilde{\cM}(\pi^{1:H}), z}(z_{h+1}\given \hat{c}_h, {\gamma}_h)$, $\hat{r}_{i, h}^{\Tilde{\cM}(\pi^{1:H})}(\hat{c}_h, {\gamma}_h)$} of the model $\Tilde{\cM}(\pi^{1:H})$, which are defined as $\PP_h^{\pi^h_{1:h-1},  \cG}(z_{h+1}\given \hat{c}_h, {\gamma}_h)$ and $\EE^{\cG}_{\pi^h_{1:h-1}}[r_{i, h}(s_h, a_h)\given \hat{c}_h, {\gamma}_h]$, respectively. The challenge here compared with the single-agent learning problem \citep{golowich2022learning} is that although ${\gamma}_h$ serves as the actions for the approximate game $\Tilde{\cM}(\pi^{1:H})$, it is not possible to enumerate all possible actions, since ${\gamma}_h$ {in general lies in continuous spaces}, and even if we only consider   \emph{deterministic}  ${\gamma}_h$, the number of all possible mappings from the private information to the real actions in $ \cG$ is still of the order $A^{P_h}$. Therefore, learning $\PP_h^{\Tilde{\cM}(\pi^{1:H}), z}(z_{h+1}\given \hat{c}_h, {\gamma}_h)$ by enumerating all possible $\hat{c}_h$ and ${\gamma}_h$ is not statistically efficient. To circumvent this issue, we observe the fact that for $\PP_h^{\Tilde{\cM}(\pi^{1:H}), z}(z_{h+1}\given \hat{c}_h, {\gamma}_h)$, it holds that
\[\PP_{h}^{\pi^{h}_{1:h-1},  \cG}(z_{h+1}\given\hat{c}_h, {\gamma}_h) = \sum_{\substack{p_h, a_h, o_{h+1}:\\\chi_{h+1}(p_h, a_h, o_{h+1})=z_{h+1}}}\PP_{h}^{\pi^{h}_{1:h-1},  \cG}(p_h, a_h, o_{h+1}\given \hat{c}_h, {\gamma}_h),\]
where we recall $\chi_{h+1}$ in Assumption \ref{evo}. Further, notice the decomposition for $\PP_{h}^{\pi^{h}_{1:h-1}, \cG}(p_h, a_h, o_{h+1}\given \hat{c}_h, {\gamma}_h)$:
\$
&\PP_{h}^{\pi^{h}_{1:h-1},  \cG}(p_h, a_h, o_{h+1}\given \hat{c}_h, {\gamma}_h)  = \PP_{h}^{\pi^h_{1:h-1},  \cG}(p_h\given \hat{c}_h)\prod_{i=1}^n\gamma_{i, h}(a_{i, h}\given p_{i, h})\PP_{h}^{\pi^h_{1:h-1},  \cG}(o_{h+1}\given \hat{c}_h, p_{h}, a_{h}).
\$
Therefore, it suffices to learn $\PP_{h}^{\pi^{h}_{1:h-1},  \cG}(p_h\given \hat{c}_h)$ and $\PP_{h}^{\pi^h_{1:h-1},  \cG}(o_{h+1}\given \hat{c}_h, p_{h}, a_{h})$. Similarly for $\hat{r}^{\Tilde{\cM}(\pi^{1:H})}$, it holds that
\[
\hat{r}^{\Tilde{\cM}(\pi^{1:H})}_{i, h}(\hat{c}_h, \gamma_h)=\sum_{p_h, a_h}\PP_{h}^{\pi^h_{1:h-1},  \cG}(p_h\given \hat{c}_h)\prod_{j=1}^n\gamma_{j, h}(a_{j, h}\given p_{j, h})r_{i, h}^{\pi^h_{1:h-1}}(\hat{c}_h, p_h, a_h),
\]
where we define $r_{i, h}^{\pi^h_{1:h-1}}(\hat{c}_h, p_h, a_h):=\EE_{\pi^h_{1:h-1}}^\cG[r_{i, h}(s_h, a_h)\given \hat{c}_h, p_{h}, a_h]$. Formally, the following algorithm learns an approximation $\hat{\cM}(\pi^{1:H})$ of $\Tilde{\cM}(\pi^{1:H})$.
The algorithm for constructing the approximation enjoys the following guarantee.

\begin{lemma}\label{lemma:estimation}
    Fix $\delta_1, \zeta_1,\zeta_2, \theta_1, \theta_2> 0$. {For Algorithm \ref{alg:construct_mg},} suppose for all $h\in [H]$, $\pi^{h}\in\Delta(\Pi^{\text{det}})$ satisfies the conditions {for $\pi^\prime$} of Lemma \ref{lemma:low_visitation}, then as long as $N_0$ in Algorithm \ref{alg:construct_mg} satisfies 
    \[N_0\ge \max \left\{\frac{C(\max_h P_h + \log\frac{4H \max_h\hat{C}_h}{\delta_1})}{\zeta_1 \theta_1^2}, \frac{CA(O + \log\frac{4H\max_h(\hat{C}_hP_h)A}{\delta_1})}{\zeta_2\theta_2^2}\right\}
    \]  for some sufficiently large constant $C$, then with probability at least $1 - \delta_1$, the following holds:  
    \begin{itemize}
        \item For all $h\in [H]$, $\hat{c}_h\not\in \cC_{h,\zeta_1}^{\text{low}}(\pi^h)$, we have that
        \begin{equation}
        	\label{eq:concentrate}
        \sum_{p_h}\left|\PP_{h}^{\hat{\cM}(\pi^{1:H})}(p_h\given\hat{c}_h) - \PP_{h}^{\pi^h_{1:h-1},  \cG}(p_h\given\hat{c}_h)\right|\le  \theta_1.
        \end{equation}
        \item For all $h\in [H]$, $(\hat{c}_h, p_h)\not\in \cV_{h,\zeta_2}^{\text{low}}(\pi^h)$, $a_h\in \cA$, we have that
        \#
        \label{eq:concentrate_2}
        \sum_{o_{h+1}}\left|\PP_{h}^{\hat{\cM}(\pi^{1:H})}(o_{h+1}\given\hat{c}_h, p_h, a_h) - \PP_{h}^{\pi^h_{1:h-1},  \cG}(o_{h+1}\given\hat{c}_h, p_h, a_h)\right|&\le \theta_2,\\
      \label{eq:concentrate_3}
        \left|\hat{r}_{i, h}^{\hat{\cM}(\pi^{1:H})}(\hat{c}_h, p_h, a_h)-r_{i, h}^{\pi^{1:H}}(\hat{c}_h, p_h, a_h) \right|&\le \theta_2.
        \#

    \end{itemize}
    We refer to the two bullets above as event $\cE_1$.
\end{lemma}
\begin{proof}
    We will prove Equation \eqref{eq:concentrate} first. Note that for any trajectory $k$ of Algorithm \ref{alg:construct_mg}, the distribution of $p_h^k$ conditioned on $\hat{c}_h^k$ is exactly $\PP_{h}^{\pi^h_{1:h-1},  \cG}(\cdot\given\hat{c}_h^k)$.

    Now consider any $\hat{c}_h\not \in \cC_{h,\zeta_1}^{\text{low}}(\pi^h)$. By the Chernoff bound, with probability at least $1-\exp(-\frac{\zeta_1 N_0}{8})$, there are at least $\frac{\zeta_1 N_0}{2}$ trajectories  {indexed by the set $\cK^1\subseteq[N_0]$}, such that for any $k\in \cK^1$, $\operatorname{Compress}_h(f_h(a_{1:h-1}^k, o_{1:h}^k)) = \hat{c}_h$. By the folklore theorem of learning a discrete probability distribution \citep{canonne2020short}, with probability {at least} $1-p^\prime$, \eqref{eq:concentrate} holds as long as
    \begin{equation}\label{eq:learn_dis}
    \frac{\zeta_1 N_0}{2}\ge \frac{C(P_h + \log\frac{1}{p^\prime})}{\theta_1^2},
    \end{equation}
    for some constant $C>1$. By a union bound over all possible $h\in [H]$  and $\hat{c}_h\in \hat{\cC}_h$, \eqref{eq:concentrate} holds with probability at least
    \[
    1-H\max_h\hat{C}_h\exp(-\frac{\zeta_1 N_0}{8}) - H\max_h\hat{C}_h p^\prime.
    \]
    Now set $p^\prime = \frac{\delta_1}{4H\max_h \hat{C}_h}$ and it is  easy to verify that \eqref{eq:learn_dis} holds since $N_0\ge \frac{C(\max_h P_h + \log\frac{4H \max_h\hat{C}_h}{\delta_1})}{\zeta_1 \theta_1^2}$. Furthermore, as long as $C$ is sufficiently large, we have that $H\max_h\hat{C}_h\exp(-\frac{\zeta_1 N_0}{8})\le \frac{\delta_1}{4}$. Therefore, we proved that with probability at least $1-\frac{\delta_1}{2}$, \Cref{eq:concentrate} holds for all $h\in [H]$, and $\hat{c}_h\not\in \cC_{h, \zeta_1}^{\operatorname{low}}(\pi^h)$.

    Similarly, consider any trajectory $k$, the distribution of $o_{h+1}$ conditioned on {any} $(\hat{c}_h, p_h, a_h)$ is exactly $\PP_{h}^{\pi^h_{1:h-1},  \cG}(\cdot\given\hat{c}_h, p_h, a_h)$. Now consider any $(\hat{c}_h, p_h)\not\in \cV_{h, \zeta_2}^{\operatorname{low}}(\pi^h)$ and $a_h\in \cA$. Note that due to the assumption on  $\pi^h$ {that takes uniform random actions after step $h-L$}, it holds that $\PP_h^{\pi^h,  \cG}(\hat{c}_h, p_h, a_h) = \PP_h^{\pi^h,  \cG}(\hat{c}_h, p_h) \PP_h^{\pi^h,  \cG}(a_h\given \hat{c}_h, p_h)\ge \frac{\zeta_2}{A}$. By the Chernoff bound, with probability at least $1-\exp(-\frac{\zeta_2 N_0}{8A})$, there are at least $\frac{\zeta_2 N_0}{2A}$ trajectories {indexed by the set $\cK^2\subseteq[N_0]$}, such that for any $k\in\cK^2$, $\operatorname{Compress}_h(f_h(a_{1:h-1}^k, o_{1:h}^k)) = \hat{c}_h, g_{h}(a_{1:h-1}^k, o_{1:h}^k) = p_h, a_{h}^k=a_h$. Again, with probability at least $1-p^\prime$, \eqref{eq:concentrate_2} and \eqref{eq:concentrate_3} hold as long as
    \[
    \frac{\zeta_2 N_0}{2A}\ge \frac{C(O + \log\frac{1}{p^\prime})}{\theta_2^2},
    \]
    for some constant $C\ge 1$. By a union bound over all possible $h\in [H]$, $\hat{c}_h, p_h, a_h$, \eqref{eq:concentrate_2} and \eqref{eq:concentrate_3} hold with probability at least
    \[
    1-H\max_h (\hat{C}_h P_h)A\exp(-\frac{\zeta_2 N_0}{8A}) - H\max_h (\hat{C}_h P_h)A p^\prime.
    \]
    Now we set $p^\prime = \frac{\delta_1}{4H\max_h (\hat{C}_h P_h) A}$. Then since $N_0>\frac{CA(O + \log\frac{4H\max_h(\hat{C}_hP_h)A}{\delta_1})}{\zeta_2\theta_2^2}$, it holds that 
    $H\max_h (\hat{C}_h P_h)A\exp(-\frac{\zeta_2 N_0}{8A})\le \frac{\delta_1}{4}$ and $H\max_h (\hat{C}_h P_h)A p^\prime\le \frac{\delta_1}{4}$ as long as the constant $C$ is sufficiently large. Therefore, we conclude that with probability at least $1-\frac{\delta_1}{2}$, \Cref{eq:concentrate_2} holds for all $h\in [H]$, $\hat{c}_h\in \hat{\cC}_h$, $p_h\in  \cP_h$, $a_h\in \cA$. 
    Finally, by a union bound, we conclude the proof.
\end{proof}

With the previous lemma, the next step is to bound the two important quantities in Definition \ref{def:ais}. In the following discussion, we will use the shorthand notation  $\Tilde{\cM}$ for $\Tilde{\cM}(\pi^{1:H})$, and $\hat{\cM}$ for $\hat{\cM}(\pi^{1:H})$. 

\begin{lemma}\label{lemma:epsilon_apx}
    Under the event $\cE_1$ in Lemma \ref{lemma:estimation}, for any $h\in [H]$, policy $\pi\in\Delta(\Pi^{\mathrm{det}})$, and prescription ${\gamma}_h\in\Gamma_h$, it holds that  
    \#
    \nonumber&\EE_{a_{1:h-1}, o_{1:h}\sim \pi}^{ \cG}\sum_{z_{h+1}}\Big|\PP_{h}^{\Tilde{\cM}, z}(z_{h+1}\given \hat{c}_h, {\gamma}_h) - \PP_{h}^{\hat{\cM}, z}(z_{h+1}\given\hat{c}_h, {\gamma}_h)\Big|\\ 
    &\qquad\le \theta_1 + 2AP_h\frac{\zeta_2}{\zeta_1} + AP_h\theta_2 + \frac{A^{2\hat{L}}O^{\hat{L}}\zeta_1}{\phi} + \bm{1}[h>\hat{L}]\cdot 2\cdot d_{\cS, h-\hat{L}}^{ \pi, \cG}(\cU_{\phi, h-\hat{L}}^ \cG(\pi^h)),\\
    \nonumber&\EE_{a_{1:h-1}, o_{1:h}\sim \pi}^{ \cG}\Big|\hat{r}_{i, h}^{\tilde{\cM}}(\hat{c}_h, {\gamma}_h)-\hat{r}_{i, h}^{\hat{\cM}}(\hat{c}_h, {\gamma}_h)\Big|\\
    \label{eq:r_ais} 
    &\qquad\le \theta_1 + 2AP_h\frac{\zeta_2}{\zeta_1} + AP_h\theta_2 + \frac{A^{2\hat{L}}O^{\hat{L}}\zeta_1}{\phi} + \bm{1}[h>\hat{L}]\cdot 2\cdot d_{\cS, h-\hat{L}}^{ \pi, \cG}(\cU_{\phi, h-\hat{L}}^ \cG(\pi^h)).
    \#
\end{lemma}
\begin{proof}
It suffices to only consider $\pi\in\Pi^{\text{det}}$, since if the statement holds for any $\pi\in\Pi^{\text{det}}$, it will hold for any $\pi\in\Delta(\Pi^{\text{det}})$ also. Under the event $\cE_1$, consider any $\hat{c}_h\not\in \cC_{h,\zeta_1}^{\text{low}}(\pi^h)$ and $\gamma_h\in\Gamma_h$:
\$
&\sum_{p_h, a_h, o_{h+1}}\left|\PP_{h}^{\Tilde{\cM}}(p_h, a_h, o_{h+1}\given \hat{c}_h, {\gamma}_h) - \PP_{h}^{\hat{\cM}}(p_h, a_h,o_{h+1}\given \hat{c}_h, {\gamma}_h)\right| \\
&=\sum_{p_h, a_h, o_{h+1}}\Big|\PP_{h}^{\pi^h,  \cG}(p_h\given \hat{c}_h)\prod_{i=1}^n\gamma_{i, h}(a_{i, h}\given p_{i, h})\PP_{h}^{\pi^h,  \cG}(o_{h+1}\given \hat{c}_h, p_{h}, a_{h}) - \PP_{h}^{\hat{\cM}}(p_h\given \hat{c}_h)\prod_{i=1}^n\gamma_{i, h}(a_{i, h}\given p_{i, h})\PP_{h}^{\hat{\cM}, o}(o_{h+1}\given\hat{c}_h, p_{h}, a_{h})\Big|\\
&\le \sum_{p_h, a_h, o_{h+1}}\prod_{i=1}^{n}\gamma_{i, h}(a_{i, h}\given p_{i, h})\left|\PP_{h}^{\pi^h,  \cG}(p_h\given\hat{c}_h) - \PP_{h}^{\hat{\cM}}(p_h\given\hat{c}_h)\right| + \\
&\qquad\prod_{i=1}^{n}\gamma_{i, h}(a_{i, h}\given p_{i, h})\PP_{h}^{\pi^h,  \cG}(p_h\given \hat{c}_h)\left|\PP_{h}^{\pi^h,  \cG}(o_{h+1}\given \hat{c}_h, p_h, a_h) - \PP_{h}^{\hat{\cM}, o}(o_{h+1}\given \hat{c}_h, p_h, a_h)\right|\\ 
&\le \|\PP_{h}^{\pi^h,  \cG}(\cdot\given\hat{c}_h) - \PP_{h}^{\hat{\cM}}(\cdot\given\hat{c}_h)\|_1  +  \sum_{p_h, a_h}\PP_{h}^{\pi^h,  \cG}(p_h\given \hat{c}_h)\| \PP_{h}^{\pi^h,  \cG}(\cdot\given\hat{c}_h, p_h, a_h)-\PP_{h}^{\hat{\cM}}(\cdot\given\hat{c}_h, p_h, a_h) \|_1\\
&\le \left(\sum_{p_h: \PP_{h}^{\pi^h,  \cG}(p_h\given\hat{c}_h)\le \frac{\zeta_2}{\zeta_1}} + \sum_{p_h: \PP_{h}^{\pi^h,  \cG}(p_h\given \hat{c}_h)> \frac{\zeta_2}{\zeta_1}}\right)\sum_{a_h}\PP_{h}^{\pi^h,  \cG}(p_h\given  \hat{c}_h)\Big\| \PP_{h}^{\pi^h,  \cG}(\cdot\given\hat{c}_h, p_h, a_h)-\PP_{h}^{\hat{\cM}}(\cdot\given\hat{c}_h, p_h, a_h) \Big\|_1+ O\theta_1\\
&\le \theta_1 + 2AP_h\frac{\zeta_2}{\zeta_1} + AP_h\theta_2,
\$
where the last inequality comes from the fact that if $\hat{c}_h\not\in \cC_{h,\zeta_1}^{\text{low}}(\pi^h)$ and $\PP_{h}^{\pi^h,  \cG}(p_h\given \hat{c}_h)> \frac{\zeta_2}{\zeta_1}$, then $(\hat{c}_h, p_h)\not \in \cV_{h,\zeta_2}^{\text{low}}(\pi^h)$. Finally, for any policy $\pi\in\Pi^{\text{det}}$, by taking expectations over $\hat{c}_h$, we conclude that
\$
&\EE_{a_{1:h-1}, o_{1:h}\sim \pi}^{ \cG}\sum_{p_h, a_h, o_{h+1}}\left|\PP_{h}^{\Tilde{\cM}}(p_h, a_h, o_{h+1}\given \hat{c}_h, {\gamma}_h) - \PP_{h}^{\hat{\cM}}(p_h, a_h,o_{h+1}\given\hat{c}_h, {\gamma}_h)\right|\\ 
&\le \theta_1 + 2AP_h\frac{\zeta_2}{\zeta_1} + AP_h\theta_2 + 2\cdot d_{\cC, h}^{ \pi, \cG}(\cC_{h,\zeta_1}^{\text{low}}(\pi^h))\\
&\le \theta_1 + 2AP_h\frac{\zeta_2}{\zeta_1} + AP_h\theta_2 + \frac{A^{2\hat{L}}O^{\hat{L}}\zeta_1}{\phi} + \bm{1}[h>\hat{L}]\cdot 2\cdot d_{\cS, h-\hat{L}}^{ \pi, \cG}(\cU_{\phi, h-\hat{L}}^ \cG(\pi^h)),
\$
where the last step comes from Lemma \ref{lemma:low_visitation}. By noticing that after marginalization, the total variation will not increase, we proved the first inequality.

Similarly, for the approximate reward, it holds that
\$
&\left|\hat{r}_{i, h}^{\tilde{\cM}}(\hat{c}_h, {\gamma}_h)-\hat{r}_{i, h}^{\hat{\cM}}(\hat{c}_h, {\gamma}_h)\right| \\
&=\Big|\sum_{p_h, a_h}\PP_{h}^{\pi^h,  \cG}(p_h\given \hat{c}_h)\prod_{i=1}^n\gamma_{i, h}(a_{i, h}\given p_{i, h})r_{i, h}^{\pi^h_{1:h-1}}(\hat{c}_h, p_{h}, a_{h}) - \PP_{h}^{\hat{\cM}}(p_h\given \hat{c}_h)\prod_{i=1}^n\gamma_{i, h}(a_{i, h}\given p_{i, h})\hat{r}^{\hat{\cM}}_{i, h}(\hat{c}_h, p_{h}, a_{h})\Big|\\
&\le \sum_{p_h, a_h}\prod_{i=1}^{n}\gamma_{i, h}(a_{i, h}\given p_{i, h})\left|\PP_{h}^{\pi^h,  \cG}(p_h\given\hat{c}_h) - \PP_{h}^{\hat{\cM}}(p_h\given\hat{c}_h)\right| + \\
&\qquad\prod_{i=1}^{n}\gamma_{i, h}(a_{i, h}\given p_{i, h})\PP_{h}^{\pi^h,  \cG}(p_h\given \hat{c}_h)\left|r_{i, h}^{\pi^h_{1:h-1}}( \hat{c}_h, p_{h}, a_{h})-\hat{r}^{\hat{\cM}}_{i, h}(\hat{c}_h, p_{h}, a_{h})\right|\\ 
&\le \|\PP_{h}^{\pi^h,  \cG}(\cdot\given\hat{c}_h) - \PP_{h}^{\hat{\cM}}(\cdot\given\hat{c}_h)\|_1 +  \sum_{p_h, a_h}\PP_{h}^{\pi^h,  \cG}(p_h\given \hat{c}_h)\left|r_{i, h}^{\pi^h_{1:h-1}}( \hat{c}_h, p_{h}, a_{h})-\hat{r}^{\hat{\cM}}_{i, h}(\hat{c}_h, p_{h}, a_{h})\right|\\
&\le  \left(\sum_{p_h: \PP_{h}^{\pi^h,  \cG}(p_h\given\hat{c}_h)\le \frac{\zeta_2}{\zeta_1}} + \sum_{p_h: \PP_{h}^{\pi^h,  \cG}(p_h\given \hat{c}_h)> \frac{\zeta_2}{\zeta_1}}\right)\sum_{a_h}\PP_{h}^{\pi^h,  \cG}(p_h\given  \hat{c}_h) \Big|r_{i, h}^{\pi^h_{1:h-1}}( \hat{c}_h, p_{h}, a_{h})-\hat{r}^{\hat{\cM}}_{i, h}(\hat{c}_h, p_{h}, a_{h})\Big| + O\theta_1\\
&\le \theta_1 + 2AP_h\frac{\zeta_2}{\zeta_1} + AP_h\theta_2.
\$
Again, by taking expectations over $\hat{c}_h$, we proved the second inequality.
\end{proof}

Finally, we are ready to prove Theorem \ref{thm:main_thm_learning} by relating $\cG$ and $\hat{\cM}(\pi^{1:H})$ through the intermediate $\Tilde{\cM}(\pi^{1:H})$.

\begin{prevproof}{thm:main_thm_learning}
    In the following proof, we will use $\Tilde{\cM}$ for $\Tilde{\cM}(\pi^{1:H})$ and $\hat{\cM}$ for $\hat{\cM}(\pi^{1:H})$. Note that for $\epsilon_{r}(\hat{\cM})$, it holds that
    \$
    \epsilon_{r}(\hat{\cM}) &= \max_{i, h} \max_{\pi\in\Pi^{\text{det}}, {\gamma}_h}\EE_{a_{1:h-1}, o_{1:h}\sim \pi}^{ \cG}|\EE^\cG[r_{i, h}(s_h, a_h)\mid c_h, {\gamma}_h]-\hat{r}^{\hat{\cM}}_{i, h}(\hat{c}_h, {\gamma}_h)|\\
    &\le \max_{i, h} \max_{\pi\in\Pi^{\text{det}}, {\gamma}_h} \EE_{a_{1:h-1}, o_{1:h}\sim \pi}^{ \cG}|\EE^\cG[r_{i, h}(s_h, a_h)\mid c_h, {\gamma}_h]-\hat{r}^{\tilde{\cM}}_{i, h}(\hat{c}_h, {\gamma}_h)|\\
    &\quad+\max_{i, h} \max_{\pi\in\Pi^{\text{det}}, {\gamma}_h} \EE_{a_{1:h-1}, o_{1:h}\sim \pi}^{ \cG}|\hat{r}^{\tilde{\cM}}_{i, h}(\hat{c}_h, {\gamma}_h)-\hat{r}^{\hat{\cM}}_{i, h}(\hat{c}_h, {\gamma}_h)|\\
    &\le \epsilon_r(\pi^{1:H}) + \epsilon_{apx}(\pi^{1:H}, \hat{L}, \zeta_1, \zeta_2, \theta_1, \theta_2, \phi),
    \$ 
where the last step comes from Lemma \ref{lemma:epsilon_apx}. Similarly, for $\epsilon_z(\hat{\cM})$, it holds that
\$
\epsilon_z(\hat{\cM}) &= \max_h \max_{\pi\in\Pi^{\text{det}}, {\gamma}_h} \EE_{a_{1:h-1}, o_{1:h}\sim\pi}^ \cG||\PP_h^ \cG(\cdot\given c_h, {\gamma}_h) - \PP_h^{\hat{\cM}, z}(\cdot\given\hat{c}_h, {\gamma}_h)||_1\\
&\le \max_h\max_{\pi\in\Pi^{\text{det}}, {\gamma}_h} \EE_{a_{1:h-1}, o_{1:h}\sim\pi}^ \cG||\PP_h^ \cG(\cdot\given c_h, {\gamma}_h) - \PP_h^{\Tilde{\cM}, z}(\cdot\given\hat{c}_h, {\gamma}_h)||_1 \\
&\qquad + \max_h\max_{\pi\in\Pi^{\text{det}}, {\gamma}_h} \EE_{a_{1:h-1}, o_{1:h}\sim\pi}^ \cG||\PP_h^{\Tilde{\cM}, z}(\cdot\given c_h, {\gamma}_h) - \PP_h^{\hat{\cM}, z}(\cdot\given\hat{c}_h, {\gamma}_h)||_1\\
	&\le \epsilon_z(\pi^{1:H}) + \epsilon_{apx}(\pi^{1:H}, \hat{L}, \zeta_1, \zeta_2, \theta_1, \theta_2, \phi),
\$
where the last step again comes from  Lemma \ref{lemma:epsilon_apx}. Therefore, with Lemma \ref{lemma:v_diff} and Theorem \ref{thm:struc}, we proved \Cref{thm:main_thm_learning}.  
\end{prevproof}
\subsection{Proof of \Cref{thm:learning_example_short}}\label{subsec:learning_proof}

Until now, we have not considered the relationship between $\Tilde{\cM}(\pi^{1:H})$ and $ \cG$, which will necessarily depend on the choice of approximate common information $\hat{c}_h$ and $\pi^{1:H}$. 
For planning, we have seen how to construct an approximate common information  $\hat{c}_h$ using finite memory. Similarly, here we will also show how to construct $\hat{c}_h$ with finite memory so that $\Tilde{\cM}(\pi^{1:H})$ is a good approximation of $ \cG$. {In the following discussions, we shall use another important policy-dependent approximate belief $\Tilde{\bb}^{\pi}_{h}(\cdot) := \bb_{h}^{\mathrm{apx},  \cG}(\cdot;d_{\cS, h-L}^{\pi,  \cG}) 
$.} We first introduce the following important lemmas.

\begin{lemma}\label{lemma:approx_belief}
There is a constant $C\ge 1$ so that the following holds. If Assumption
\ref{observa} holds, then for any $\epsilon, \phi> 0, L\in \mathbb{N}$ so that $L\ge C\gamma^{-4}\log (\frac{1}{\epsilon \phi})$, it holds that for any policies $ \pi,\pi^\prime\in\Delta(\Pi^{\mathrm{det}})$,
\$
\mathbb{E}_{ \pi^\prime}^{\mathcal{G}}&\left\|\bb_h\left(a_{1: h-1}, o_{1: h}\right)-\widetilde{\bb}_h^{\pi}\left(a_{h-L: h-1}, o_{h-L+1: h}\right)\right\|_1 \leq \epsilon+\bm{1}[h>L] \cdot 6 \cdot d_{\cS, h-L}^{\pi^\prime, \mathcal{G}}\left(\mathcal{U}_{\phi, h-L}^{\mathcal{G}}\left(\pi\right)\right),\\
\mathbb{E}_{\pi^\prime}^{\mathcal{G}}&\left\|\bb_h\left(a_{1: h-1}, o_{1: h-1}\right)-\widetilde{\bb}_h^{\pi}\left(a_{h-L: h-1}, o_{h-L+1: h-1}\right)\right\|_1 \leq \epsilon+\bm{1}[h>L] \cdot 6 \cdot d_{\cS, h-L}^{ \pi^\prime, \mathcal{G}}\left(\mathcal{U}_{\phi, h-L}^{\mathcal{G}}\left(\pi\right)\right),\\
\mathbb{E}_{\pi^\prime}^{\mathcal{G}}&\Big\|\bb_h\left(a_{1: h-1}, o_{1: h-1}, o_{1, h}\right)-\widetilde{\bb}_h^{\pi}\Big(a_{h-L: h-1}, o_{h-L+1: h-1}, o_{1, h}\Big)\Big\|_1 \leq \epsilon+\bm{1}[h>L] \cdot 6 \cdot d_{\cS, h-L}^{\pi^\prime, \mathcal{G} }\left(\mathcal{U}_{\phi, h-L}^{\mathcal{G}}\left(\pi\right)\right).
\$

Furthermore, for any finite domain $Y$, conditional probability $q(y\given s)$, and the posterior update operator $F^q:\Delta(\cS)\rightarrow\Delta(\cS)$ as defined in Lemma \ref{lemma:poster}, it holds that
\[
\EE^\cG_{\pi^\prime}\EE_{y\sim q\cdot \bb_{h}(a_{1:h-1}, o_{1:h})}|| F^q(\bb_{h}(a_{1:h-1}, o_{1:h}); y) - F^q(\bb_{h}^{\prime}(a_{h-L:h-1}, o_{h-L+1:h}); y)||_1\le \epsilon.
\]

\end{lemma}
\begin{proof}
    It directly follows from our Theorem \ref{noah_theo}, and Lemma 12.2 of \cite{golowich2022learning}.  
\end{proof}
The lemma shows that if we use the $d_{\cS, h-\hat{L}}^{ \pi, \cG}$ instead of a $\operatorname{Unif}(\cS)$ as the prior, the approximate belief will suffer from an additional error term $d_{\cS, h-L}^{\pi^\prime, \mathcal{G}}\left(\mathcal{U}_{\phi, h-L}^{\mathcal{G}}\left(\pi\right)\right)$. The following lemma shows that there already exists an efficient algorithm for finding $\pi$ to minimize $d_{\cS, h-L}^{\pi^\prime, \mathcal{G}}\left(\mathcal{U}_{\phi, h-L}^{\mathcal{G}}\left(\pi\right)\right)$.
\begin{lemma}\label{lemma:explore_s}
    Given $\alpha, \beta> 0$, 
    $\hat{L}\ge C\frac{\log (H S O /(\alpha \gamma))}{\gamma^4}$, 
    and $\phi = \frac{\alpha \gamma^2}{C^3 H^{10} S^5 O^4}$ for some constant $C>0$. There exists an algorithm \texttt{BaSeCAMP} (Algorithm 3 of \cite{golowich2022learning}) with both computation and sample complexity bounded by $(O A)^{\hat{L}} \log (\frac{1}{\beta})$, outputting $K=2HS$ groups of policies $\{\pi^{{1:H}, j}\}_{j=1}^{K}$, where $\pi^{h, j}\in\Delta(\Pi^{\text{det}})$ and $\pi^{h, j}_{h^\prime} = \operatorname{Unif}(\cA)$ for $h^\prime\ge h-\hat{L}, j\in [K]$. It holds that with probability at least $1-\beta$, there is at least one $j^\star\in [K]$ such that for any $h>\hat{L}$, policy $\pi\in\Pi^{\mathrm{det}}$:
        \[
    d_{\cS, h-\hat{L}}^{ \pi, \cG}(\cU_{\phi, h-\hat{L}}^{ \cG}(\pi^{h, j^\star}))\le \frac{\alpha}{CH^2}.
    \]
\end{lemma}
\begin{proof}
    It follows from Theorem 3.1 in \cite{golowich2022learning}.  
\end{proof}
\begin{corollary}\label{corr:expl}
    Given $\epsilon, \delta_2> 0$, 
    $L\ge C\frac{\log (H S O /(\epsilon \gamma))}{\gamma^4}$
    , and $\phi = \frac{\epsilon \gamma^2}{C^2 H^{8} S^5 O^4}$ 
    for some constant $C>0$. There exists an algorithm \texttt{BaSeCAMP} (Algorithm 3 of \cite{golowich2022learning}) with both computation and sample complexity bounded by $N_1 = (O A)^{L} \log (\frac{1}{\delta_2})$,  outputting $K=2HS$ groups of policies $\{\pi^{{1:H}, j}\}_{j=1}^{K}$, where $\pi^{h, j}\in\Delta(\Pi^{\mathrm{det}})$ and $\pi^{h, j}_{h^\prime} = \operatorname{Unif}(\cA)$ for $h\in [H]$, $h^\prime\ge h-L$, $j\in [K]$. The following event $\cE_2$ holds with probability at least $1-\delta_2$: there is at least one $j^\star\in [K]$ such that for any $h>L$, policy $\pi^\prime\in\Delta(\Pi^{\mathrm{det}})$: 
    \$
    d_{\cS, h-L}^{ \pi^\prime, \cG}(\cU_{\phi, h-L}^{ \cG}(\pi^{h, j^\star}))&\le \epsilon.
    \$
\end{corollary}
\begin{proof}
    This is by letting $\alpha = \frac{CH^2\epsilon}{2}$, $\delta_2 = \beta$, and $L\ge \max\{C\frac{\log (\frac{1}{\epsilon\phi})}{\gamma^4}, C\frac{\log (H S O /(\alpha \gamma))}{\gamma^4}\}$ in \Cref{lemma:explore_s}.
\end{proof}

In the discussion thereafter, we will use $\Tilde{\cM}$ for $\Tilde{\cM}(\pi^{1:H, j^\star})$ and $\hat{{\cM}}$ for $\hat{\cM}(\pi^{1:H, j^\star})$, and $\hat{r}_{i, h}$ for $\hat{r}_{i, h}^{j^\star}$ interchangeably.
There is still one issue unsolved, which is that \texttt{BaSeCAMP} does not tell us which $j\in [K]$ is the $j^\star$ we want. Therefore, we have to evaluate the policies $\{\pi^{\star, j}\}_{j=1}^K$, {which are generated by running Algorithm \ref{alg:avi} on the candidate models $\{\hat{\cM}(\pi^{1:H, j})\}_{j\in [K]}$}. The policy evaluation and selection algorithm is described in Algorithm \ref{alg:selection}.  

\begin{lemma}\label{lemma:selection}
    For Algorithm \ref{alg:selection}, suppose that the $K$ groups of policies $\{\pi^{1:H, j}\}_{j=1}^{K}$ and $K$ reward functions $\{(\hat{r}^j_i)_{i=1}^n\}_{j=1}^K$ satisfy that there exists some $j^\star\in [K]$ such that for any policy $\pi\in\Pi$, $i\in [n]$, we have 
    \[
    \left|V_{i, 1}^{\pi,  \cG }(\emptyset) - V_{i, 1}^{\pi, \hat{\cM}(\pi^{1:H, j^\star})}(\emptyset)\right|\le \epsilon.
    \]
    If $N_2\ge C\frac{H^2\log\frac{K^2n}{\delta_3}}{\epsilon^2}$ for some constant $C>0$, then with probability at least $1-\delta_3$, the following event $\cE_3$ holds 
    \[
    \operatorname{NE/CE/CCE-gap}(\pi^{\star, \hat{j}})\le \operatorname{NE/CE/CCE-gap}(\pi^{\star, j^\star}) + 6\epsilon + H\epsilon_e .
    \]
\end{lemma}
\begin{proof}
For NE/CCE, note that $V_{i, 1}^{\pi_{i}^{\star, j, m}\times \pi_{-i}^{\star, j}, \hat{\cM}(\pi^{1:H, m})}(\emptyset) \ge \max_{\pi_{i}}V_{i, 1}^{\pi_i\times \pi_{-i}^{\star, j}, \hat{\cM}(\pi^{1:H, m})}(\emptyset) -H\epsilon_e $ {according to Corollary \ref{corr:ne_br}} for $m\in [K]$. By the concentration bound on the relationship between the accumulated rewards and the value function for all policies $\pi^{\star, j}, \pi_i^{\star, j, m}\times\pi_{-i}^{\star, j}$, and further a union bound {over all $i\in[n]$, $j\in[K]$, and $m\in[K]$}, with probability at least $1-\delta_3$, the following event $\cE_3$ holds for any $i\in [n], j\in [K], m\in [K]$:
\[
\left|R_{i}^j - V_{i, 1}^{\pi^{\star, j},  \cG }(\emptyset)\right|\le \epsilon,\qquad\qquad 
\left|R_{i}^{j, m} - V_{i, 1}^{\pi_i^{\star, j, m}\times \pi_{-i}^{\star, j},  \cG }(\emptyset)\right|\le\epsilon.
\]
In the following proof, we will assume the previous event holds. Define $m_{i, j}^\star \in \arg\max_m R_{i}^{j, m}$. Now we will firstly show that $\max_m R_{i}^{j, m}$ approximates the best response of $\pi_{-i}^{\star, j}$. Note that for any $i\in [n], j\in [K]$:
\[
\max_{\pi_i}V_{i, 1}^{\pi_i\times \pi_{-i}^{\star, j},  \cG }(\emptyset) - \max_m R_{i}^{j, m}\ge \max_{\pi_i}V_{i, 1}^{\pi_i\times \pi_{-i}^{\star, j},  \cG }(\emptyset) - V_{i, 1}^{\pi_i^{\star, j, m_{i, j}^\star}\times \pi_{-i}^{\star, j},  \cG }(\emptyset)-\epsilon\ge -\epsilon.
\]
On the other hand, 
\$
\max_{\pi_i}V_{i, 1}^{\pi_i\times \pi_{-i}^{\star, j},  \cG }(\emptyset) &- \max_m R_{i}^{j, m}\le \max_{\pi_i}V_{i, 1}^{\pi_i\times \pi_{-i}^{\star, j},  \cG }(\emptyset)-\max_m V_{i, 1}^{\pi_i^{\star, j, m}\times \pi_{-i}^{\star, j},  \cG }(\emptyset) +\epsilon\\
&\le \max_{\pi_i}V_{i, 1}^{\pi_i\times \pi_{-i}^{\star, j},  \cG }(\emptyset)-\max_m V_{i, 1}^{\pi_i^{\star, j, m}\times \pi_{-i}^{\star, j}, \hat{\cM}(\pi^{1:H, j^\star})}(\emptyset) +2\epsilon\\
&\le \max_{\pi_i}V_{i, 1}^{\pi_i\times \pi_{-i}^{\star, j},  \cG }(\emptyset)-V_{i, 1}^{\pi_i^{\star, j, j^\star}\times \pi_{-i}^{\star, j}, \hat{\cM}(\pi^{1:H, j^\star})   }(\emptyset) +2\epsilon\\
&\le \max_{\pi_i}V_{i, 1}^{\pi_i\times \pi_{-i}^{\star, j},  \cG }(\emptyset)-\max_{\pi_i}V_{i, 1}^{\pi_i\times \pi_{-i}^{\star, j}, \hat{\cM}(\pi^{1:H, j^\star})   }(\emptyset) +2\epsilon + H\epsilon_e\\
&\le 3\epsilon + H\epsilon_e ,
\$
where the second last step comes from Corollary   \ref{corr:ne_br} and the last step comes from the fact that the max-operator is non-expansive. Now we are ready to evaluate $\pi^{\star, \hat{j}}$:
\$
\operatorname{NE/CCE-gap}(\pi^{\star, \hat{j}}) &= \max_i \max_{\pi_i}\left(V_{i, 1}^{\pi_i\times \pi_{-i}^{\star, \hat{j}},  \cG }(\emptyset) - V_{i, 1}^{\pi^{\star, \hat{j}},  \cG }(\emptyset)\right)\\
&\le \max_i \max_{\pi_i}\left(V_{i, 1}^{\pi_i\times \pi_{-i}^{\star, \hat{j}},  \cG }(\emptyset) - R_{i}^{\hat{j}} \right) + \epsilon \le \max_i \left(\max_m R_{i}^{\hat{j}, m} - R_{i}^{\hat{j}}\right) + 4\epsilon + H\epsilon_e .
\$
Meanwhile for $\pi^{\star, j^\star}$, we have that  
\$
\operatorname{NE/CCE-gap}(\pi^{\star, j^\star}) &= \max_i \max_{\pi_i}\left(V_{i, 1}^{\pi_i\times \pi_{-i}^{\star, j^\star},  \cG }(\emptyset) - V_{i, 1}^{\pi^{\star, j^\star},  \cG }(\emptyset)\right)\\
&\ge \max_i \max_{\pi_i}\left(V_{i, 1}^{\pi_i\times \pi_{-i}^{\star, j^\star},  \cG }(\emptyset) - R_{i}^{j^\star} \right) - \epsilon\\
&\ge \max_i \left(\max_m R_{i}^{j^\star, m} - R_{i}^{j^\star}\right) - 2\epsilon.
\$
Recall the definition of $\hat{j} \in \arg\min_{j} \left(\max_{i}\max_m (R_{i}^{j, m} - R_{i}^j)\right)$, we conclude that $\operatorname{NE/CCE-gap}(\pi^{\star, \hat{j}})\le \operatorname{NE-gap}(\pi^{\star, j^\star})  + 6\epsilon +H\epsilon_e$.

For CE, note that \$V_{i, 1}^{\pi_{i}^{\star, j, m}\odot \pi_{-i}^{\star, j}, \hat{\cM}(\pi^{1:H, m})}(\emptyset) \ge \max_{\phi_{i}}V_{i, 1}^{(\phi_i\diamond \pi_{i}^{\star, j})\odot \pi_{-i}^{\star, j}, \hat{\cM}(\pi^{1:H, m})}(\emptyset) -H\epsilon_e. \$ Similarly, by a concentration bound and then a union bound, with probability at least $1-\delta_3$, the following event $\cE_3$  holds for any $i\in [n], j\in [K], m\in [K]$:
\[
\left|R_{i}^j - V_{i, 1}^{\pi^{\star, j},  \cG }(\emptyset)\right|\le \epsilon,\qquad\qquad 
\left|R_{i}^{j, m} - V_{i, 1}^{\pi_i^{\star, j, m}\odot \pi_{-i}^{\star, j},  \cG }(\emptyset)\right|\le\epsilon. 
\]
In the following proof, we will assume the previous event holds. Define $m_{i, j}^\star = \arg\max_m R_{i}^{j, m}$. Now we will firstly show that $\max_m R_{i}^{j, m}$ approximates the best strategy modification with respect to $\pi_{-i}^{\star, j}$. Note that for any $i\in [n], j\in [K]$:
\$
&\max_{\phi_i}V_{i, 1}^{(\phi_i\diamond \pi_{i}^{\star, j})\odot \pi_{-i}^{\star, j},  \cG }(\emptyset) - \max_m R_{i}^{j, m}\\
&\qquad\ge \max_{\phi_i}V_{i, 1}^{(\phi_i\diamond \pi_{i}^{\star, j})\odot \pi_{-i}^{\star, j},  \cG }(\emptyset) - V_{i, 1}^{\pi_i^{\star, j, m_{i, j}^\star}\odot \pi_{-i}^{\star, j},  \cG }(\emptyset)-\epsilon\\
&\qquad\ge -\epsilon.
\$
On the other hand, 
\$
\max_{\phi_i}&V_{i, 1}^{(\phi_i\diamond \pi_{i}^{\star, j})\odot \pi_{-i}^{\star, j},  \cG }(\emptyset) - \max_m R_{i}^{j, m}\\
&\le \max_{\phi_i}V_{i, 1}^{(\phi_i\diamond \pi_{i}^{\star, j})\odot \pi_{-i}^{\star, j},  \cG }(\emptyset)-\max_m V_{i, 1}^{\pi_i^{\star, j, m}\odot \pi_{-i}^{\star, j},  \cG }(\emptyset) +\epsilon\\
&\le \max_{\phi_i}V_{i, 1}^{(\phi_i\diamond \pi_{i}^{\star, j})\odot \pi_{-i}^{\star, j},  \cG }(\emptyset)-\max_m V_{i, 1}^{\pi_i^{\star, j, m}\odot \pi_{-i}^{\star, j}, \hat{\cM}(\pi^{1:H, j^\star})}(\emptyset) +2\epsilon\\
&\le \max_{\phi_i}V_{i, 1}^{(\phi_i\diamond \pi_{i}^{\star, j})\odot \pi_{-i}^{\star, j},  \cG }(\emptyset)-V_{i, 1}^{\pi_i^{\star, j, j^\star}\odot \pi_{-i}^{\star, j}, \hat{\cM}(\pi^{1:H, j^\star})}(\emptyset) +2\epsilon\\
&\le \max_{\phi_i}V_{i, 1}^{(\phi_i\diamond \pi_{i}^{\star, j})\odot \pi_{-i}^{\star, j},  \cG }(\emptyset)-\max_{\phi_i}V_{i, 1}^{(\phi_i\diamond \pi_{i}^{\star, j})\odot \pi_{-i}^{\star, j}, \hat{\cM}(\pi^{1:H, j^\star})}(\emptyset) +2\epsilon + H\epsilon_e\\
&\le 3\epsilon + H\epsilon_e ,
\$
where the second last step comes from Corollary \ref{corr:ce_br} and the last step comes from the fact that the max-operator is non-expansive. Now we are ready to evaluate $\pi^{\star, \hat{j}}$:
\$
\operatorname{CE-gap}(\pi^{\star, \hat{j}}) &= \max_i \max_{\phi_i}\left(V_{i, 1}^{(\phi_i\diamond \pi_{i}^{\star, \hat{j}})\odot \pi_{-i}^{\star, \hat{j}},  \cG }(\emptyset) - V_{i, 1}^{\pi^{\star, \hat{j}},  \cG }(\emptyset)\right)\\
&\le \max_i \max_{\phi_i}\left(V_{i, 1}^{(\phi_i\diamond \pi_i^{\star, \hat{j}})\odot \pi_{-i}^{\star, \hat{j}},  \cG }(\emptyset) - R_{i}^{\hat{j}} \right) + \epsilon\\
&\le \max_i \left(\max_m R_{i}^{\hat{j}, m} - R_{i}^{\hat{j}}\right) + 4\epsilon + H\epsilon_e .
\$
Meanwhile for $\pi^{\star, \hat{j}}$, we have that
\$
\operatorname{CE-gap}(\pi^{\star, j^\star}) &= \max_i \max_{\phi_i}\left(V_{i, 1}^{(\phi_i\diamond \pi_i^{\star, j^\star})\odot \pi_{-i}^{\star, j^\star},  \cG }(\emptyset) - V_{i, 1}^{\pi^{\star, j^\star},  \cG }(\emptyset)\right)\\
&\ge \max_i \max_{\phi_i}\left(V_{i, 1}^{(\phi_i\diamond \pi_i^{\star, j^\star})\odot \pi_{-i}^{\star, j^\star},  \cG }(\emptyset) - R_{i}^{j^\star} \right) - \epsilon\\
&\ge \max_i \left(\max_m R_{i}^{j^\star, m} - R_{i}^{j^\star}\right) - 2\epsilon.
\$
Recall the definition of $\hat{j} = \arg\min_{j} \left(\max_{i}\max_m (R_{i}^{j, m} - R_{i}^j)\right)$, we conclude that $\operatorname{CE-gap}(\pi^{\star, \hat{j}})\le \operatorname{CE-gap}(\pi^{\star, j^\star})  + 6\epsilon + H\epsilon_e$.
\end{proof}

We put together the entire learning procedure in Algorithm \ref{alg:learning}. {Before diving into the examples in \Cref{sec:example}, the proof for the first part of Theorem \ref{thm:main_learning_full} follows from the fact that both the computation and sample complexities depend on $\max_h C_h$ and $\max_h P_h$. Therefore, if we can find $\pi^{1:H}$ and $\operatorname{Compress}_h$ for $h\in [H]$ such that the relevant errors are minimized while $\max_h C_h$ and $\max_h P_h$ are of quasi-polynomial size, then there exists a quasi-polynomial sample and time algorithm learning $\epsilon$-NE if $\cG$ is zero-sum or cooperative and $\epsilon$-CE/CCE if $\cG$ is general-sum}. In the following discussion, we will see the sample complexity of our algorithm instantiated with specific information structures. 

\paragraph{One-step delayed information sharing.} In this case, the information structure gives $c_h=\{a_{1:h-1}, o_{1:h-1}\}$, $p_{i, h} = \{o_{i, h}\}$, $z_{h+1} = \{o_{h}, a_{h}\}$. Fix $L>0$, we define the approximate common information as $\hat{c}_h = \{a_{h-L:h-1}, o_{h-L+1:h-1}\}$. For any $\pi^{1:H}$, where $\pi^h\in\Delta(\Pi^{\text{det}})$ for $h\in [H]$, it is direct to verify that
\[\PP_h^{\Tilde{\cM}(\pi^{1:H}), c}(s_h, p_h\given \hat{c}_h) =\PP_h^{\pi^h,  \cG}(s_h, p_h\given \hat{c}_h) =\Tilde{\bb}_h^{\pi^h}(a_{h-L:h-1}, o_{h-L+1:h-1})(s_h)\OO_{h}(o_h\given s_h),
\]
where we recall the definition of $\Tilde{\bb}_h^{\pi^h}$ in \Cref{subsec:learning_proof}. Meanwhile, according to Definition \ref{def:L_main}, it is direct to verify that $\hat{L} = L$. Hereafter in the proof, we use $\Tilde{\cM}$ to denote $\Tilde{\cM}(\pi^{1:H, j^\star})$ for short. Therefore, we conclude that if $L\ge C\frac{\log (H S O /(\epsilon \gamma))}{\gamma^4}$, {by a union bound of the high probability event $\cE_1$ in Lemma \ref{lemma:estimation}, $\cE_2$ in Corollary \ref{corr:expl}, and $\cE_3$ in Lemma \ref{lemma:selection}}, with probability at least $1-\delta_1-\delta_2-\delta_3$, it holds that for any $i\in [n]$
\$
&\epsilon_{r}(\pi^{1:H, j^\star}) \\
&= \max_{i, h}\max_{\pi\in\Pi^{\text{det}}, {\gamma}_h}\EE_{a_{1:h-1}, o_{1:h}\sim \pi}^{ \cG}\left|\EE^{ \cG}[r_{i, h}(s_h, a_h)\mid c_h, {\gamma}_h]-\hat{r}^{\Tilde{\cM}}_{i, h}(\hat{c}_h, {\gamma}_h)\right|\\
&\le \max_{h}\max_{\pi\in\Pi^{\text{det}}}\EE_{a_{1:h-1}, o_{1:h}\sim \pi}^{{ \cG}}\|  \bb_h(a_{1:h-1}, o_{1:h-1})-\Tilde{\bb}_h^{\pi^{h, j^\star}}(a_{h-L:h-1}, o_{h-L+1:h-1})\|_1\\
&\le \epsilon + \max_{h}\max_{\pi\in\Pi^{\text{det}}}\bm{1}[h>L] \cdot 6 \cdot d_{\cS, h-L}^{\pi, \mathcal{G}}\left(\mathcal{U}_{\phi, h-L}^{\mathcal{G}}\left(\pi^{h, j^\star}\right)\right),
\$  
and moreover 
\$
&\epsilon_z(\pi^{1:H, j^\star}) = \max_h\max_{\pi\in\Pi^{\text{det}}, {\gamma}_h} \EE_{a_{1:h-1}, o_{1:h}\sim \pi}^{ \cG}\left\| \PP_{h}^{\cG}(\cdot\given c_h, {\gamma}_h) - \PP_{h}^{\Tilde{\cM}, z}(\cdot\given \hat{c}_h, {\gamma}_h) \right\|_1\\
&\le \max_h\max_{\pi\in\Pi^{\text{det}}, {\gamma}_h}\EE_{a_{1:h-1}, o_{1:h}\sim \pi}^{{ \cG}}\left\|  \bb_h(a_{1:h-1}, o_{1:h-1})-\Tilde{\bb}_h^{\pi^{h, j^\star}}(a_{h-L:h-1}, o_{h-L+1:h-1})\right\|_1\\
&\le \epsilon + \max_{h}\max_{\pi\in\Pi^{\text{det}}}\bm{1}[h>L] \cdot 6 \cdot d_{\cS, h-L}^{\pi, \mathcal{G}}\left(\mathcal{U}_{\phi, h-L}^{\mathcal{G}}\left(\pi^{h, j^\star}\right)\right).
\$ 
According to the choice of $\pi^{1:H, j^\star}$ and {Corollary \ref{corr:expl}}, it holds that 
\[\max_h\max_{\pi}\bm{1}[h>L] \cdot 6 \cdot d_{\cS, h-L}^{\pi, \mathcal{G}}\left(\mathcal{U}_{\phi, h-L}^{\mathcal{G}}\left(\pi^{h, j^\star}\right)\right)\le 6\epsilon.
\]
Therefore, for any $\alpha, \delta>0$,  setting 
$\epsilon=\frac{\alpha}{200(H+1)^2}$, $\theta_1 = \frac{\alpha}{200(H+1)^2O}$, $\zeta_2=\zeta_1^2$, $\theta_2 = \frac{\alpha}{200(H+1)^2A\max_h P_h}$, $\zeta_1 = \min\left\{\frac{\alpha\phi}{200(H+1)^2A^{2L}O^L}, \frac{\alpha}{400(H+1)^2A\max_h P_h}\right\}$, $\phi = \frac{\epsilon \gamma^2}{C^2 H^{8} S^5 O^4}$, $\epsilon_e = \frac{\alpha}{200H}$, $\delta_1 = \delta_2 = \delta_3 = \frac{\delta}{3}$, $\Tilde{\cM}(\pi^{1:H, j^\star})$   is an $(\epsilon_r, \epsilon_z)$-expected-approximate common information model of $ \cG$, where $\epsilon_r, \epsilon_z \le \frac{14\alpha}{200(H+1)^2}$. This leads to that $\pi^{\star, j^\star}$ is a $\frac{15\alpha}{200}$-NE{/CE/CCE}, and $|V_{i, 1}^{\pi, \cG }(\emptyset) - V_{i, 1}^{\pi, \hat{\cM}(\pi^{1:H, j^\star})   }(\emptyset)|\le \frac{15\alpha}{200}$ for any policy $\pi\in\Pi$ by Lemma \ref{lemma:v_diff}. By Lemma  \ref{lemma:selection}, $\operatorname{NE/CE/CCE-gap}(\pi^{\star, \hat{j}})\le \operatorname{NE/CE/CCE-gap}(\pi^{\star, j^\star}) + \frac{91\alpha}{200}\le\alpha$.  Finally, we are ready to analyze the computation and sample complexities of our algorithm.

\begin{theorem}
Let $\alpha, \delta, \gamma>0$. Algorithm \ref{alg:learning} given a $\gamma$-observable POSG of one-step delayed information sharing structure outputs an $\alpha$-NE \xynew{if the POSG is zero-sum or cooperative, or $\alpha$-CE/CCE if the POSG is general-sum}, with probability at least $1-\delta$, with time and sample complexities bounded by $(AO)^{C\gamma^{-4}\log\frac{SHO}{\gamma\alpha}}\log\frac{1}{\delta}$ for some universal constant $C>0$.
\end{theorem}
\begin{proof}
    Recall that $\hat{C}_h\le (OA)^{L}$, $P_h\le O$, $N_0= \max\left\{\frac{C(\max_h P_h + \log\frac{4H \max_h\hat{C}_h}{\delta_1})}{\zeta_1 \theta_1^2}, \frac{CA(O + \log\frac{4H\max_h(\hat{C}_hP_h A)}{\delta_1})}{\zeta_2\theta_2^2}\right\}$, $N_1 = (O A)^{L} \log (\frac{1}{\delta_2})$, and $N_2=C \frac{H^2\log\frac{K^2n}{\delta_3}}{\epsilon^2}$ for some constant $C> 0$\xynew{, and we have set $\delta_1=\delta_2=\delta_3=\frac{\delta}{3}$.} The total number of samples used is $KN_0 + N_1 + (K + nK^2)N_2$. Substituting the choices of parameters into $N_0$, $N_1$, and $N_2$, we proved the sample complexity. Furthermore, for time complexity, since our algorithm only calls the \texttt{BaSeCAMP} and our planning algorithm a polynomial number of times, the time complexity is also bounded by $(OA)^{C\gamma^{-4}\log\frac{SHO}{\gamma\alpha}}\log\frac{1}{\delta}$. 
\end{proof}
\paragraph{State controlled by one controller with asymmetric delay sharing.} The information structure is given as $c_{h} = \{o_{1, 1:h}, o_{2, 1:h-d}, a_{1, 1:h-1}\}$, $p_{1, h} = \emptyset$, $p_{2, h} = \{o_{2, h-d+1:h}\}$. Fix some $L> 0$, the approximate common information is constructed as $\hat{c}_h :=\{o_{1, h-d-L+1:h}, o_{2, h-d-L+1:h-d}, a_{1, h-d-L:h-1}\}$. Then for any given policy $\pi^{1:H}$, where $\pi^h\in\Delta(\Pi^{\text{det}})$, following exactly the same derivation as in \Cref{subsec:finite_memory}, it holds that
\$
&\PP_{h}^{\Tilde{\cM}(\pi^{1:H}), c}(s_h, p_h \given \hat{c}_h) = \PP_{h}^{\pi^h,  \cG}(s_h, p_h\given \hat{c}_h) \\
& \quad=\sum_{s_{h-d}}\PP^ \cG(s_h, p_h\given s_{h-d}, f_a, f_o)F^{P(\cdot\given \cdot, f_a)}(\Tilde{\bb}^{\pi^h}_{h-d}(a_{h-d-L:h-d-1}, o_{h-d-L+1:h-d}); f_o)(s_{h-d}).
\$
Meanwhile, it is direct to verify that $\hat{L} = L+d$ by Definition \ref{def:L_main}. Therefore, we conclude that if $L\ge C\frac{\log (H S O /(\epsilon \gamma))}{\gamma^4}$, {by a union bound of the high probability event $\cE_1$ in Lemma \ref{lemma:estimation}, $\cE_2$ in Corollary \ref{corr:expl}, and $\cE_3$ in Lemma \ref{lemma:selection}}, with probability at least $1-\delta_1-\delta_2-\delta_3$, it holds that for any $i\in [n]$:  
\$
&\epsilon_{r}(\pi^{1:H, j^\star}) \\
&= \max_{i, h}\max_{\pi\in\Pi^{\text{det}}, {\gamma}_h}\EE_{a_{1:h-1}, o_{1:h}\sim \pi}^{ \cG}\left|\EE^{ \cG}[r_{i, h}(s_h, a_h)\mid c_h, {\gamma}_h]-\hat{r}^{\Tilde{\cM}}_{i, h}(\hat{c}_h, {\gamma}_h)\right|\\
&\le + \max_{h}\max_{\pi\in\Pi^{\text{det}}}\EE_{a_{1:h-1}, o_{1:h}\sim \pi}^{{ \cG}}\Big\|F^{P(\cdot\given \cdot, f_a)}(\bb_{h-d}(a_{1:h-d-1}, o_{1:h-d}); f_o) \\
&\qquad\qquad\qquad\qquad\qquad\qquad- F^{P(\cdot\given \cdot, f_a)}(\Tilde{\bb}_{h-d}^{\pi^{h, j^\star}}(a_{h-d-L:h-d-1}, o_{h-d-L+1:h-d}); f_o)\Big\|_1\\
&\le \epsilon + \max_{h}\max_{\pi\in\Pi^{\text{det}}}\bm{1}[h>\hat{L}] \cdot 6 \cdot d_{\cS, h-\hat{L}}^{\pi, \mathcal{G}}\left(\mathcal{U}_{\phi, h-\hat{L}}^{\mathcal{G}}\left(\pi^{h, j^\star}\right)\right),
\$ 
and moreover
\$
&\epsilon_z(\pi^{1:H, j^\star}) = \max_h\max_{\pi\in\Pi^{\text{det}}, {\gamma}_h} \EE_{a_{1:h-1}, o_{1:h}\sim \pi}^{ \cG}\left\|\PP_{h}^{ \cG}(\cdot\given c_h, {\gamma}_h) - \PP_{h}^{\Tilde{\cM}, z}(\cdot\given c_h, {\gamma}_h) \right\|_1\\
&\le\max_h\max_{\pi\in\Pi^{\text{det}}, {\gamma}_h}\EE_{a_{1:h-1}, o_{1:h}\sim \pi^\prime}^{{ \cG}}\Big\|F^{P(\cdot\given \cdot, f_a)}(\bb_{h-d}(a_{1:h-d-1}, o_{1:h-d}); f_o) \\
&\qquad\qquad\qquad\qquad\qquad\qquad- F^{P(\cdot\given \cdot, f_a)}(\Tilde{\bb}_{h-d}^{\pi^{h, j^\star}}(a_{h-d-L:h-d-1}, o_{h-d-L+1:h-d}); f_o)\Big\|_1\\
&\le \epsilon + \max_{h}\max_{\pi\in\Pi^{\text{det}}}\bm{1}[h>\hat{L}] \cdot 6 \cdot d_{\cS, h-\hat{L}}^{\pi, \mathcal{G}}\left(\mathcal{U}_{\phi, h-\hat{L}}^{\mathcal{G}}\left(\pi^{h, j^\star}\right)\right).
\$
According to the choice of  $\pi^{1:H, j^\star}$ and Corollary \ref{corr:expl}, it holds that 
\[\max_h\max_{\pi\in\Pi^{\text{det}}}\bm{1}[h>\hat{L}] \cdot 6 \cdot d_{\cS, h-\hat{L}}^{\pi, \mathcal{G}}\left(\mathcal{U}_{\phi, h-\hat{L}}^{\mathcal{G}}\left(\pi^{h, j^\star}\right)\right)\le 6\epsilon.
\]
Therefore, for any $\alpha, \delta>0$,   setting 
$\epsilon=\frac{\alpha}{200(H+1)^2}$, $\theta_1 = \frac{\alpha}{200(H+1)^2O}$, $\zeta_2=\zeta_1^2$, $\theta_2 = \frac{\alpha}{200(H+1)^2A\max_h P_h}$, $\zeta_1 = \min\left\{\frac{\alpha\phi}{200(H+1)^2A^{2(L+d)}O^{L+d}}, \frac{\alpha}{400(H+1)^2A\max_h P_h}\right\}$, $\phi = \frac{\epsilon \gamma^2}{C^2 H^{8} S^5 O^4}$, $\epsilon_e = \frac{\alpha}{200H}$, $\delta_1 = \delta_2 = \delta_3 = \frac{\delta}{3}$, $\Tilde{\cM}(\pi^{1:H, j^\star})$ is an $(\epsilon_r, \epsilon_z)$-expected-approximate common information model of $ \cG$, where $\epsilon_r, \epsilon_z \le \frac{14\alpha}{200(H+1)^2}$. This leads to that $\pi^{\star, j^\star}$ is a $\frac{15\alpha}{200}$-NE{/CE/CCE}, and $|V_{i, 1}^{\pi, \cG}(\emptyset) - V_{i, 1}^{\pi, \hat{\cM}(\pi^{1:H, j^\star})   }(\emptyset)|\le \frac{15\alpha}{200}$ for any policy $\pi\in\Pi$ by Lemma \ref{lemma:v_diff}. By Lemma \ref{lemma:selection}, $\operatorname{NE/CE/CCE-gap}(\pi^{\star, \hat{j}})\le \operatorname{NE/CE/CCE-gap}(\pi^{\star, j^\star}) + \frac{91\alpha}{200}\le\alpha$. Finally, we are ready to analyze the computation and sample complexities of our algorithm.

\begin{theorem}
Let $\alpha, \delta, \gamma>0$. Algorithm \ref{alg:learning} given a $\gamma$-observable POSG of state controlled by one controller with asymmetric delay sharing outputs an $\alpha$-NE \xynew{if the POSG is zero-sum or cooperative, or $\alpha$-CE/CCE if the POSG is general-sum}, with probability at least $1-\delta$, with time and sample complexities bounded by $(OA)^{C(\gamma^{-4}\log\frac{SHO}{\gamma\alpha} + d)}\log\frac{1}{\delta}$ for some universal constant $C>0$. 
\end{theorem}
\begin{proof}{}
    Recall that $\hat{C}_h\le (AO)^{L}$, $P_h\le (AO)^d$, $N_0= \max\left\{\frac{C(\max_h P_h + \log\frac{4H \max_h\hat{C}_h}{\delta_1})}{\zeta_1 \theta_1^2}, \frac{CA(O + \log\frac{4H\max_h(\hat{C}_hP_h)A}{\delta_1})}{\zeta_2\theta_2^2}\right\}$, $N_1 = (OA)^{\hat{L}} \log (\frac{1}{\delta_2})$, and $N_2=C \frac{H^2\log\frac{K^2n}{\delta_3}}{\epsilon^2}$ for some constant $C> 0$\xynew{, and we have set $\delta_1=\delta_2=\delta_3=\frac{\delta}{3}$.} The total number of samples used is $KN_0 + N_1 + (K + nK^2)N_2$. Substituting the choices of parameters into $N_0$, $N_1$, and $N_2$, we proved the sample complexity. Furthermore, for time complexity analysis, since our algorithm only calls the \texttt{BaSeCAMP} and our planning algorithm polynomial number of times, the time complexity is also bounded by $(OA)^{C(\gamma^{-4}\log\frac{SHO}{\gamma\alpha} + d)}\log\frac{1}{\delta}$. 
\end{proof}

\paragraph{Information sharing with one-directional-one-step delay.} For this case, we have $c_h = \{o_{1, 1:h}, o_{2, 1:h-1}, a_{1:h-1}\}$, $p_{1, h} = \emptyset$, $p_{2, h} = \{o_{2, h}\}$, and $z_{h+1} = \{o_{1, h+1}, o_{2, h}, a_h\}$. Fix $L>0$, we construct the approximate common information as $\hat{c}_h = \{o_{1, h-L+1:h}, o_{2, h-L+1:h-1}, a_{h-L:h-1}\}$. For any $\pi^{1:H}$, where $\pi^h\in \Delta(\Pi^{\text{det}})$ for $h\in [H]$, it is easy to verify that
\[
    \PP_h^{\pi^h,  \cG}(s_h, p_h\given \hat{c}_h)=
    \Tilde{\bb}_{h}^{\pi^h}(o_{1, h-L+1:h}, o_{2, h-L+1:h-1}, a_{h-L:h-1})(s_h)\PP_h(o_{2, h}\given s_h, o_{1, h})
\]
where $\PP_h(o_{2, h}\given s_h, o_{1, h}) = \frac{\OO_{h}(o_{1, h}, o_{2, h}\given s_h)}{\sum_{o_{2, h}^\prime}\OO_h (o_{1, h}, o_{2, h}^\prime\given s_h)}$. Furthermore, it is direct to verify that $\hat{L} = L$. Therefore, we conclude that if $L\ge C\frac{\log (H S O /(\epsilon \gamma))}{\gamma^4}$, {by a union bound of the high probability event $\cE_1$ in Lemma \ref{lemma:estimation}, $\cE_2$ in Corollary \ref{corr:expl}, and $\cE_3$ in Lemma \ref{lemma:selection}}, with probability at least $1-\delta_1-\delta_2-\delta_3$, it holds that for any $i\in [n]$:
\$
&\epsilon_{r}(\pi^{1:H, j^\star}) \\
&= \max_{i, h}\max_{\pi\in\Pi^{\text{det}}, {\gamma}_h}\EE_{\pi}^{ \cG}\left|\EE^{ \cG}[r_{i, h}(s_h, a_h)\mid c_h, {\gamma}_h]-\hat{r}^{\Tilde{\cM}}_{i, h}(\hat{c}_h, {\gamma}_h)\right|\\
&\le \max_{h}\max_{\pi\in\Pi^{\text{det}}}\EE_{\pi}^{{\cG}}\left\|  \bb_h(a_{1:h-1}, o_{1:h-1}, o_{1, h})-\Tilde{\bb}_h^{\pi^{h, j^\star}}(a_{h-L:h-1}, o_{h-L+1:h-1}, o_{1, h})\right\|_1\\
&\le \epsilon + \max_{h}\max_{\pi\in\Pi^{\text{det}}}\bm{1}[h>L] \cdot 6 \cdot d_{\cS, h-L}^{\pi, \mathcal{G}}\left(\mathcal{U}_{\phi, h-L}^{\mathcal{G}}\left(\pi^{h, j^\star}\right)\right).
\$
Moreover, we have
\$
&\epsilon_z(\pi^{1:H, j^\star}) \\
&= \max_h\max_{\pi\in\Pi^{\text{det}}, {\gamma}_h} \EE_{\pi}^{ \cG}\left\| \PP_{h}^{ \cG}(\cdot\given c_h, {\gamma}_h) - \PP_{h}^{\Tilde{\cM}, z}(\cdot\given c_h, {\gamma}_h) \right\|_1\\
&\le \max_h\max_{\pi\in\Pi^{\text{det}}, {\gamma}_h}\EE_{ \pi^\prime}^{{ \cG}}\left\|  \bb_h(a_{1:h-1}, o_{1:h-1}, o_{1, h})-\Tilde{\bb}_h^{\pi^{h, j^\star}}(a_{h-L:h-1}, o_{h-L+1:h-1}, o_{1, h})\right\|_1\\
&\le \epsilon + \max_{h}\max_{\pi\in\Pi^{\text{det}}}\bm{1}[h>L] \cdot 6 \cdot d_{\cS, h-L}^{\pi, \mathcal{G}}\left(\mathcal{U}_{\phi, h-L}^{\mathcal{G}}\left(\pi^{h, j^\star}\right)\right).
\$
According to the choice of  $\pi^{1:H, j^\star}$ and Corollary \ref{corr:expl}, it holds that 
\[\max_h\max_{\pi\in\Pi^{\text{det}}}\bm{1}[h>L] \cdot 6 \cdot d_{\cS, h-L}^{\pi, \mathcal{G}}\left(\mathcal{U}_{\phi, h-L}^{\mathcal{G}}\left(\pi^{h, j^\star}\right)\right)\le 6\epsilon.
\]
Therefore, for any $\alpha, \delta>0$, setting 
$\epsilon=\frac{\alpha}{200(H+1)^2}$, $\theta_1 = \frac{\alpha}{200(H+1)^2O}$, $\zeta_2=\zeta_1^2$, $\theta_2 = \frac{\alpha}{200(H+1)^2A\max_h P_h}$, $\zeta_1 = \min\left\{\frac{\alpha\phi}{200(H+1)^2A^{2L}O^L}, \frac{\alpha}{400(H+1)^2A\max_h P_h}\right\}$, $\phi = \frac{\epsilon \gamma^2}{C^2 H^{8} S^5 O^4}$, $\epsilon_e = \frac{\alpha}{200H}$, $\delta_1 = \delta_2 = \delta_3 = \frac{\delta}{3}$, $\Tilde{\cM}(\pi^{1:H, j^\star})$ is an $(\epsilon_r, \epsilon_z)$-expected-approximate common information model of $ \cG$, where $\epsilon_r, \epsilon_z \le \frac{14\alpha}{200(H+1)^2}$. This leads to that $\pi^{\star, j^\star}$ is a $\frac{15\alpha}{200}$-NE{/CE/CCE}, and $|V_{i, 1}^{\pi, \cG }(\emptyset) - V_{i, 1}^{\pi, \hat{\cM}(\pi^{1:H, j^\star})   }(\emptyset)|\le \frac{15\alpha}{200}$ for any policy $\pi\in\Pi$ by Lemma \ref{lemma:v_diff}. By Lemma  \ref{lemma:selection}, $\operatorname{NE/CE/CCE-gap}(\pi^{\star, \hat{j}})\le \operatorname{NE/CE/CCE-gap}(\pi^{\star, j^\star}) + \frac{91\alpha}{200}\le\alpha$. Finally, we are ready to analyze the computation and sample complexities of our algorithm.

\begin{theorem}
Let $\alpha, \delta, \gamma>0$. Algorithm \ref{alg:learning} given a $\gamma$-observable POSG of one-directional-one-step delayed information sharing structure outputs an $\alpha$-NE \xynew{if the POSG is zero-sum or cooperative, or $\alpha$-CE/CCE if the POSG is general-sum}, with probability at least $1-\delta$, with time and sample complexities bounded by $(AO)^{C\gamma^{-4}\log\frac{SHO}{\gamma\alpha}}\log\frac{1}{\delta}$ for some universal constant $C>0$.
\end{theorem}
\begin{proof}
    Recall that $\hat{C}_h\le (OA)^{L}$, $P_h\le O$, $N_0= \max \left\{\frac{C(\max_h P_h + \log\frac{4H \max_h\hat{C}_h}{\delta_1})}{\zeta_1 \theta_1^2}, \frac{CA(O + \log\frac{4H\max_h(\hat{C}_hP_h A)}{\delta_1})}{\zeta_2\theta_2^2}\right\}$, $N_1 = (O A)^{L} \log (\frac{1}{\delta_2})$, and $N_2=C \frac{H^2\log\frac{K^2n}{\delta_3}}{\epsilon^2}$ for some constant $C> 0$\xynew{, and we have set $\delta_1=\delta_2=\delta_3=\frac{\delta}{3}$.} The total number of samples used is $KN_0 + N_1 + (K + nK^2)N_2$. Substituting the choices of parameters into $N_0$, $N_1$, and $N_2$, we proved the sample complexity. Furthermore, for time complexity analysis, since our algorithm only calls the \texttt{BaSeCAMP} and our planning algorithm polynomial number of times, the time complexity is also bounded by $(OA)^{C\gamma^{-4}\log\frac{SHO}{\gamma\alpha}}\log\frac{1}{\delta}$. 
\end{proof}

\paragraph{Uncontrolled state process with delayed sharing.}
The information structure gives that  $c_{h} = \{o_{1:h-d}\}$, $p_{i, h} = \{o_{i, h-d+1:h}\}$, and $z_{h+1} = \{o_{h-d+1}\}$. Fix a $L>0$, the approximate common information is $\hat{c}_h = \{o_{h-d-L+1:h-d}\}$. For any policy $\pi^{1:H}$, where $\pi^{h}\in \Delta(\Pi^{\text{det}})$ for $h\in [H]$, it is easy to verify that
\$
\PP_h^{\Tilde{\cM}(\pi^{1:H}), c}(s_h, p_h\given \hat{c}_h)&=\PP_h^{\pi^h,  \cG}(s_h, p_h\given \hat{c}_h)  = \sum_{s_{h-d}}\Tilde{\bb}^{\pi^{h}}_{h-d}(o_{h-d-L+1:h-d})(s_{h-d})\PP(s_{h}, o_{h-d+1:h}\given s_{h-d}).
\$
Furthermore, it is direct to verify that $\hat{L} = L+d$ by Definition \ref{def:L_main}. Therefore, we conclude that if $L\ge C\frac{\log (H S O /(\epsilon \gamma))}{\gamma^4}$, {by a union bound of the high probability event $\cE_1$ in Lemma \ref{lemma:estimation}, $\cE_2$ in Corollary \ref{corr:expl}, and $\cE_3$ in Lemma \ref{lemma:selection}}, with probability at least $1-\delta_1-\delta_2-\delta_3$, it holds that for any $i\in [n]$:
\$
\epsilon_{r}(\pi^{1:H, j^\star}) 
&= \max_{i, h}\max_{\pi\in\Pi^{\text{det}}, {\gamma}_h}\EE_{a_{1:h-1}, o_{1:h}\sim \pi}^{ \cG}\left|\EE^{ \cG}[r_{i, h}(s_h, a_h)\mid c_h, {\gamma}_h]-\hat{r}^{\Tilde{\cM}}_{i, h}(\hat{c}_h, {\gamma}_h)\right|\\
&\le  \max_{h}\max_{\pi\in\Pi^{\text{det}}}\EE_{a_{1:h-1}, o_{1:h}\sim \pi}^{{ \cG}}\left\|\bb_{h-d}( o_{1:h-d})-\Tilde{\bb}_{h-d}^{\pi^{h, j^\star}}(o_{h-d-L+1:h-d})\right\|_1\\
&\le \epsilon + \max_{ h}\max_{\pi\in\Pi^{\text{det}}}\bm{1}[h>\hat{L}] \cdot 6 \cdot d_{\cS, h-\hat{L}}^{\pi, \mathcal{G}}\left(\mathcal{U}_{\phi, h-\hat{L}}^{\mathcal{G}}\left(\pi^{h, j^\star}\right)\right).
\$
Moreover, we also have
\$
\epsilon_z(\pi^{1:H, j^\star}) &= \max_h\max_{\pi\in\Pi^{\text{det}}, {\gamma}_h} \EE_{a_{1:h-1}, o_{1:h}\sim \pi}^{ \cG}\left\| \PP_{h}^{ \cG}(\cdot\given c_h, {\gamma}_h) - \PP_{h}^{\Tilde{\cM}, z}(\cdot\given c_h, {\gamma}_h) \right\|_1\\
&\le \max_h\max_{\pi\in\Pi^{\text{det}}, {\gamma}_h}\EE_{a_{1:h-1}, o_{1:h}\sim \pi^\prime}^{{ \cG}}\left\|  \bb_{h-d}( o_{1:h-d})-\Tilde{\bb}_{h-d}^{\pi^{h, j^\star}}(o_{h-d-L+1:h-d})\right\|_1\\
&\le \epsilon + \max_{h}\max_{\pi\in\Pi^{\text{det}}}\bm{1}[h>\hat{L}] \cdot 6 \cdot d_{\cS, h-\hat{L}}^{\pi, \mathcal{G}}\left(\mathcal{U}_{\phi, h-\hat{L}}^{\mathcal{G}}\left(\pi^{h, j^\star}\right)\right).
\$
According to the choice of  $\pi^{1:H, j^\star}$ and Corollary \ref{corr:expl}, it holds that 
\[\max_h\max_{\pi\in\Pi^{\text{det}}}\bm{1}[h>\hat{L}] \cdot 6 \cdot d_{\cS, h-\hat{L}}^{\pi, \mathcal{G}}\left(\mathcal{U}_{\phi, h-\hat{L}}^{\mathcal{G}}\left(\pi^{h, j^\star}\right)\right)\le 6\epsilon.
\]

Therefore, for any $\alpha, \delta>0$,   setting 
$\epsilon=\frac{\alpha}{200(H+1)^2}$, $\theta_1 = \frac{\alpha}{200(H+1)^2O}$, $\zeta_2=\zeta_1^2$, $\theta_2 = \frac{\alpha}{200(H+1)^2A\max_h P_h}$, $\zeta_1 = \min\left\{\frac{\alpha\phi}{200(H+1)^2A^{2(L+d)}O^{L+d}}, \frac{\alpha}{400(H+1)^2A\max_h P_h}\right\}$, $\phi = \frac{\epsilon \gamma^2}{C^2 H^{8} S^5 O^4}$, $\epsilon_e = \frac{\alpha}{200H}$, $\delta_1 = \delta_2 = \delta_3 = \frac{\delta}{3}$, $\Tilde{\cM}(\pi^{1:H, j^\star})$ is an $(\epsilon_r, \epsilon_z)$-expected-approximate common information model of $ \cG$, where $\epsilon_r, \epsilon_z \le \frac{14\alpha}{200(H+1)^2}$. This leads to that $\pi^{\star, j^\star}$ is a $\frac{15\alpha}{200}$-NE{/CE/CCE}, and $|V_{i, 1}^{\pi,  \cG }(\emptyset) - V_{i, 1}^{\pi, \hat{\cM}(\pi^{1:H, j^\star})   }(\emptyset)|\le \frac{15\alpha}{200}$ for any policy $\pi$ by Lemma \ref{lemma:v_diff}. By Lemma \ref{lemma:selection}, \$\operatorname{NE/CE/CCE-gap}(\pi^{\star, \hat{j}})\le \operatorname{NE/CE/CCE-gap}(\pi^{\star, j^\star}) + \frac{91\alpha}{200}\le\alpha.\$ Finally, we are ready to analyze the computational and sample complexities of our algorithm. 

\begin{theorem}
Let $\alpha, \delta, \gamma>0$. Algorithm \ref{alg:learning} given a $\gamma$-observable POSG of uncontrolled state process and delayed information sharing structure outputs an $\alpha$-NE \xynew{if the POSG is zero-sum or cooperative, or $\alpha$-CE/CCE if the POSG is general-sum}, with probability at least $1-\delta$, with time and sample complexities bounded by $(OA)^{C(\gamma^{-4}\log\frac{SHO}{\gamma\alpha} + d)}\log\frac{1}{\delta}$ for some universal constant $C>0$. 
\end{theorem}
\begin{proof}
    Recall that $\hat{C}_h\le O^{L}$, $P_h\le O^d$, $N_0= \max \left\{\frac{C(\max_h P_h + \log\frac{4H \max_h\hat{C}_h}{\delta_1})}{\zeta_1 \theta_1^2}, \frac{CA(O + \log\frac{4H\max_h(\hat{C}_hP_h)A}{\delta_1})}{\zeta_2\theta_2^2}\right\}$, $N_1 = (OA)^{\hat{L}} \log (\frac{1}{\delta_2})$, and $N_2=C \frac{H^2\log\frac{K^2n}{\delta_3}}{\epsilon^2}$ for some constant $C> 0$\xynew{, and we have set $\delta_1=\delta_2=\delta_3=\frac{\delta}{3}$.} The total number of samples used is $KN_0 + N_1 + (K + nK^2)N_2$. Substituting the choices of parameters into $N_0$, $N_1$, and $N_2$, we proved the sample complexity. Furthermore, for time complexity analysis, since our algorithm only calls the \texttt{BaSeCAMP} and our planning algorithm polynomial number of times, the time complexity is also bounded by $(OA)^{C(\gamma^{-4}\log\frac{SHO}{\gamma\alpha} )}\log\frac{1}{\delta}$.
\end{proof}

\paragraph{Symmetric information game.}
For symmetric information game, $c_{h} = \{o_{1:h}, a_{1:h-1}\}$, $p_{i, h} = \emptyset$, and $z_{h+1} = \{a_{h}, o_{h+1}\}$. Fix $L> 0$, we construct the approximate common information as $\hat{c}_{h} = \{o_{h-L+1:h}, a_{h-L:h-1}\}$. For any $\pi^{1:H}$, where $\pi^h\in\Delta(\Pi^{\text{det}})$ for $h\in[H]$, it is easy to verify that
\[
\PP_h^{\Tilde{\cM}(\pi^{1:H}), c}(s_h, p_h\given \hat{c}_h)=\PP_h^{\pi^h,  \cG}(s_h, p_h\given \hat{c}_h) =\Tilde{\bb}^{\pi^h}_{h}(a_{h-L:h-1}, o_{h-L+1:h})(s_h).
\]
Meanwhile, it is direct to verify that $\hat{L} = L$ by Definition \ref{def:L_main}. Therefore, we conclude that if $L\ge C\frac{\log (H S O /(\epsilon \gamma))}{\gamma^4}$, {by a union bound of the high probability event $\cE_1$ in Lemma \ref{lemma:estimation}, $\cE_2$ in Corollary \ref{corr:expl}, and $\cE_3$ in Lemma \ref{lemma:selection}}, with probability at least $1-\delta_1-\delta_2-\delta_3$, it holds that for any $i\in[n]$:
\$
&\epsilon_{r}(\pi^{1:H, j^\star}) 
= \max_{i, h}\max_{\pi\in\Pi^{\text{det}}, {\gamma}_h}\EE_{a_{1:h-1}, o_{1:h}\sim \pi}^{ \cG}\left|\EE^{ \cG}[r_{i, h}(s_h, a_h)\mid c_h, {\gamma}_h]-\hat{r}^{\Tilde{\cM}}_{i, h}(\hat{c}_h, {\gamma}_h)\right|\\
&\quad\le \max_{h}\max_{\pi\in\Pi^{\text{det}}}\EE_{a_{1:h-1}, o_{1:h}\sim \pi}^{{ \cG}}\left\|  \bb_h(a_{1:h-1}, o_{1:h})-\Tilde{\bb}_h^{\pi^{h, j^\star}}(a_{h-L:h-1}, o_{h-L+1:h})\right\|_1\\
&\quad\le \epsilon + \max_{h}\max_{\pi\in\Pi^{\text{det}}}\bm{1}[h>L] \cdot 6 \cdot d_{\cS, h-L}^{\pi, \mathcal{G}}\left(\mathcal{U}_{\phi, h-L}^{\mathcal{G}}\left(\pi^{h, j^\star}\right)\right).
\$
Moreover, we have 
\$
&\epsilon_z(\pi^{1:H, j^\star}) = \max_h\max_{\pi\in\Pi^{\text{det}}, {\gamma}_h} \EE_{a_{1:h-1}, o_{1:h}\sim \pi}^{ \cG}\left\| \PP_{h}^{ \cG}(\cdot\given c_h, {\gamma}_h) - \PP_{h}^{\Tilde{\cM}, z}(\cdot\given c_h, {\gamma}_h) \right\|_1\\
&\quad\le \max_h\max_{\pi\in\Pi^{\text{det}}, {\gamma}_h}\EE_{a_{1:h-1}, o_{1:h}\sim \pi^\prime}^{{ \cG}}\left\|  \bb_h(a_{1:h-1}, o_{1:h})-\Tilde{\bb}_h^{\pi^{h, j^\star}}(a_{h-L:h-1}, o_{h-L+1:h})\right\|_1\\
&\quad\le \epsilon + \max_{h}\max_{\pi\in\Pi^{\text{det}}}\bm{1}[h>L] \cdot 6 \cdot d_{\cS, h-L}^{\pi, \mathcal{G}}\left(\mathcal{U}_{\phi, h-L}^{\mathcal{G}}\left(\pi^{h, j^\star}\right)\right).
\$
According to the choice of  $\pi^{1:H, j^\star}$ and Corollary \ref{corr:expl}, it holds that 
\[\max_h\max_{\pi\in\Pi^{\text{det}}}\bm{1}[h>L] \cdot 6 \cdot d_{\cS, h-L}^{\pi, \mathcal{G}}\left(\mathcal{U}_{\phi, h-L}^{\mathcal{G}}\left(\pi^{h, j^\star}\right)\right)\le 6\epsilon.
\]
Therefore, for any $\alpha, \delta>0$, setting 
$\epsilon=\frac{\alpha}{200(H+1)^2}$, $\theta_1 = \frac{\alpha}{200(H+1)^2O}$, $\zeta_2=\zeta_1^2$, $\theta_2 = \frac{\alpha}{200(H+1)^2A\max_h P_h}$, $\zeta_1 = \min\left\{\frac{\alpha\phi}{200(H+1)^2A^{2L}O^L}, \frac{\alpha}{400(H+1)^2A\max_h P_h}\right\}$,  $\phi = \frac{\epsilon \gamma^2}{C^2 H^{8} S^5 O^4}$, $\epsilon_e = \frac{\alpha}{200H}$, $\delta_1 = \delta_2 = \delta_3 = \frac{\delta}{3}$, $\Tilde{\cM}(\pi^{1:H, j^\star})$ is an $(\epsilon_r, \epsilon_z)$-expected-approximate common information model of $ \cG$, where $\epsilon_r, \epsilon_z \le \frac{14\alpha}{200(H+1)^2}$. This leads to that $\pi^{\star, j^\star}$ is a $\frac{15\alpha}{200}$-NE{/CE/CCE}, and $|V_{i, 1}^{\pi,  \cG }(\emptyset) - V_{i, 1}^{\pi, \hat{\cM}(\pi^{1:H, j^\star})   }(\emptyset)|\le \frac{15\alpha}{200}$ for any policy $\pi\in\Pi$ by Lemma \ref{lemma:v_diff}. By Lemma \ref{lemma:selection}, $\operatorname{NE/CE/CCE-gap}(\pi^{\star, \hat{j}})\le \operatorname{NE/CE/CCE-gap}(\pi^{\star, j^\star}) + \frac{91\alpha}{200}\le\alpha$. Finally, we are ready to analyze the computation and sample complexities of our algorithm.

\begin{theorem}
Let $\alpha, \delta, \gamma>0$. Algorithm \ref{alg:learning} given a $\gamma$-observable POSG of symmetric information sharing structure outputs an $\alpha$-NE \xynew{if the POSG is zero-sum or cooperative, or $\alpha$-CE/CCE if the POSG is general-sum}, with probability at least $1-\delta$, with time and sample complexities bounded by $(AO)^{C\gamma^{-4}\log\frac{SHO}{\gamma\alpha}}\log\frac{1}{\delta}$ for some universal constant $C>0$. 
\end{theorem}
\begin{proof}
    Recall that $\hat{C}_h\le (OA)^{L}$, $P_h=1$, $N_0= \max \left\{\frac{C(\max_h P_h + \log\frac{4H \max_h\hat{C}_h}{\delta_1})}{\zeta_1 \theta_1^2}, \frac{CA(O + \log\frac{4H\max_h(\hat{C}_hP_h)A}{\delta_1})}{\zeta_2\theta_2^2}\right\}$, $N_1 = (O A)^{L} \log (\frac{1}{\delta_2})$, and $N_2=C \frac{H^2\log\frac{K^2n}{\delta_3}}{\epsilon^2}$ for some constant $C> 0$\xynew{, and we have set $\delta_1=\delta_2=\delta_3=\frac{\delta}{3}$.} The total number of samples used is $KN_0 + N_1 + (K + nK^2)N_2$. Substituting the choices of parameters into $N_0$, $N_1$, and $N_2$, we proved the sample complexity. Furthermore, for time complexity analysis, since our algorithm only calls the \texttt{BaSeCAMP} and our planning algorithm polynomial number of times, the time complexity is also bounded by $(OA)^{C(\gamma^{-4}\log\frac{SHO}{\gamma\alpha} )}\log\frac{1}{\delta}$.
\end{proof}

\subsection{Missing details in \Cref{dec-pomdp}}\label{supp-sec-5} 
Now we prove \Cref{prop:static-hard}, where the hardness follows from the hardness of the one-step Dec-POMDP in \Cref{one-step}.
\begin{prevproof}{prop:static-hard}
Note that for \Cref{eq:team-Q}, if we take the underlying Dec-POMDP $\cG$ to be $H=1$, $n=2$ without any information-sharing, and the approximate belief is constructed to be the ground-truth belief of the underlying Dec-POMDP $\cG$, the optimal prescription solved by \Cref{eq:team-Q} is then exactly the optimal policy of the underlying $\cG$. By the hardness from \Cref{one-step}, we conclude that solving \Cref{eq:team-Q}  
is also \texttt{NP-hard}.
\end{prevproof}
\begin{proposition}\label{lem:single-controller}
	Given any approximate common information model $\cM$ that is consistent with a belief {\\$\{\PP_{h^\prime}^{\cM, c}(s_{h^\prime}, p_{h^\prime}\given \hat{c}_{h^\prime})\}_{h^\prime\in [H]}$}, if \textbf{Condition 1} holds, we have for any $h\in[H]$, $\hat{c}_h\in\hat{\cC}_h$, $\gamma_{h}\in\Gamma_{h}$ 
	\#\label{eq:cond1-q}
	Q_h^{\star, \cM}(\hat{c}_h, \gamma_{h})=\sum_{j\in[n]}U_{j, h}(\hat{c}_h, \gamma_{j, h}),
	\#
	for some functions $\{U_{j, h}\}_{j\in[n]}$. Correspondingly, \Cref{eq:team-Q} can be solved exactly in time complexity $\texttt{poly}(S, A, P_h)$.
\end{proposition}

\begin{prevproof}{lem:single-controller}
	By the definition of $Q_h^{\star, \cM}(\hat{c}_h, \gamma_{h})$ and \Cref{def:consistency}, it holds that
	\$
	&Q_h^{\star, \cM}(\hat{c}_h, \gamma_{h})\\
	&=\sum_{s_h, p_h, a_h, s_{h+1}, o_{h+1}}\PP_{h}^{\cM, c}(s_h, p_h\given \hat{c}_h)\prod_{i=1}^n\gamma_{i, h}(a_{i, h}\given p_{i, h})\TT_h(s_{h+1}\given s_h, a_{\cth, h})\\
	&\qquad\OO_{h+1}(o_{h+1}\given s_{h+1})\left[r_{h}(s_h, a_{h}) + V_{h+1}^{\star,\cM}(\hat{c}_{h+1})\right]\\
	&=\sum_{j\not=\cth}\sum_{s_h, p_h, a_{j, h}}\PP_{h}^{\cM, c}(s_h, p_h\given \hat{c}_h)\gamma_{j, h}(a_{j, h}\given p_{j, h})\left [r_{j, h}(s_h, a_{j, h})\right]+  \\
	& \sum_{s_h, p_h, a_{\cth, h}, s_{h+1}, o_{h+1}}\PP_{h}^{\cM, c}(s_h, p_h\given \hat{c}_h)\gamma_{\cth, h}(a_{\cth, h}\given p_{\cth, h})\TT_h(s_{h+1}\given s_h, a_{\cth,h})\OO_{h+1}(o_{h+1}\given s_{h+1})\\
	&\qquad\qquad\qquad\qquad\times\left[ r_{\cth, h}(s_h, a_{\cth, h})+ V_{h+1}^{\star,\cM}(\hat{c}_{h+1})\right]\\
	&:= \sum_{j\in[n]}U_{j, h}(\hat{c}_h, \gamma_{j, h})
	\$
	where the last step is due to the assumption that that $\hat{c}_{h+1}=\hat{\phi}_{h+1}(\hat{c}_h, z_{h+1})$ and $z_{h+1}=\chi_{h+1}(p_h, a_{\cth, h}, o_{h+1})$. Now, to solve \Cref{eq:team-Q}, we only need to optimize w.r.t. each $\gamma_{j, h}$ for $j\in[n]$ \emph{individually}, which is a linear program with the constraint set of $\gamma_{j, h}$ to be a concatenation of simplex by \Cref{prop:linear}. Hence, \Cref{eq:team-Q} can be solved even \emph{exactly} in time complexity $\texttt{poly}(S, A, P_h)$.
\end{prevproof}
\begin{proposition}\label{nested}
	Suppose \textbf{Condition 2} holds, \Cref{alg:agent-dp} returns $\gamma_{1:n, h}^\star$ such that
	\[
		\gamma_{1:n, h}^\star\in\arg\max_{\gamma_{1, h}, \cdots, \gamma_{n, h}} Q_{h}^{\star, \cM}(\hat{c}_h, \gamma_{1, h}, \cdots, \gamma_{n, h}),
	\]
	with time complexity $\texttt{poly}(P_h, A, S)$.
\end{proposition}

\begin{prevproof}{nested}
 We slightly abuse our notation for the $Q^{\star, \cM}_h$ as below to define for any $u_{i}\in\cU_i:=\{(\times_{j=1}^i\cP_{j, h})\times(\times_{j=1}^{i-1}\cA_{j})\rightarrow \Delta(\cA_i)\}$ and $i\in[n]$ that
\$
&Q_{h}^{\star, \cM}(\hat{c}_h, u_{1}, \cdots, u_{n})\\
	&:=\sum_{s_h, p_h, a_h, s_{h+1}, o_{h+1}}\PP_{h}^{\cM, c}(s_h, p_h\given \hat{c}_h)\prod_{i=1}^n u_{i}(a_{i, h}\given p_{1:i, h}, a_{1:i-1, h})\TT_h(s_{h+1}\given s_h, a_{h}) \OO_{h+1}(o_{h+1}\given s_{h+1})\left[r_{h}(s_h, a_{h}) + V_{h+1}^{\star,\cM}(\hat{c}_{h+1})\right].
\$
By the standard result of value iteration for POMDPs, we have that $u^\star_{1:n}$ is an optimal policy for the POMDP $\hat{\cP}(n)$ in the sense that
\$
Q_{h}^{\star, \cM}(\hat{c}_h, u_{1}^\star, \cdots, u_{n}^\star)&=\max_{\{u_i\in\cU_i\}_{i\in[n]}}Q_{h}^{\star, \cM}(\hat{c}_h, u_{1}, \cdots, u_{n}) \ge \max_{\{\gamma_{i, h}\in\Gamma_{i, h}\}_{i\in[n]}}Q_{h}^{\star, \cM}(\hat{c}_h, \gamma_{i, h}, \cdots, \gamma_{n, h}),
\$
where the inequality comes from the fact that any $\gamma_{i, h}\in\Gamma_{i, h}$ can be realized by an equivalent $u_i\in\cU_i$ such that the value is the same. Meanwhile, due to the nested information-sharing structure, for any $p_h\in\cP_h$, it holds that $u_{1:n}^\star$ and $\gamma^\star_{1:n, h}$ outputs the same action deterministically according to the second for-loop of \Cref{alg:agent-dp}. Hence, we conclude that
\[
Q_{h}^{\star, \cM}(\hat{c}_h, \gamma_{1, h}^\star, \cdots, \gamma_{n, h}^\star)=Q_{h}^{\star, \cM}(\hat{c}_h, u_{1}^\star, \cdots, u_{n}^\star),
\]
which further concludes that $\gamma_{1:n, h}^\star$ returned by \Cref{alg:agent-dp} is an exact solution of \Cref{eq:team-Q}. Finally, the time complexity scales with the size of the history space of $\hat{\cP}(n)$, which is $\prod_{i=1}^n P_{i, h}A_{i, h}=AP_h$. The additional polynomial dependency on $S$ comes from computing the posterior distribution for the initialization step in \Cref{alg:agent-dp}. 
\end{prevproof}
\begin{proposition}\label{factor}
	Suppose \textbf{Condition 3} holds. For each $h\in[H]$, there exist $n$ functions $\{F_{i, h}\}_{i\in[n]}$ such that
	\#\label{eq:cond3-q}
	Q_{h}^{\star,\cM}(\hat{c}_h, \gamma_{1, h}, \cdots, \gamma_{n, h})=\sum_{i=1}^n F_{i, h}(\hat{c}_{i, h}, \gamma_{i, h}).
	\#
	Correspondingly, \Cref{eq:team-Q} can be solved in time $\sum_{i\in[n]}\texttt{poly}(S_i, A_i, P_{i, h})$.
\end{proposition}

\begin{prevproof}{factor}
	We prove our result by backward induction on $h$. Obviously, it holds for $h=H+1$. Now suppose the proposition holds for $h+1$. For step $h$, it holds that 
	\$
	&Q_h^{\star, \cM}(\hat{c}_h, \gamma_{1, h}, \cdots, \gamma_{n, h})\\
	&=\sum_{s_h, p_h, a_h, s_{h+1}, o_{h+1}}\PP_{h}^{\cM, c}(s_h, p_h\given \hat{c}_h)\prod_{j=1}^n\gamma_{j, h}(a_{j, h}\given p_{j, h})\TT_h(s_{h+1}\given s_h, a_{h})\\
	&\qquad\qquad\OO_{h+1}(o_{h+1}\given s_{h+1})\left[\sum_{i=1}^n r_{i, h}(s_{i, h}, a_{i, h})+F_{i, h+1}(\hat{c}_{i, h+1}, \gamma_{i, h+1}^\star(\hat{c}_{i, h+1}))\right]\\
	&=\sum_{i=1}^n\sum_{s_{i, h}, p_{i, h}, a_{i, h}, s_{i,h+1}, o_{i, h+1}}\PP_{i, h}^{\cM, c}(s_{i, h}, p_{i, h}\given \hat{c}_{i, h})\gamma_{i, h}(a_{i, h}\given p_{i, h})\TT_h(s_{i, h+1}\given s_{i, h}, a_{i, h})\\
	&\qquad\qquad\OO_{i, h+1}(o_{i, h+1}\given s_{i, h+1})\left[r_{i, h}(s_{i, h}, a_{i, h})+F_{i, h+1}(\hat{c}_{i, h+1}, \gamma_{i, h+1}^\star(\hat{c}_{i, h+1}))\right]\\
	&:=\sum_{i=1}^{n}F_{i, h}(\hat{c}_{i, h}, \gamma_{i, h}),
	\$
	where for the first equality, we defined $\gamma_{i, h+1}^\star(\hat{c}_{i, h+1})\in\arg\max_{\gamma_{i, h+1}\in\Gamma_{i, h+1}}F_{i, h+1}(\hat{c}_{i, h+1}, \gamma_{i, h+1})$, thus proving the decomposition. Therefore, to solve \Cref{eq:team-Q}, it suffices to optimize each $F_{i, h}(\hat{c}_{i, h},\gamma_{i, h})$ individually w.r.t. $\gamma_{i, h}$, which is a linear program with the concatenation of simplex as the constraint by \Cref{prop:linear}. Thus, the time complexity is $\sum_{i=1}^n\texttt{poly}(S_i, A_i, P_{i, h})$.
\end{prevproof}
\begin{remark}\label{remark:factor}
In fact, under \textbf{Condition 3}, \Cref{alg:avi} and its time complexity can be further improved, where for each $h\in[H]$, we do not necessarily need to enumerate all possible joint approximate common information $\hat{c}_h$, but only the individual approximate common information $\hat{c}_{i, h}$ for each $i\in[n]$. This allows the final time complexity to depend only on $\max_{h\in[H]}\sum_{i\in[n]}\hat{C}_{i, h}P_{i, h}$ instead of $\max_{h\in[H]}\hat{C}_{h}P_{h}$, thus \emph{not} suffering from the exponential dependency on the number of agents anymore.
\end{remark}

\begin{prevproof}{thm:plan_cases_dec}
	The first step is to show that $\hat{\pi}^\star$, i.e., the return of \Cref{alg:avi} with the equilibrium-computation  subroutine replaced as \Cref{eq:team-Q} is a near-optimal policy for the underlying Dec-POMDP $\cG$. To begin with, for any policy $\pi\in\Pi$, we shall prove inductively that for any $h\in[H]$, $c_h\in\cC_h$ that
	\[
	V_{h}^{\pi, \cM}(c_h)\le V_{h}^{\hat{\pi}^\star, \cM}(\hat{c}_h).
	\]
	It is direct to verify that the inequality holds for $h=H+1$. Now suppose it holds for step $h+1$. For step $h$, note that
	\$
	V_{h}^{\pi, \cM}(c_h)&=\EE_{\{\omega_{j, h}\}_{j\in[n]}}\EE^{\cM}[\hat{r}_{h}^\cM + V_{h+1}^{\pi, \cM}(c_{h+1})\given \hat{c}_h, \{\pi_{j, h}(\cdot\given \omega_{j, h}, c_h, \cdot)\}_{j\in[n]}]\\
	&\le \EE_{\{\omega_{j, h}\}_{j\in[n]}}\EE^{\cM}[\hat{r}_{h}^\cM + V_{h+1}^{\hat{\pi}^\star,\cM}(\hat{c}_{h+1})\given \hat{c}_h, \{\pi_{j, h}(\cdot\given \omega_{j, h}, c_h, \cdot)\}_{j\in[n]}]\\
	&\le \EE^{\cM}[\hat{r}_{h}^\cM + V_{h+1}^{\hat{\pi}^\star,\cM}(\hat{c}_{h+1})\given \hat{c}_h, \{\hat{\pi}^{\star}_{j, h}(\cdot\given \hat{c}_h, \cdot)\}_{j\in[n]}]\\
	&=V^{\pi^\star, \cM}_h(\hat{c}_h),
	\$
	where the first inequality is by inductive hypothesis, and the second inequality is due to  $V_{h+1}^{\hat{\pi}^\star,\cM}(\hat{c}_{h+1})=V_{h+1}^{\star,\cM}(\hat{c}_{h+1})$ and $\{\hat{\pi}^{\star}_{j, h}(\cdot\given \hat{c}_h, \cdot)\}_{j\in[n]}$ is a solution of \Cref{eq:team-Q}. Now under the ground-truth model $\cG$, for any $\pi\in\Pi$, $h\in[H]$, $c_h\in\cC_h$, by \Cref{lemma:v_diff}, it holds that
	\[
	V_1^{\pi, \cG}(\emptyset)-V_1^{\hat{\pi}^\star, \cG}(\emptyset)\le V_1^{\pi, \cM}(\emptyset)-V_1^{\hat{\pi}^\star, \cM}(\emptyset) +2(H\epsilon_r + H^2\epsilon_z) \le 2(H\epsilon_r + H^2\epsilon_z).
	\]
	
	To analyze the time complexity, we observe that \Cref{alg:avi} needs to solve \Cref{eq:team-Q} for $\hat{C}_h$ times for each $h\in[H]$. Therefore, if \Cref{eq:team-Q} can be solved with time complexity $\texttt{poly}(S, A, P_h)$ for each $h\in[H]$, the total time complexity of \Cref{alg:avi} is $H\max_{h\in[H]}\hat{C}_h\times\texttt{poly}(S, A, P_h)$.
	
	Now we are ready to instantiate the guarantees for the examples in \Cref{sec:example}. Specifically, it is direct to verify that \Cref{exam-2} and \Cref{exam-5} together with the approximate belief constructed in \Cref{subsec:finite_memory} satisfy \textbf{Condition 1} (turned-based structures), while \Cref{exam:sym} and \Cref{exam-5} together with the approximate belief constructed in \Cref{subsec:finite_memory} satisfy \textbf{Condition 2} (the nested information-sharing structure). Therefore, by \Cref{lem:single-controller} and \Cref{nested}, \Cref{eq:team-Q} can be solved with time complexity $\texttt{poly}(S, A, P_h)$ for each $h\in[H]$, and the total time complexity of planning such a $2(H\epsilon_r + H^2\epsilon_z)-$team-optimal solution for the Dec-POMDP is $\max_{h\in[H]}\hat{C}_h\cdot \texttt{poly}(S, A, P_h, H)$. Finally, by \Cref{thm:plan_cases_full}, for all examples in \Cref{sec:example}, there exists an approximate model $\cM$ such that $\max\{\epsilon_r, \epsilon_z\}\le \cO(\frac{\epsilon}{H^2})$, while $\max_h\hat{\cC}_hP_h$ is only quasi-polynomial of the problem instance size. Hence, the time complexity for planning the $\epsilon$-team-optimal solution for those examples is also quasi-polynomial.
		
		For \Cref{exam-1}, i.e., the one-step delayed sharing case, if we additionally  assume the \textbf{Part (1)}  of \textbf{Condition 3} (factorized structures) holds, the approximate belief we constructed in  \Cref{subsec:finite_memory} also satisfies the \textbf{Part (2)} of \textbf{Condition 2}. Thus, by the improved algorithm and guarantees in \Cref{remark:factor} and \Cref{factor}, the total time complexity is $n\max_{i\in[n], h\in[H]}\hat{C}_{i, h}\times\texttt{poly}(S_i, A_i, P_{i, h}, H)$. Meanwhile, by our construction of the approximate belief, we can ensure $\max\{\epsilon_r, \epsilon_z\}\le \cO(\frac{\epsilon}{H^2})$, while $\hat{C}_{i, h}P_{i, h}\le (A_i O_i)^{\cO(\log(SH/\epsilon)/\gamma^4)}$. Therefore, the total time complexity of planning the $\epsilon$-team-optimal solution is $n(A_i O_i)^{\cO(\log(SH/\epsilon)/\gamma^4)}$, without suffering from the exponential dependency on $n$.
\end{prevproof}

\paragraph{(Quasi-)Efficient learning in Dec-POMDPs without model knowledge.} Based on such planning algorithms, we are ready to extend our MARL algorithm   to the Dec-POMDP setting {for finding the team optimum}. Specifically, we only need to replace line 4 of \Cref{alg:learning}, i.e., planning for equilibria of the POSG with the planning algorithm for the team-optimal solution of the Dec-POMDP discussed above. Meanwhile, the line 7 of \Cref{alg:learning} for policy selection (\Cref{alg:selection}) can be greatly simplified, where we can directly choose 
\[
\hat{j}\leftarrow \arg\max_{j\in[K]} R^j,
\] 
i.e., the policy with the highest empirical rewards. For completeness, we provided the modified policy selection algorithm in \Cref{alg:selection-dec}.
Meanwhile, \Cref{alg:construct_mg} of learning the approximate model $\hat{\cM}(\pi^{1:H})$ also needs to ensure that planning in the learned $\hat{\cM}(\pi^{1:H})$ is computationally quasi-efficient. Specifically, \Cref{eq:team-Q} needs to be solved computationally efficiently for $\hat{\cM}(\pi^{1:H})$ by enforcing that \textbf{Condition 1}, \textbf{2}, or \textbf{3} holds. This can be done by slightly adjusting \Cref{alg:construct_mg} as in the proof of the following theorem.

\begin{theorem}\label{thm:learning_dec} 
	Fix $\epsilon, \delta>0$. Under Assumption \ref{observa}, for the one-step delayed sharing example under the assumption of \textbf{Part (1)} in \textbf{Condition 3} and all the other information-sharing structure examples in \Cref{sec:example}, there exists a multi-agent RL algorithm that learns an $\epsilon$-team optimal solution with probability at least $1-\delta$,  with \emph{both} quasi-polynomial time and sample complexities. 
\end{theorem}

To prove \Cref{thm:learning_dec}, the major step is to prove the correctness of the simplified policy selection procedure, i.e., the counterpart of \Cref{lemma:selection} for the Dec-POMDP setting.
\begin{lemma}\label{lemma:selection-dec}
	    Fix $\epsilon$, $\delta_3>0$. For Algorithm \ref{alg:selection-dec}, suppose that the $K$ groups of policies $\{\pi^{1:H, j}\}_{j=1}^{K}$ satisfy that there exists some $j^\star\in [K]$ such that for any policy $\pi\in\Pi$, we have 
    \[
    \left|V_{1}^{\pi,  \cG }(\emptyset) - V_{1}^{\pi, \hat{\cM}(\pi^{1:H, j^\star})}(\emptyset)\right|\le \epsilon.
    \]
    If $N_2\ge C\frac{H^2\log\frac{K^2n}{\delta_3}}{\epsilon^2}$ for some constant $C>0$, then with probability at least $1-\delta_3$, it holds that 
    \[
    V^{\pi^{\star, \hat{j}}, \cG}_1(\emptyset)\ge \max_{\pi\in\Pi}V^{\pi, \cG}_1(\emptyset) - 4\epsilon.
    \]
\end{lemma}
\begin{proof}
	By the concentration bound on the accumulated rewards of policies $\pi^{\star, j}$, and further a union bound over all $j\in[n]$, with probability at least $1-\delta_3$, the following event $\cE_3$ holds for any $j\in [K]$:
\[
\left|R^j - V_{1}^{\pi^{\star, j},  \cG }(\emptyset)\right|\le \epsilon.
\]
Therefore, it holds that
\[
V^{\pi^{\star, \hat{j}}, \cG}_1(\emptyset)\ge R^{\hat{j}}-\epsilon\ge R^{j^\star}-\epsilon\ge V^{\pi^{\star, j^\star}, \cG}_1(\emptyset) - 2\epsilon.
\]
Meanwhile, by denoting $\pi^\star\in\arg\max_{\pi\in\Pi}V_1^{\pi, \cG}(\emptyset)$, we have
\[
V^{\pi^{\star, j^\star}, \cG}_1(\emptyset)-V^{\pi^\star, \cG}_1(\emptyset)\ge V^{\pi^{\star, j^\star}, \hat{\cM}(\pi^{1:H, j^\star})}_1(\emptyset)-V^{\pi^\star, \hat{\cM}(\pi^{1:H, j^\star})}_1(\emptyset)-2\epsilon\ge-2\epsilon,
\]
where the last step is due to the fact that $\pi^{\star, j^\star}$ is the optimal policy of $\hat{\cM}(\pi^{1:H, j^\star})$. Therefore, we conclude that that $V^{\pi^{\star, \hat{j}}, \cG}_1(\emptyset)\ge V^{\pi^{\star, j^\star}, \cG}_1(\emptyset) - 2\epsilon\ge \max_{\pi\in\Pi}V_1^{\pi, \cG}(\emptyset)-4\epsilon$.
\end{proof}
Finally, we are ready to prove \Cref{thm:learning_dec}.
\begin{prevproof}{thm:learning_dec}
	The correctness of the extended learning algorithm follows similarly as the proof of \Cref{thm:main_learning_full}, where for any $\alpha, \delta>0$, under the exactly the same choices of all parameters (cf. \Cref{subsec:learning_proof}) as for learning the equilibrium, with probability $1-\delta$, there exists $j^\star\in[K]$ such that $|V_{1}^{\pi, \cG}(\emptyset) - V_{1}^{\pi, \hat{\cM}(\pi^{1:H, j^\star})}(\emptyset)|\le\frac{15\alpha}{200}$ for any $\pi\in\Pi$. Now by \Cref{lemma:selection-dec}, it holds that $V^{\pi^{\star, \hat{j}}, \cG}_1(\emptyset)\ge \max_{\pi\in\Pi}V^{\pi, \cG}_1(\emptyset) - \frac{60\alpha}{200}\ge \max_{\pi\in\Pi}V^{\pi, \cG}_1(\emptyset) - \frac{60\alpha}{200}$, thus concluding that $\pi^{\star, \hat{j}}$ is an $\alpha$-team-optimal solution. For the sample complexity, since the choice of all parameters remains the same as that for learning the equilibrium, the sample complexity remains the same as for learning the equilibrium, i.e., quasi-polynomial. 
	
	For the time complexity, as we mentioned above, we need to adjust \Cref{alg:construct_mg} to ensure \Cref{eq:team-Q} can be solved computationally efficiently for the learned model. Specifically, 
	\begin{itemize}
		\item For \textbf{Example 2} and \textbf{Example 5} that satisfy the \textbf{Condition 1}, it suffices to estimate $\PP_{h}^{\hat{\cM}(\pi^{1:H})}(o_{h+1}\given \hat{c}_h, p_{h}, a_{\cth})$, $\hat{r}_{i, h}^{\hat{\cM}(\pi^{1:H})}(\hat{c}_h, p_h, a_{i, h})$ for each $i\in[n]$ instead of the original $\PP_{h}^{\hat{\cM}(\pi^{1:H})}(o_{h+1}\given \hat{c}_h, p_{h}, a_{h})$, $\hat{r}_{i, h}^{\hat{\cM}(\pi^{1:H})}(\hat{c}_h, p_h, a_{h})$. Then \Cref{eq:trans-es} and \Cref{eq:reward-es} in \Cref{alg:construct_mg} can be replaced as follows:
	\$
	&\PP_{h}^{\hat{\cM}(\pi^{1:H}), z}(z_{h+1}\given \hat{c}_h, \gamma_{\cth, h})\leftarrow \sum_{\substack{p_h, a_{\cth, h}, o_{h+1}}}\bm{1}[\chi_{h+1}(p_h, a_{\cth, h}, o_{h+1})=z_{h+1}]\\
	&\qquad\qquad\qquad\qquad\times\PP_{h}^{\hat{\cM}(\pi^{1:H})}(p_h\given \hat{c}_h)\gamma_{\cth, h}(a_{\cth, h}\given p_{\cth, h})\PP_{h}^{\hat{\cM}(\pi^{1:H})}(o_{h+1}\given \hat{c}_h, p_{h}, a_{\cth, h})\\
	&\hat{r}_{i, h}^{\hat{\cM}(\pi^{1:H})} (\hat{c}_h, \gamma_{i, h})\leftarrow \sum_{p_h, a_{i, h}}\PP_{h}^{\hat{\cM}(\pi^{1:H})}(p_h\given \hat{c}_h)\gamma_{i, h}(a_{i, h}\given p_{i, h})\hat{r}_{i, h}^{\hat{\cM}(\pi^{1:H})}(\hat{c}_h, p_h, a_{i, h}).
\$
	With the modified construction, it is direct to verify that \Cref{eq:cond1-q} of \Cref{lem:single-controller} still holds for $\hat{\cM}(\pi^{1:H})$. Thus, one can solve \Cref{eq:team-Q} computationally efficiently.
	\item For \textbf{Example 3} and \textbf{Example 4} that satisfy \textbf{Condition 2}, \Cref{alg:construct_mg} requires no modifications since the learned model automatically satisfies \textbf{Condition 2}.
	\item For \textbf{Example 1} under the assumption of \textbf{Part (1)} in \textbf{Condition 3}, it suffices to estimate $\PP^{\hat{\cM}(\pi^{1:H})}_h(p_{i, h}\given \hat{c}_{i, h})$, $\PP_{h}^{\hat{\cM}(\pi^{1:H})}(o_{i, h+1}\given \hat{c}_{i, h}, p_{i, h}, a_{i, h})$, $\hat{r}_{i, h}^{\hat{\cM}(\pi^{1:H})}(\hat{c}_{i, h}, p_{i, h}, a_{i, h})$ separately for each $i\in[n]$ instead of the original joint $\PP^{\hat{\cM}(\pi^{1:H})}_h(p_{ h}\given \hat{c}_{ h})$, $\PP_{h}^{\hat{\cM}(\pi^{1:H})}(o_{h+1}\given \hat{c}_h, p_{h}, a_{h})$, $\hat{r}_{i, h}^{\hat{\cM}(\pi^{1:H})}(\hat{c}_h, p_h, a_{h})$. Then \Cref{eq:trans-es} and \Cref{eq:reward-es} can be replaced as follows for each $i\in[n]$
	\$
	&\PP_{h}^{\hat{\cM}(\pi^{1:H}), z}(z_{i, h+1}\given \hat{c}_{i, h}, \gamma_{i, h})\leftarrow \sum_{\substack{p_{i, h}, a_{i, h}, o_{i, h+1}}}\bm{1}[\chi_{i, h+1}(p_{i, h}, a_{i, h}, o_{i, h+1})=z_{i, h+1}]\\
	&\qquad\qquad\qquad\qquad\times\PP_{h}^{\hat{\cM}(\pi^{1:H})}(p_{i, h}\given \hat{c}_{i, h})\gamma_{i, h}(a_{i, h}\given p_{i, h})\PP_{h}^{\hat{\cM}(\pi^{1:H})}(o_{i, h+1}\given \hat{c}_{i, h}, p_{i, h}, a_{i, h})\\
	&\hat{r}_{i, h}^{\hat{\cM}(\pi^{1:H})} (\hat{c}_{i, h}, \gamma_{i, h})\leftarrow \sum_{p_{i, h}, a_{i, h}}\PP_{h}^{\hat{\cM}(\pi^{1:H})}(p_{i, h}\given \hat{c}_{i, h})\gamma_{i, h}(a_{i, h}\given p_{i, h})\hat{r}_{i, h}^{\hat{\cM}(\pi^{1:H})}(\hat{c}_{i, h}, p_{i,h}, a_{i, h}).
	\$
	With the modified construction, it is direct to verify that \Cref{eq:cond3-q} of \Cref{factor} still holds for $\hat{\cM}(\pi^{1:H})$. Thus, one can solve \Cref{eq:team-Q} computationally efficiently.
	\end{itemize}
	Now, since we have called the planning algorithm (i.e., \Cref{alg:avi}) only polynomial times, the total time complexity is also quasi-polynomial by \Cref{thm:plan_cases_dec}.
\end{prevproof}

\end{document}